\RequirePackage{silence}
\RequirePackage{fix-cm}
\documentclass[pdflatex, sn-nature]{sn-jnl}

\usepackage{graphicx}%
\usepackage{microtype}

\graphicspath{
  {figures/experiment-1/}
  {figures/experiment-1a/}
  {figures/experiment-2/}
  {figures/experiment-3/}
  {figures/experiment-4/}
}

\usepackage{multirow}%
\usepackage{amsmath,amssymb,amsfonts}%
\usepackage{amsthm}%
\usepackage[title]{appendix}%
\usepackage{xcolor}%
\usepackage{textcomp}%
\usepackage{manyfoot}%
\usepackage{booktabs}
\usepackage{bookmark}
\usepackage{algorithm}%
\usepackage{algorithmicx}%
\usepackage{algpseudocode}%
\usepackage{listings}%
\usepackage{cleveref}
\usepackage{placeins}
\usepackage{subcaption}            
\usepackage{tabularx}     
\usepackage{graphicx}
\usepackage{makecell}
\usepackage{longtable}
\newcolumntype{C}{>{\centering\arraybackslash}X}
\newcolumntype{L}{>{\raggedright\arraybackslash}X}

\usepackage{bm,amssymb,amsmath}

\makeatletter
\def\th@plain{%
  \thm@notefont{}
  \itshape 
}
\def\th@definition{%
  \thm@notefont{}
  \normalfont 
}
\makeatother

\def\1{\bm{1}}

\def\rvtheta{{\boldsymbol{\theta}}}

\def\rvs{{\mathbf{s}}}

\def\rvy{{\mathbf{y}}}

\def\rmB{{\mathbf{B}}}

\def\rmI{{\mathbf{I}}}

\def\rmM{{\mathbf{M}}}

\def\rmS{{\mathbf{S}}}

\def\rmX{{\mathbf{X}}}
\def\rmY{{\mathbf{Y}}}
\def\rmZ{{\mathbf{Z}}}

\def\vzero{{\bm{0}}}

\def\vh{{\bm{h}}}

\DeclareMathAlphabet{\mathsfit}{\encodingdefault}{\sfdefault}{m}{sl}
\SetMathAlphabet{\mathsfit}{bold}{\encodingdefault}{\sfdefault}{bx}{n}

\newcommand{\rmOmega}{\bm{\Omega}}

\begin{document}

\title[]{Meta-Amortized Cognitive Modeling}


\author*[1]{\fnm{Jerry M.} \sur{Huang}}\email{huangm13@rpi.edu}

\author[2]{\fnm{Lukas} \sur{Schumacher}}\email{lu.schumacher@unibas.ch}

\author[3]{\fnm{Niek} \sur{Stevenson}}\email{niek.n.stevenson@gmail.com}

\author[4]{\fnm{Mischa} \sur{von Krause}}\email{mischa.vonkrause@uni-siegen.de}

\author[1]{\fnm{Stefan T.} \sur{Radev}}\email{radevs@rpi.edu}

\affil*[1]{\orgdiv{Center for Modeling, Simulation, and Imaging in Medicine (CeMSIM)}, \orgname{Rensselaer Polytechnic Institute}, \orgaddress{\street{110 8th Street}, \city{Troy}, \postcode{12180}, \state{NY}, \country{USA}}}

\affil[2]{\orgdiv{Center for Economic Psychology}, \orgname{University of Basel}, \orgaddress{\city{Basel}, \country{Switzerland}}}

\affil[3]{\orgname{University of Amsterdam}, \orgaddress{\city{Amsterdam}, \country{Netherlands}}}

\affil[4]{\orgname{University of Siegen}, \orgaddress{\city{Siegen}, \country{Germany}}}


\abstract{
Cognitive modeling is highly iterative. 
Researchers repeatedly revise model structures, parameterizations, covariates, and experimental designs as theories evolve and data accumulates.
Amortized Bayesian inference with neural networks provides rapid estimation for individual models, but every revision requires new simulations and training.
Here, we introduce CogFormerV2, a meta-amortized framework that enables researchers to specify, revise, and fit cognitive models with a single network, carrying amortization across an entire space of models all at once.
Trained once in about a day on a single GPU, it delivers accurate, well-calibrated posteriors across more than 14.6 billion model configurations spanning binary, multi-alternative, and continuous decisions, with varying parameter sets, covariate structures, trial counts, and contamination levels.
It can fit empirical data from distinct behavioral tasks and extends to new models through continual learning and representation transfer. 
Released as an open-source interactive application, CogFormerV2 provides a reusable infrastructure for meta-amortized inference in cognitive modeling, while carrying the potential to accelerate computational modeling workflows at large.
}


\maketitle
\pagebreak

\section*{Introduction}
\label{sec:intro}

Cognitive models connect theories of the mind to empirical data through mathematical models with interpretable parameters. For instance, evidence accumulation models formalize decision-making as the accumulation of (noisy) evidence toward a response threshold \cite{ratcliffDiffusionModel2004,brownSimplestComplete2008,donkin2018response}. By explicating how \emph{forward} models generate observable behavior (e.g., response time and choices), researchers can compare different explanations of decision-making phenomena (e.g., which cognitive constructs account for speed-accuracy trade-offs) and test them against empirical data. Testing these explanations requires solving the \emph{inverse} problem, that is, inferring the parameters from the observations. For complex models, the inverse problem can be computationally challenging \citep{palestro2018likelihood, voss2019sequential, turner2018approximating, radev2020bayesflow, fengler2021likelihood}.

Neural simulation-based inference \citep[SBI;][]{cranmer2020frontier, deistler2025simulation} provides a solution for any model that can be simulated.
Specifically, \emph{amortized SBI} \cite{burkner2023some, zammitmangion2025neural} concentrates the bulk of the computation in an initial phase of simulation and neural network training. Once trained, a generative network can approximate the posterior distribution $p(\rvtheta \mid \rvy^*, \mathcal{M})$ under model $\mathcal{M}$ for any new observation $\rvy^*$ almost instantly.
Recent advances in amortized inference have led to a growing body of successful applications across a diverse array of cognitive models such as speeded decision-making, dynamic decision processes, judgments, risky choice, and cognitive conflicts, to name a few \cite{radev2020amortized, wieschen2020jumping, fengler2021likelihood, boelts2022flexible, von2022mental, ghaderi2023general, sokratous2023ask, schumacher2023superstat, schumacher2024validation, rmus2024artificial, hato2025, schaefer2025conflictModels, scholten2026diffusionLens, kvam2024using, kvam2025comparing, belov2026supervised, wu2026testing}.

However, scientific inquiry rarely involves fitting a single model. Before collecting data, researchers use \emph{in silico} experiments to iteratively modify model assumptions and evaluate how these choices affect their conclusions \citep{Heck2019Maximizing}. For example, an experimental manipulation may affect the speed of evidence accumulation, response caution, or both \citep{voss2019sequential, von2022mental}. Researchers may also define alternative model configurations by introducing different covariates and examining whether additional parameters, such as across-trial variability \citep{evans2019assessing} and urgency \citep{hawkins2015revisiting}, are necessary to fit the data. Each such modification introduces a new inverse problem, requiring the model to be fitted and evaluated anew.

While modern software frameworks, \citep[e.g.,][]{fengler2026hssm, stevenson2026bayesian}, simplify the process of specifying complex models, conventional amortized inference remains tied to individual model specifications. Each newly specified model requires its own training and validation to verify computational faithfulness, a process that often takes hours to days. Thus, as researchers repeatedly amortize single models within \emph{spaces of candidate models}, the computational cost can quickly add up, especially for large-scale datasets \citep{von2022mental, von2026scaling}. 

This discussion reveals the need to extend amortization beyond individual models, harmonizing it with an iterative Bayesian workflow of model fitting, checking, and refinement across candidate models \cite{gelman2020bayesian, gelman2026bayesian}. Thus, the computational bottleneck shifts to a higher level of model abstraction. What is needed is a modeling taxonomy and a flexible inference engine operating over a large sets of model configurations.

Meta-amortization addresses this issue by extending amortized inference over a space of model configurations, reducing the need to retrain neural networks for each change in model assumption \citep{wu2020meta}.
Recent work in SBI has already begun to demonstrate this possibility \citep{huang2026cogformer, chang2025amortized, kipnis2025metabeta, elsemuller2023sensitivity, schroder2023simultaneous, gloeckler2024all}. Further, prior-fitted networks such as TabPFN have further illustrated how pretraining a network across diverse tasks can produce highly reusable estimators \cite{hollmann2022tabpfn, hollmann2025accurate, vetter2026effortless}. However, the benefits of meta-amortization as enabled by these developments have remained largely untapped in cognitive modeling.

\begin{figure}
    \centering
    \includegraphics[width=\linewidth]{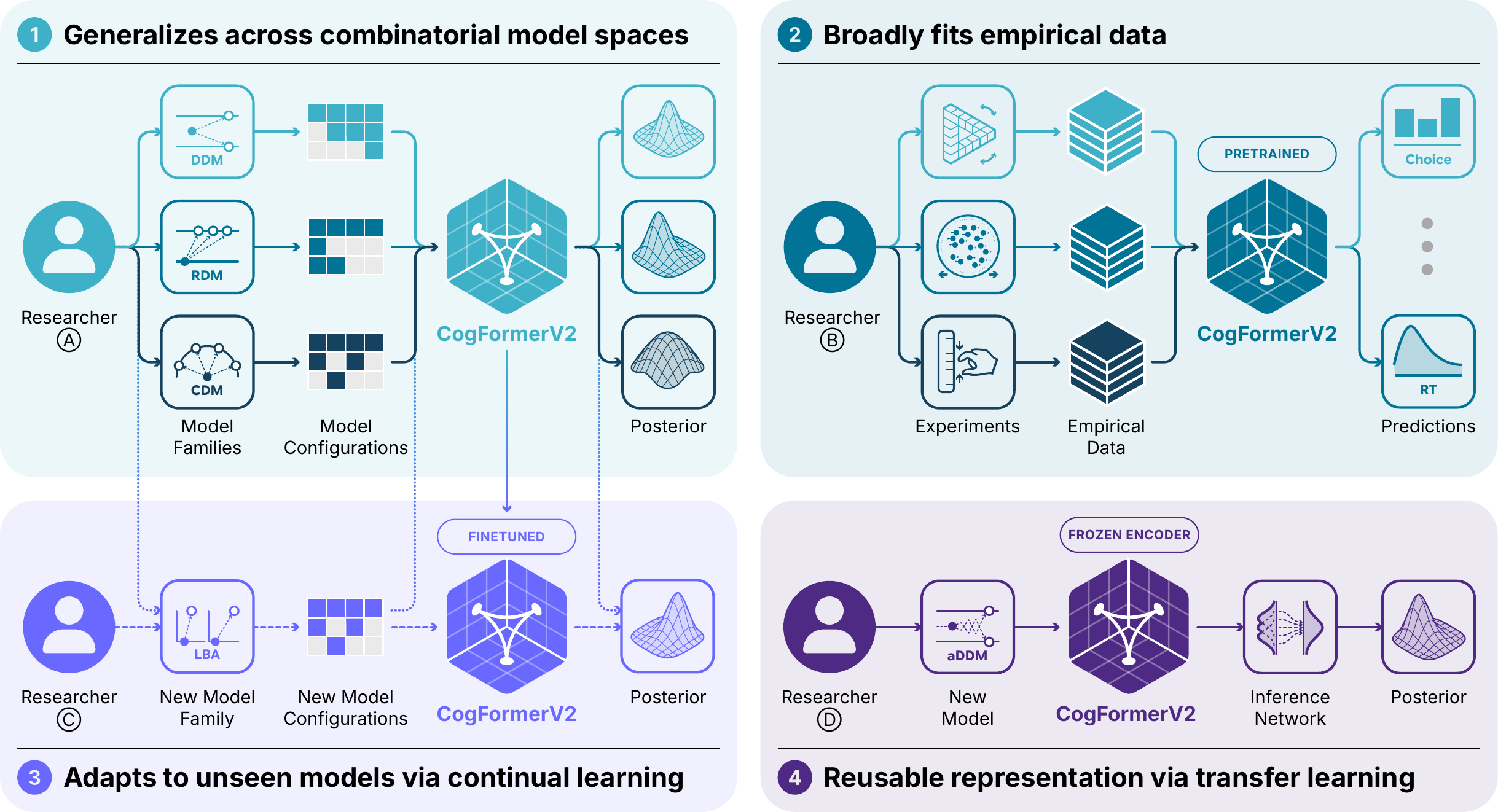}
    \caption{\textbf{Overview of CogFormerV2's capabilities}. Trained for decision-making processes involving binary, multi-alternative, and continuous responses, CogFormerV2 1) enables accurate and well-calibrated inference across combinatorial model spaces, 2) provides excellent posterior predictive performance when applied to empirical datasets, 3) can adapt to new model families via continual learning, and 4) has reusable representations that can adapt to new models via transfer learning.}
    \label{fig:cogformerv2-main}
\end{figure}

In this paper, we develop CogFormerV2, a meta-amortized inference framework that brings this vision to models of decision-making and aims to provide a blueprint for amortizing any domain of cognitive modeling. Building on an initial prototype \cite{huang2026cogformer}, CogFormerV2 uses a single generative architecture pretrained for only a day on a single GPU. It delivers fast posterior inference across a combinatorial space of more than 14.6 billion configurations of decision model families of binary, multi-alternative, and continuous responses. Researchers can vary which parameters to estimate or to fix, assign discrete or continuous covariates to parameters, and accommodate different numbers of trials or degrees of contamination, all without retraining.

Across extensive simulations and empirical applications, CogFormerV2 demonstrates excellent estimation and predictive performance across distinct decision-making tasks. It enables static design optimization by estimating the information gain over candidate designs, varying trial counts, covariates, or conditioning. Beyond this scope, continual learning incorporates previously unseen model families, while learned representations transfer to new decision-making models (for an overview, see \autoref{fig:cogformerv2-main}). An interactive, open-source application makes these \emph{in silico} capabilities directly available to researchers (\href{https://jerrymhuang--cogformer-demo-web.modal.run/}{https://jerrymhuang--cogformer-demo-web.modal.run/}).

These capabilities extend amortized inference from a single-use tool for individual models to a reusable engine for iterative model testing. Since CogFormerV2 learns across a broad space of modeling choices, it brings fast Bayesian inference closer to the iterative process of theory testing, so that researchers can refine model and experiment design in parallel and at a greatly accelerated pace.

\section*{Results}
\label{sec:results}

We trained CogFormerV2 on simulated data from three widely used evidence accumulation model families: 1) the diffusion decision model \citep[DDM;][]{ratcliff2008diffusion}, 2) the racing diffusion model \citep[RDM;][]{tillman2020racingDiffusion}, and 3) the circular diffusion model \citep[CDM;][]{smith2016circularDiffusion}. We assess its inference accuracy against thousands of model configurations. Through continual learning and transfer learning, we extend the pretrained CogFormerV2 to previously unseen model families, specifically the linear ballistic accumulator \citep[LBA;][]{brown2008simplest} and the attentional drift diffusion model \citep[aDDM;][]{krajbich2012attentional}. 
The following four large-scale experiments evaluate different aspects of CogFormerV2 across these models.
Further details on the simulation setup, training procedure, model architectures, and evaluation metrics are provided in \autoref{sec:methods}.

\subsection*{Experiment 1: CogFormerV2 generalizes across model spaces}
\label{ssec:generalization}

\begin{figure}[tp]
  \centering
  \includegraphics[width=\textwidth]{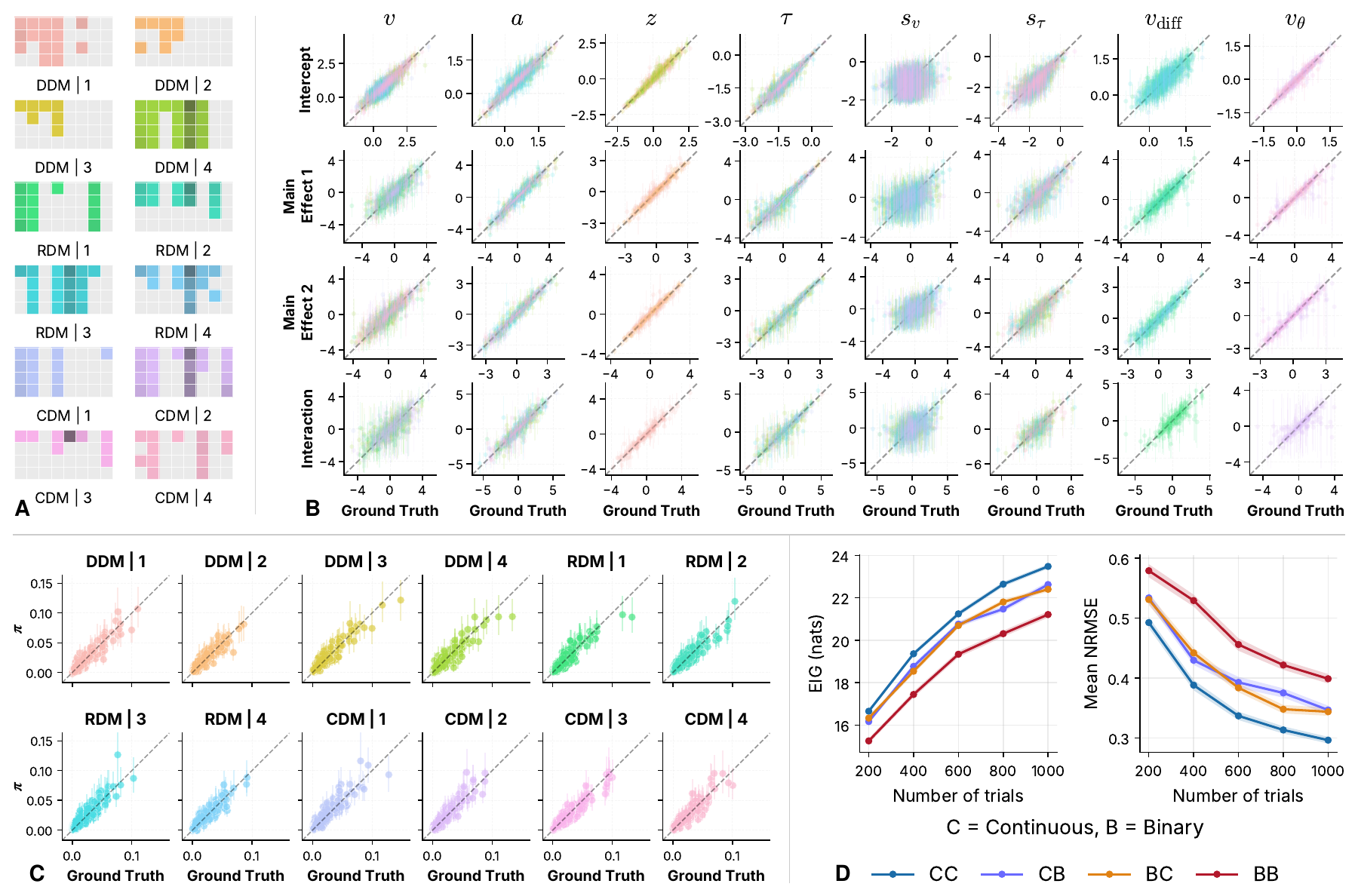}
  \caption{\textbf{Amortized inference across model spaces}. A) 12 randomly sampled model configurations within a model space of two interacting predictors, with each predictor being either continuous or dichotomous. Their respective inference targets in the model space are indicated by non-gray color shades. Brighter hues suggest better parameter recovery. B) Parameter recovery for the sampled model configurations in the model space. C) Estimates of the contamination probability $\pi$ for each design. D) Expected information gain and NRMSE measured per increasing trial counts when each of the two DDM predictors is continuous [C] or binary [B].}
  \label{fig:exp1-main}
\end{figure}

The consistency of CogFormerV2's inference quality across the model space must be examined to ensure that its statistical accuracy is systematic within its amortization scope. To evaluate, we first test whether our model trained on the DDM, RDM, and CDM model families can accurately amortize across their combinatorial model configurations. Here, a \emph{model configuration} comprises choices of which model parameters are estimated or fixed, which categorical or continuous predictors affect them, and which interactions between predictors are included. After just one day of training on a single GPU, CogFormerV2 opens up accurate Bayesian estimation across a combinatorial space of 14,609,856,204 model configurations without retraining, not counting generalization over sample sizes. 

Crucially, we achieve this generalization scope with only 192 million simulated training datasets, accounting for about 1.3\% of the entire model space. This simulation budget is far smaller than the total number of possible models. For comparison, training a separate neural estimator for each model configuration would require approximately 36.5 billion training hours at the assumed per-design training cost (\autoref{ssec:exp1-protocol}). Thus, training CogFormerV2 only once makes an otherwise prohibitive volume of candidate models accessible for inference and static design optimization.

\begin{table}[!htp]
\centering
\small
\caption{Comparison of absolute metrics (median $\pm$ SEM) between BayesFlow (baseline) and CogFormerV2, aggregated across the five benchmark model configurations and all parameters in each configuration.}
\label{tab:amortization-absolute-metrics}
\begin{tabular*}{\textwidth}{@{\extracolsep{\fill}}llccc@{}}
\toprule
\textbf{Model} & \textbf{Method} & \textbf{NRMSE} & \makecell{\textbf{Calibration}\\\textbf{error}}
& \makecell{\textbf{Posterior}\\\textbf{contraction}} \\
\midrule
DDM & BayesFlow & $0.046 \pm 0.007$ & $0.017 \pm 0.002$ & $0.978 \pm 0.023$ \\
    & CogFormerV2 & $0.054 \pm 0.007$ & $0.023 \pm 0.002$ & $0.968 \pm 0.024$ \\
\midrule
RDM & BayesFlow & $0.075 \pm 0.007$ & $0.019 \pm 0.002$ & $0.928 \pm 0.027$ \\
    & CogFormerV2 & $0.082 \pm 0.007$ & $0.018 \pm 0.002$ & $0.936 \pm 0.027$ \\
\midrule
CDM & BayesFlow & $0.043 \pm 0.006$ & $0.013 \pm 0.001$ & $0.985 \pm 0.022$ \\
    & CogFormerV2 & $0.049 \pm 0.007$ & $0.020 \pm 0.002$ & $0.978 \pm 0.023$ \\
\bottomrule
\end{tabular*}
\end{table}

Across 1000 randomly sampled model configurations, CogFormerV2 shows well-calibrated posteriors ($\mathrm{ECE} = 0.021$) with accurate parameter recovery ($\mathrm{NRMSE} = 0.077$) (see \autoref{fig:exp1-main}-A and \autoref{fig:exp1-main}-B for results on 12 of these configurations). Despite model-specific identifiability constraints on some parameters, such as the drift rate variability $s_v$, CogFormerV2 also shows robustness to data contamination (see \autoref{fig:exp1-main}-C), with contamination probability $\pi$ also recovered consistently. Compared with compute-matched single-model networks, CogFormerV2 retains competitively low estimation error and excellent marginal calibration (\autoref{tab:amortization-absolute-metrics}).

Beyond parameter estimation, CogFormerV2 can help researchers ask which experimental design will inform them more about the model's parameters before they collect empirical data. To demonstrate, we compared model configurations with different mixtures of continuous and categorical predictors using expected information gain (EIG; see also \autoref{par:eig}), the anticipated reduction in uncertainty about model parameters after observing data. We observe that EIG increases and mean $\mathrm{NRMSE}$ steadily decreases with both the number of observations and the number of continuous predictors in model configurations (see \autoref{fig:exp1-main}-D). Specifically, increasing the number of trials from 200 to 1,000 increased expected information gain (EIG) by 5.96–6.83 nats (Shannon bits) and reduced NRMSE by 31–40\%.

\subsection*{Experiment 2: CogFormerV2 can broadly fit empirical data.}
\label{ssec:real-data}

\begin{figure}[htbp]
  \centering
  \includegraphics[width=0.96\textwidth]{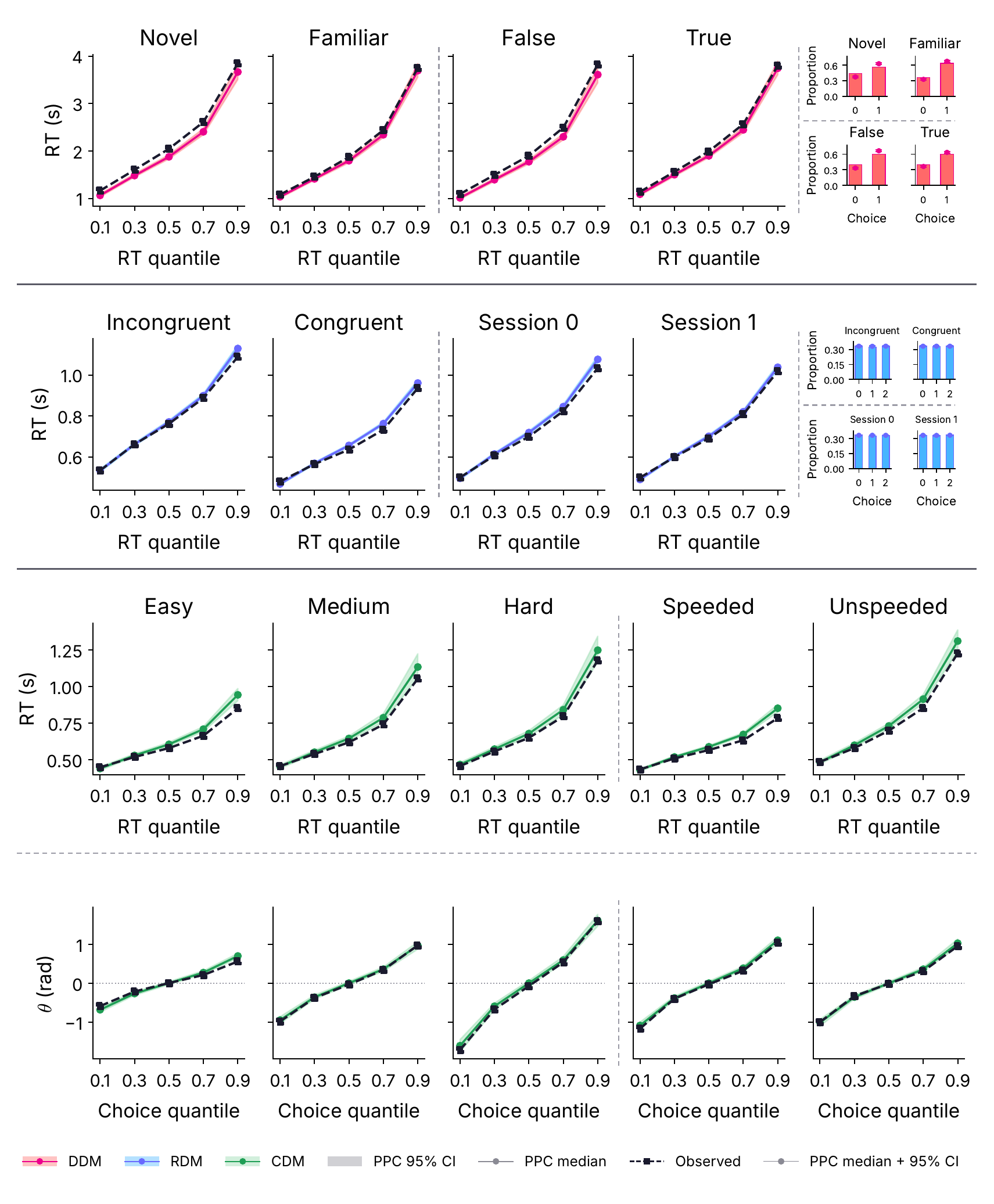}
  \caption{\textbf{Posterior predictive check on empirical data}. Observed data against the posterior predictive of CogFormer's fits, pooled over subjects: DDM (72 subjects; \textit{red}), RDM (149; \textit{blue}), and CDM (12; \textit{green}). Quantile panels show the response time (RT) and, for the CDM, the continuous response angle $\theta$ (final row) for each level of each experimental condition: observed (black squares, dashed) against the predictive median (colored line) with its 95\% credible interval. For the DDM and RDM, the bar plots on the right (first and second row) compare, per condition level, the observed proportion of each response (in bars) and the predictive median and 95\% interval (points with intervals).}
  \label{fig:exp2-main}
\end{figure}

CogFormerV2's ability to generalize over a broad spectrum of experimental tasks cannot be established by assessing its statistical accuracy on simulated data alone. To test whether its promising estimation performance translates to empirical data, we applied the pretrained CogFormerV2 from Experiment 1 to a truth judgment task using the DDM \cite{stump2026cognitive}, a Stroop task using the RDM \cite{stevensonBayesianHierarchical2026}, and a continuous-orientation judgment task using the CDM \cite{kvam2019geometric}. Posterior predictive checks suggest that all models are able to capture the distributions of RTs and choices and experimental effects across all three tasks, with closely aligned response error prediction, measured in angle, for the continuous-orientation judgment task \autoref{fig:exp2-main}. This experiment demonstrates CogFormerV2’s reusability across distinct empirical data under different model families, highlighting its practicality.

The remaining discrepancy in prediction focuses on slower responses and effect magnitudes. The DDM tended to underestimate RTs (quantile $\mathrm{MAE} = 110\,\mathrm{ms}$), whereas the CDM overestimated slower responses (quantile $\mathrm{MAE} = 34\,\mathrm{ms}$) despite closely matching response angle distributions. The RDM also showed minor RT discrepancies towards slower RTs (quantile $\mathrm{MAE} = 15\,\mathrm{ms}$). Some observed summaries fell outside the posterior predictive intervals, indicating that CogFormerV2's generalizability across tasks did not eliminate detectable condition-specific misfit, which may be attributed to model specification and may not be considered one of CogFormerV2's shortcomings.

\subsection*{Experiment 3: CogFormerV2 can adapt to unseen models via fine-tuning.}
\label{ssec:fine-tuning}

\begin{figure}[htbp]
  \centering
  \includegraphics[width=\textwidth]{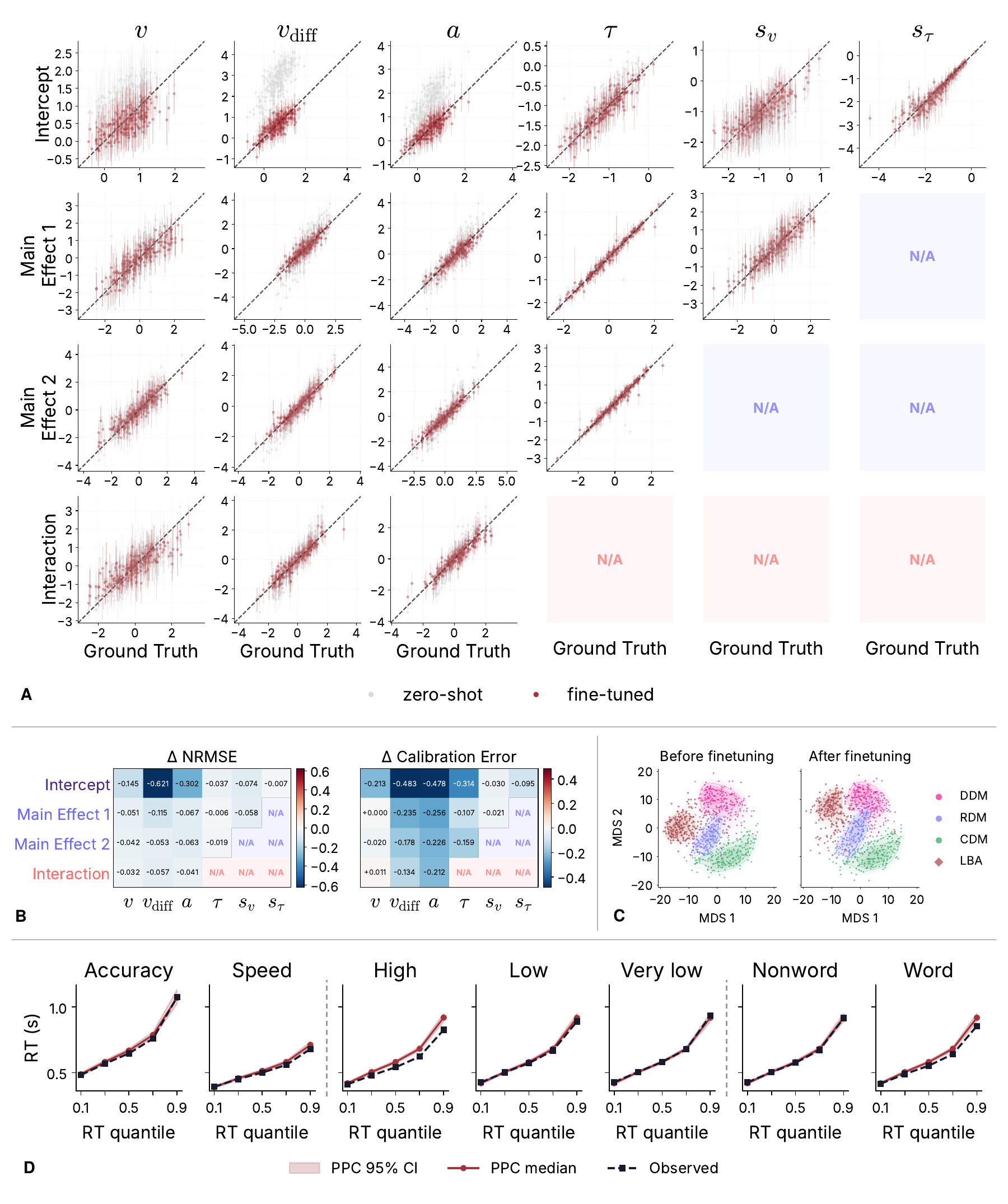}
  \caption{\textbf{Fine-tuning to unseen models}. A) Parameter recovery for the LBA before (\textit{gray}) and after fine-tuning (\textit{red}). B) Change in each coefficient's NRMSE and calibration error with fine-tuning on the same design. Negative values (\textit{blue}) reflect improvements. C) Multi-dimensional scaling (MDS) showing how the encoder representation of LBA shifts before and after the fine-tuning procedure while keeping the existing model families unchanged. D) Observed (\textit{black}) and resimulated (\textit{red}) response time (RT) quantiles for the lexical-decision task data from \cite{wagenmakers2008diffusion} obtained after fine-tuning.}
  \label{fig:exp3-main}
\end{figure}

Adaptation to previously unseen model families could significantly expand CogFormerV2's meta-amortization scope whenever needed, which can further enhance its reusability. To test this, we take the pretrained CogFormerV2 from Experiments 1 and 2 to account for the LBA model family \cite{brown2008simplest}, which has a different data-generating structure for evidence accumulation. 

Even before fine-tuning, CogFormerV2 can already partially recover LBA parameters, although the marginal posteriors are miscalibrated. Fine-tuning improved both parameter recovery and calibration.
For instance, in the illustrated benchmark model configuration (\autoref{fig:exp3-main}-AB), the median $\mathrm{NRMSE}$ dropped from $0.228$ to $0.171$ after continual learning with only an additional 12.8 million simulations (6.6\% of the original simulation budget).
Thus, CogFormerV2 is easily extensible to spaces of unseen model configurations beyond the original training distributions. 

\begin{table}[t]
  \centering
  \small
  \caption{Validation metrics on the original model families in CogFormerV2's amortization scope before and after fine-tuning with the LBA, evaluated on the same benchmark model configuration in \autoref{fig:exp3-main}-A. }
  \label{tab:exp3-forgetting-metrics}
  \medskip
  \begin{tabularx}{\textwidth}{@{}l*{3}{>{\centering\arraybackslash}X}@{}}
    \toprule
    \textbf{Metric} & \textbf{DDM} & \textbf{RDM} & \textbf{CDM} \\
    \midrule
    NRMSE ($\downarrow$)
      & $0.058 \to 0.058$ $(0.000)$
      & $0.100 \to 0.100$ $(0.000)$
      & $0.059 \to 0.059$ $(0.000)$ \\
    \addlinespace
    Calibration error ($\downarrow$)
      & $0.023 \to 0.014$ $(-0.009)$
      & $0.020 \to 0.017$ $(-0.003)$
      & $0.019 \to 0.017$ $(-0.002)$ \\
    \addlinespace
    Posterior contraction
      & $0.963 \to 0.964$ $(+0.001)$
      & $0.864 \to 0.861$ $(-0.003)$
      & $0.974 \to 0.974$ $(+0.001)$ \\
    \bottomrule
  \end{tabularx}
\end{table}

The expanded inference generalization also couples with the stable encoder representation and parameter recovery for the original model families (see \autoref{fig:exp3-main}-C and \autoref{tab:exp3-forgetting-metrics}). This allows CogFormerV2 to accommodate the LBA along with its existing inference tasks. While the estimator remains generally well-calibrated after fine-tuning, adaptability comes with a minor cost in accuracy. Across LBA benchmark designs and parameters, median NRMSE was $0.134$ for the fine-tuned network versus $0.076$ for single-model baselines, with generally lower posterior contraction. Nevertheless, one can easily mitigate this trade-off with more fine-tuning, without discounting CogFormerV2's flexible generalization.

The extension of the amortization scope also translates to empirical applications. To demonstrate, we applied the fine-tuned CogFormerV2 to the lexical-decision data of \cite{wagenmakers2008diffusion} (see \autoref{fig:exp3-main}-D). The adapted LBA estimator reproduced the broad RT distributions and speed-accuracy shift in these data (quantile $\mathrm{MAE} = 18\,\mathrm{ms}$ across all three conditions). Agreement was closest for faster and intermediate responses, while over-prediction of slower responses becomes the most apparent for high-frequency words, whose predicted 90th-percentile RT was approximately $0.919$ s compared with the observed $0.826$ s. Overall, these results reinforce CogFormerV2's reusability, showing how it can expand to additional cognitive models and adapt to new experimental designs, while retaining meaningful inference for the models from its original amortization scope. 

\subsection*{Experiment 4: CogFormerV2's encoder representation can be reused in new models.}
\label{ssec:encoder-rep}

\begin{figure}[t]
  \centering
  \includegraphics[width=0.29\textwidth]{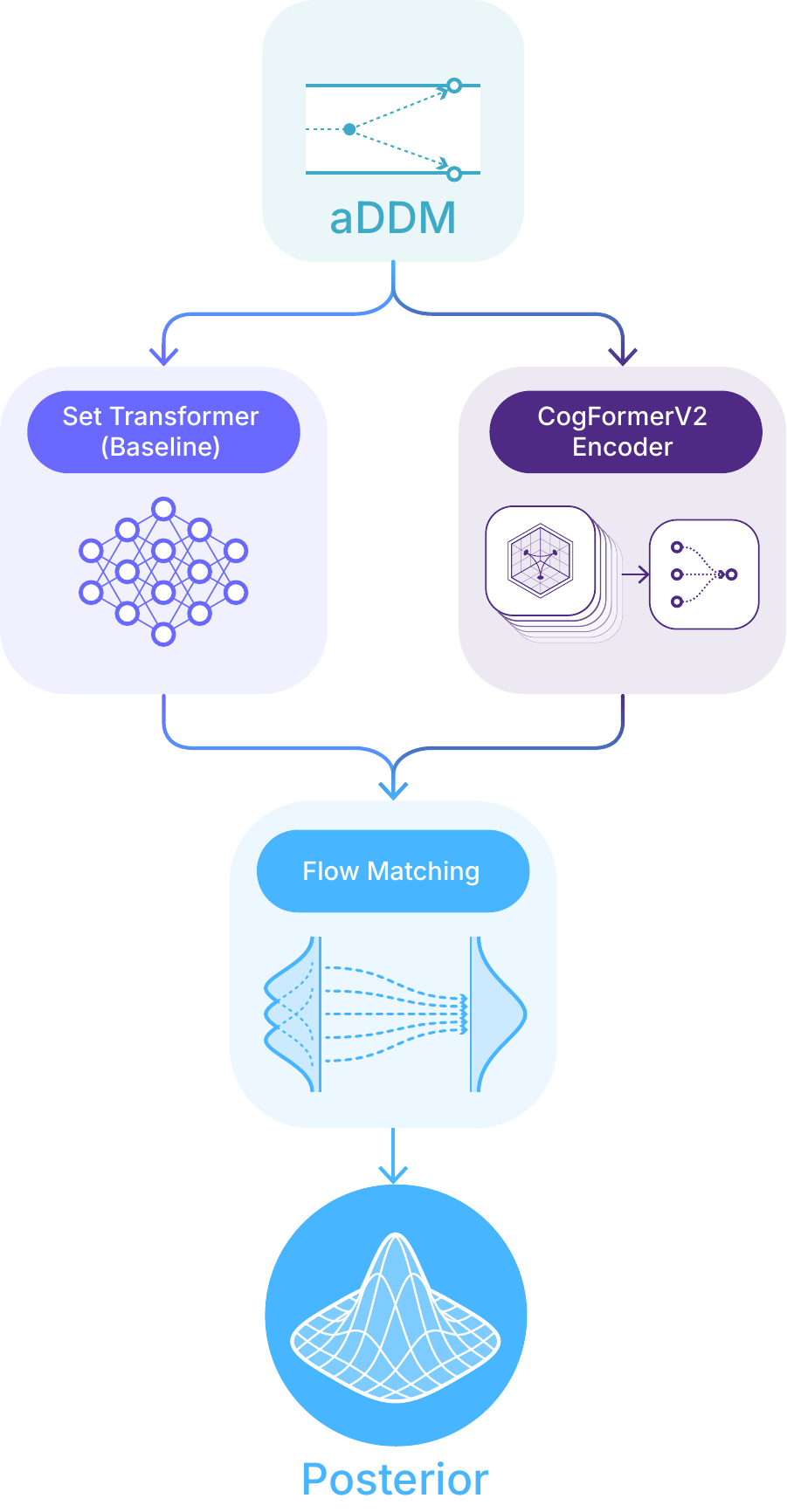}
  \includegraphics[width=0.70\textwidth]{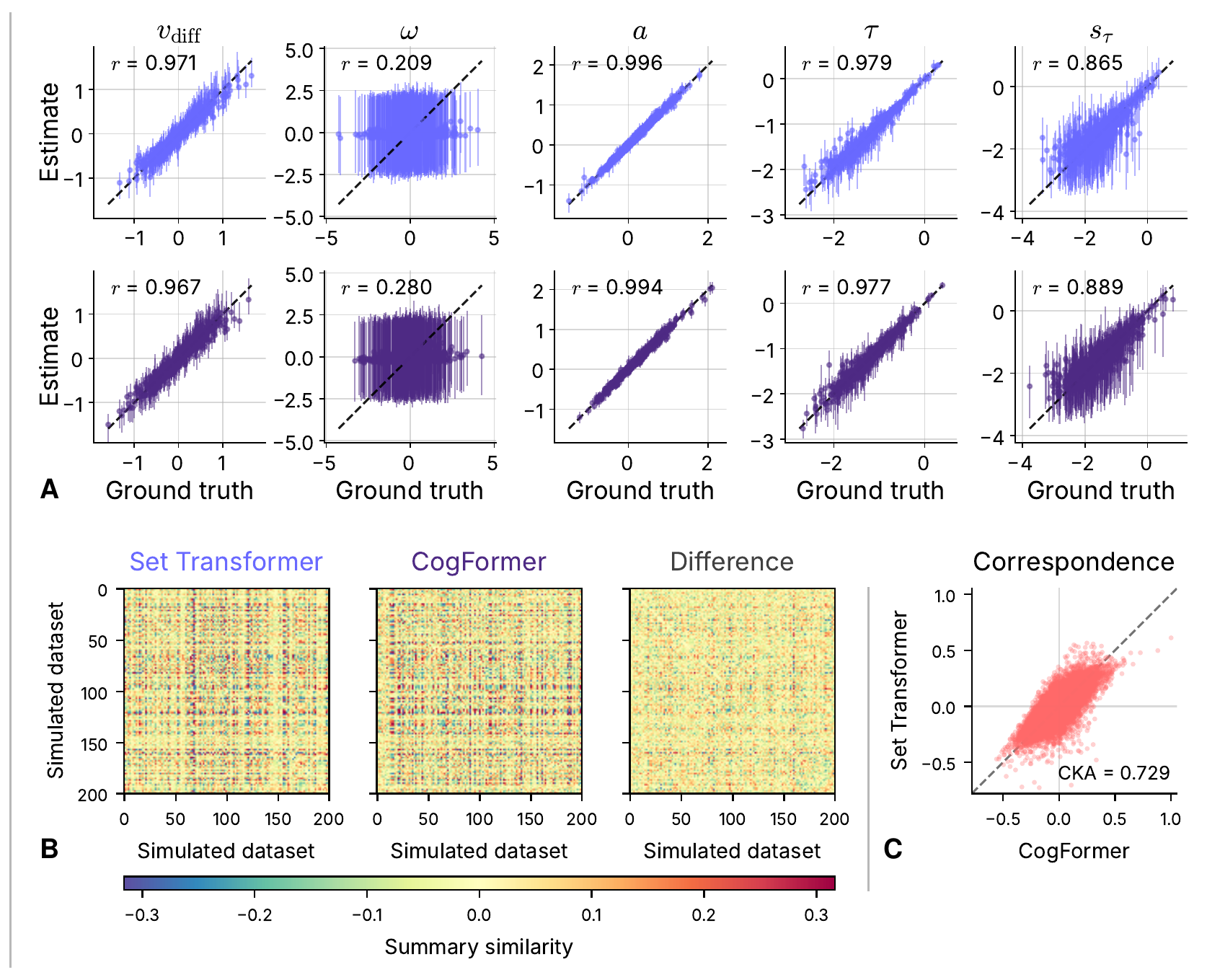}
  \caption{\textbf{CogFormerV2's frozen encoder transfers to new models.} A) attentional drift-diffusion model \citep[aDDM;][]{krajbich2012attentional}) parameter recovery using a task-specific Set Transformer summary (upper row) or CogFormerV2's frozen encoder (lower row), with matched posterior architectures and training budgets. Points show posterior medians and 95\% credible intervals against ground truth for $500$ validation datasets per arm; $r$ denotes correlation.  B) similarity matrices for the learned and frozen aDDM summaries, their difference, and C) pairwise correspondence, quantified by linear center kernal alignment (CKA).}
  \label{fig:exp4-main}
\end{figure}

Beyond its adaptability via fine-tuning, CogFormerV2's encoder summaries could also be reused for an unseen model without updates. In neural SBI, a summary network typically compresses a dataset into a learned representation that the inference network uses to infer the posteriors \citep{radev2020bayesflow}. Here, we train only a new inference network on CogFormerV2's summaries and extend inference to the aDDM. We compare the resulting network to an estimator whose summary network was trained specifically for the aDDM. As observed from \autoref{fig:exp4-main} (\emph{top}), parameter recovery is largely comparable between the two approaches. Neither approach reliably identified the attentional discount parameter under this design, which indicates a lack of identifiability. However, the encoder summary retained sufficient information to recover the other free parameters, extending its utility beyond the models that shaped its training.

Successful representation transfer did not require the frozen encoder to reproduce the representation learned specifically for the aDDM. Centered kernel alignment \cite{kornblith2019similarity}, a measure of similarity between how representations organize the same datasets, was $0.729$, compared with a mean of $0.795$ for random projections of the frozen representation to the task-specific summary's dimensionality. The observed alignment did not exceed this reference, despite comparable parameter recovery. Together, these results show that CogFormerV2's encoder can generate reusable summaries for a new inference task without reproducing the geometry of a dedicated summary network.

\section*{Discussion}
\label{sec:discussions}

The main contribution of CogFormerV2 is to bring amortized Bayesian inference closer to the iterative workflows of cognitive modeling. Instead of treating every revision of a model as a new inference problem that requires separate simulation and training, CogFormerV2 makes a vast space of modeling choices accessible through a single pretrained neural estimator. The latter allows modelers to explore alternative explanations of behavioral effects, revise parameter constraints, add covariates, and evaluate the informativeness of experimental designs before collecting data, all without retraining the network for each new model specification.

Our results demonstrate that this flexibility can be achieved while retaining competitive approximation accuracy across a combinatorial space of model configurations. Applications to empirical data further show that a single pretrained estimator can fit multiple model families to empirical data across different experimental tasks without retraining. Experiments 3 and 4 address a different question: whether and how the estimator can be adapted to model families beyond its original training scope. These extensions, however, come with limitations concerning continual learning, the generality of pretrained representations, and the difficulty of diagnosing errors.

Amortizing many inference problems may come at the cost of statistical accuracy. Experiment 1 shows that CogFormerV2 retains accurate, well-calibrated inference across model configurations, although at the cost of slightly elevated approximation error. This difference is referred to as an \emph{amortization gap} \citep{cremer2018inference, marino2018iterative}. Despite minor performance differences, CogFormerV2 allows researchers to investigate models they would otherwise leave unexplored by avoiding repeated training for the same model configuration.

The empirical applications in Experiment 2 raise a separate question: how well do the available models explain behavior? Across three distinct tasks, CogFormerV2 broadly reproduces RT distributions and experimental effects, despite mild but systematic over-prediction of longer RTs. These discrepancies illustrate the \emph{theory gap} between obtaining a posterior and establishing the adequacy of model assumptions. While accurate posterior predictions can allow researchers to stratify the inferential quality of the candidate models, one cannot automatically assume that the best model that explains the data is always the most optimal in its inference accuracy. In other words, meta-amortization alone does not provide sufficient information about how well-specified a model is against context-specific empirical observations. Thus, diagnostics for model misspecification could provide further evidence about when an apparently plausible analysis needs closer examination \citep{schmitt2024detecting, ward2022robust, wehenkel2024addressing}, and adaptive design optimization \citep{bracher2025jadai} with a pre-trained meta-amortizer could uncover the model configuration within the training distribution that best serves a designed experiment.

Experiments 3 and 4 also highlight the limitations of extending meta-amortization beyond its initial training scope. When zero-shot inference is insufficient, incorporating a new model family requires fine-tuning. However, this is considerably cheaper than training a new estimator from scratch, as demonstrated by the LBA fine-tuning in Experiment 3. A remaining challenge is to incrementally incorporate new model families without compromising inference on those already within the amortization scope. Since simulations from earlier model families can be generated on demand, replaying them alongside the new families offers a straightforward way to mitigate catastrophic forgetting \citep{rolnick2019experience}. Whether the original inference capabilities are fully retained after such updates nevertheless requires verification.

A further limitation concerns the generality of the learned representations. Experiment 4 shows that CogFormerV2's pretrained encoder enables transfer learning with a small inference head. However, this approach can only succeed if the encoder preserves the information needed by the new model. A model may depend on aspects of the data that are irrelevant for all families encountered during pretraining, such as the extreme tails of the RT distribution. There is no guarantee that the encoder retains such information, and training a new inference network cannot recover what has already been discarded. Whether a pretrained representation is informative for a new model requires additional verification.

Finally, we note that poor parameter recovery can reflect either weak identifiability in the generative model or a failure of the encoder to preserve relevant information, which we refer to as a \emph{representation gap}. The weak recovery of the attentional discount parameter in Experiment 4 illustrates this difficulty. In our case, we can compare the pretrained encoder against a task-specific network trained from scratch. Without such a reference, it would be difficult to determine whether poor recovery reflects intrinsic limits on identifiability or the learned representation. A general method for distinguishing these two scenarios remains an open problem for the field.

Despite the limitations discussed above, the present work opens up multiple potential directions for further extending the scope of meta-amortization. For instance, CogFormerV2 is implemented without accounting for non-stationary parameter fluctuations. Future work can integrate neural superstatistics \citep{schumacher2023superstat, schumacher2024validation} and compositional modeling \citep{arruda2026compositional} to extend meta-amortization to time-varying model parameters and hierarchical models. In addition, while CogFormerV2 provides reusable inference over evidence accumulation models, it could also fold in other models with different generative structures and observables into its meta-amortization scope (e.g., reinforcement learning), so that researchers can apply it to an entire domain of cognitive models.

\section*{Conclusion}

Cognitive science advances by positing theories of behavior, representing them as computational models, and confronting them with empirical evidence. Meta-amortization can accelerate this progress by providing a shared, adaptable computational infrastructure that allows researchers to apply inference tools to their chosen models and datasets without repeatedly training new estimators. This infrastructure also allows modelers to evaluate information gain before costly data collection. More broadly, the same approach could incorporate models in other scientific domains, helping meet the emerging demands of meta-modeling across science.

\pagebreak

\section*{Methods}
\label{sec:methods}

\subsection*{Taxonomy of amortization scopes}
\label{ssec:scopes}

\begin{figure}[t]
    \centering
    \includegraphics[width=\linewidth]{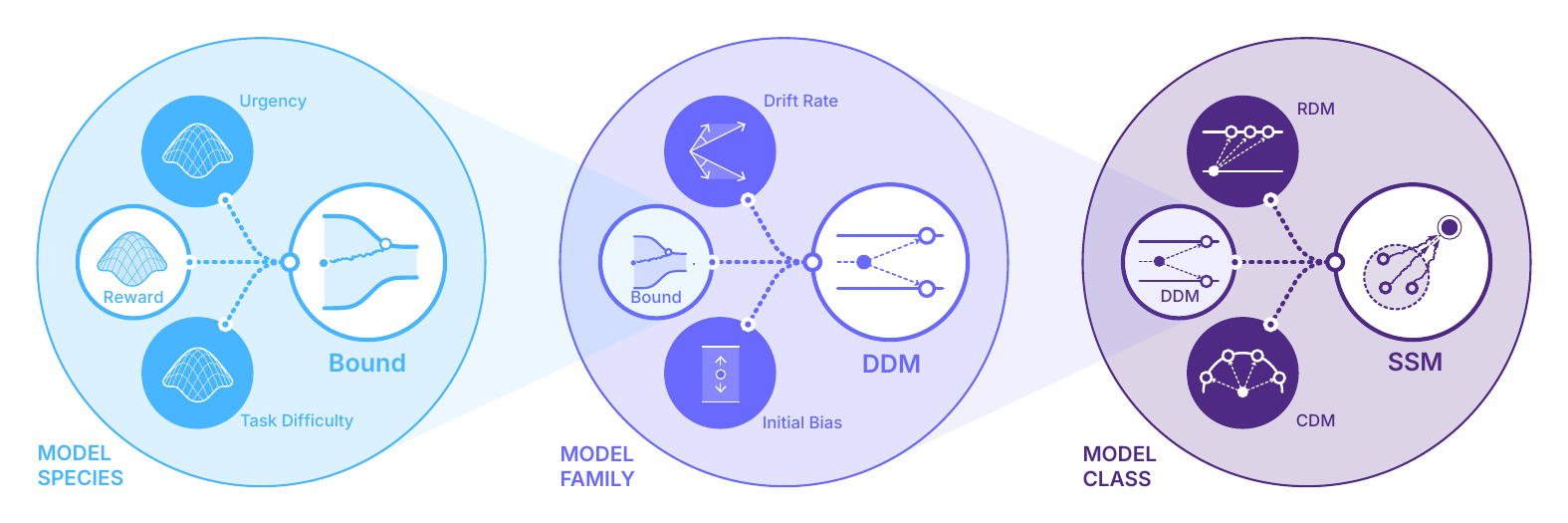}
    \caption{\textbf{Taxonomy of amortization scopes}. In cognitive modeling, one fits concrete model species $\mathcal{M}$ nested within model families $\mathcal{F}$ as part of a larger model class $\mathcal{C}$. Here, we train a neural architecture that generalizes across multiple model families in a given model class (here, evidence accumulation models).}
    \label{fig:levels}
\end{figure}

Amortized inference in cognitive modeling requires an explicit statement of what an estimator is amortized over. Borrowing the terminology of Linnaean ranks in ecology \cite{linné1774caroli}, we organize models into a three-level taxonomy: \textbf{model species}, \textbf{model family}, and \textbf{model class}, as shown in \autoref{fig:levels}. Multiple models at the same level of abstraction are grouped at the next higher level. The taxonomy fixes the level at which inference is amortized and, by extension, the generalization scope of a trained neural estimator.

A \textbf{model species} $\mathcal{M}$ is a single, fully specified model instance, with a fixed computational structure and a fixed set of parameters and predictors. For instance, a two-choice decision experiment may have reward, task difficulty, and urgency as predictors, which may be specified to modulate a subset of DDM parameters, such as the drift rate $\vartheta$ and the boundary separation $a$.  
This is the inference scope of conventional SBI applications in cognitive modeling, and any change in parameterization or generative structure requires retraining the neural network.

A \textbf{model family} $\mathcal{F}$ is a collection of multiple model species that share a computational structure but differ in parameterization or auxiliary assumptions.
For instance, variants of the DDM \citep{ratcliff1978diffusionModel, ratcliffDiffusionModel2004} differ in collapsing versus fixed boundaries \citep{bogaczPhysicsOptimal2006, ditterichEvidenceTimevariant2006, rasanan2025collapsing}, starting-point variability, non-decision time structure, or which parameters carry predictors.
Such specifications are typically nested: simpler variants follow from more general ones by fixing parameters to zero or another constant. Amortizing over a model family therefore requires generalizing across different model configurations within the same generative mechanism \cite{huang2026cogformer}.

A \textbf{model class} $\mathcal{C}$ is a collection of multiple model families that implement a shared computational principle.
In decision-making, a class of evidence accumulation models can consist of the DDM, the RDM, and the CDM together.
These families are not nested and cannot be reached from one another by parameter constraints.
Amortizing over a model class therefore requires generalizing across structurally distinct generative mechanisms while using what they share, such as their observables.

The meta-simulator under which we amortize functions at the level of the model class:
\begin{equation}\label{eq:generative}
  p(\mathcal{F},\mathcal{M},\mathcal{D},\rvtheta_{\mathcal{F}},\rmY \mid \mathcal{C})
  = p(\mathcal{F}\mid\mathcal{C})\,
      p(\mathcal{M},\mathcal{D}\mid\mathcal{F})\,
      p(\rvtheta_{\mathcal{F}}\mid\mathcal{M},\mathcal{D})\,
      p(\rmY\mid\rvtheta_{\mathcal{F}},\mathcal{M},\mathcal{D}),
\end{equation}
where $\mathcal{D}$ denotes a model species over the intrinsic parameters $\rvtheta_{\mathcal{F}}$ of family $\mathcal{F}$ and $\rmY$ denotes the model observables. Here, parameters are \emph{intrinsic} in the sense that they are the original parameters of the model family without considering any configurations introduced by the regression structure. The next sections describe how the abstract model ($\mathcal{M}$) and design ($\mathcal{D}$) variables are represented numerically, through a design matrix $\rmX$, a parameter mask $\rmM$, and a model-family embedding, and how CogFormerV2 processes them to approximate the amortized posterior. A summary of all notation is given in \autoref{tab:notation}.

\subsection*{Meta-simulator}
\label{sec:meta-simulator}

\begin{figure}
    \centering
    \includegraphics[width=0.99\linewidth]{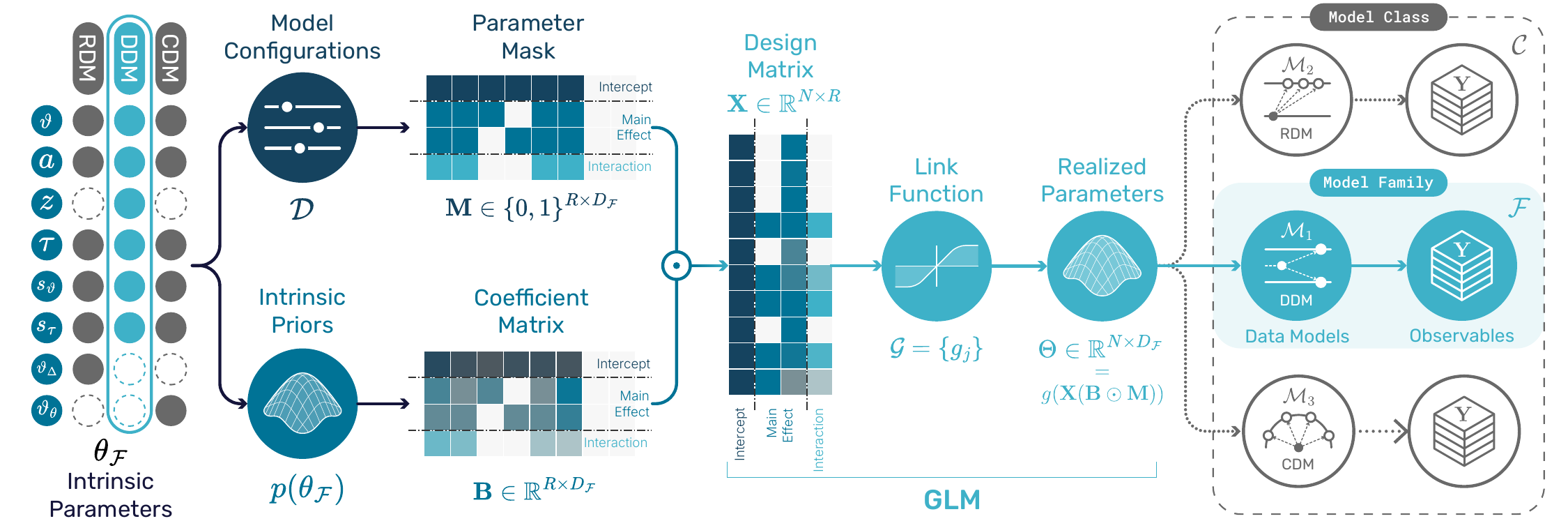}
    \caption{\textbf{Architecture of the meta-simulator}. For a single model family (\textit{middle}), a coefficient matrix $\rmB$ defines a generalized linear model (GLM), and a random binary parameter mask $\rmM$ selects its active entries. Together with a sampled design matrix $\rmX$ and column-wise link functions $g$, $\rmB$ determines the trial-level parameters $\Theta$, from which the observation model $\mathcal{M}$ generates the observables $\rmY$. The same generative process applies to every model in the class $\mathcal{C}$.}
    \label{fig:meta-simulator}
\end{figure}

Building on \autoref{eq:generative}, we construct a meta-simulator that expresses the parameters of a cognitive model as functions of predictor variables (see~\autoref{fig:meta-simulator}). Each model family $\mathcal{F}$ has $D_{\mathcal{F}}$ intrinsic parameters $\rvtheta_{\mathcal{F}}$; for example, the DDM may include $\vartheta$, $a$, $z$, $\tau$, $s_\vartheta$, and $s_\tau$. Let $\rmX \in \mathbb{R}^{N \times I}$ be a design matrix over $N$ trials with $I$ columns, $\rmB \in \mathbb{R}^{I \times D_{\mathcal{F}}}$ a coefficient matrix with entries $\beta_{ij}$, and $\rmM \in \{0,1\}^{I \times D_{\mathcal{F}}}$ a binary mask indicating which entries of $\rmB$ are active. Trial-level parameters and observations are then generated as
\begin{equation}
\label{eq:meta-simulator}
  \Theta = g\bigl(\rmX\,(\rmM \odot \rmB)\bigr),
  \qquad
  \rvy_n \sim p_{\mathcal{F}}(\rvy \mid \rvtheta_n),
  \quad n = 1, \dots, N,
\end{equation}
where row $\rvtheta_n$ of $\Theta \in \mathbb{R}^{N \times D_{\mathcal{F}}}$ holds the model parameters for trial $n$ and $\rvy_n$ is the corresponding row of $\rmY$. Thus, $\rmB$ determines how predictor variables affect each cognitive-model parameter, while $\rmM$ determines which of those effects are present in a particular simulated model. The link functions $g = (g_1, \dots, g_{D_{\mathcal{F}}})$ act column-wise and map each linear predictor to the valid domain of its parameter, using softplus transformations for positive parameters and scaled sigmoid transformations for bounded parameters such as $z$ (\autoref{tab:model-priors}). In the formulation of \autoref{eq:generative}, the pair $(\rmX, \rmM)$ encodes $(\mathcal{M}, \mathcal{D})$, and $\rmB$ determines $\rvtheta_{\mathcal{F}}$ for every trial, so $\rmB$ is the inference target.

Not all intrinsic parameters are treated identically. Core parameters, such as $\vartheta$, $a$, $z$, and $\tau$, are always estimated and may depend on their respective predictor variables, whereas optional parameters such as $s_\vartheta$ and $s_\tau$ may instead be fixed at zero. The contamination probability $\pi$ is treated separately as a global dataset-level parameter, estimated once per dataset and is never regressed on the design. When contamination is included, the observation model in \autoref{eq:meta-simulator} becomes
\begin{equation}
\label{eq:contamination}
  \rvy_n \sim
  (1-\pi)\,p_{\mathcal{F}}(\rvy\mid\rvtheta_n)
  +
  \pi\,\mathcal{U}(\mathcal{Y}_{\mathcal{F}}),
  \qquad
  \pi \sim \mathrm{Beta}(1,50),
\end{equation}
where $\mathcal{Y}_{\mathcal{F}}$ is the response domain of family $\mathcal{F}$. Contaminant observations are sampled uniformly over the response domain. Specifically, RT contaminants are drawn from $\mathcal{U}(0,20)$, and choice contaminants are drawn uniformly over the available alternatives, or uniformly over angles for the CDM.

For each simulated dataset, we sample both the design and the structure of the regression model. We draw the number of predictors as $R \sim \mathcal{U}\{0,\ldots, R_{\max}\}$, and dataset-specific inclusion probabilities control whether optional parameters are estimated and which model parameters receive main effects and interactions. These sampling procedures generate a wide range of model specifications, from intercept-only models to regressed designs. During validation, these random specifications are replaced by the fixed benchmark designs in \autoref{tab:benchmark-configs}.

The design matrix $\rmX$ contains an intercept followed by main effects and pairwise interactions. Predictors may be categorical or continuous, and the simulator deliberately varies their balance and dependence so that the network is exposed to both well-conditioned and strongly correlated experimental designs. All main-effect predictors of a dataset are generated jointly from a single latent Gaussian. For a design with $R_{\mathrm{cont}}$ continuous and $R_{\mathrm{cat}}$ categorical predictors ($R = R_{\mathrm{cont}} + R_{\mathrm{cat}}$), where categorical predictor $r$ has $C_r \sim \mathcal{U}\{2, \ldots, C_{\max}\}$ levels, each trial $n$ receives a latent vector
\begin{equation}
\label{eq:design-latent}
  \vh_n \sim \mathcal{N}(\vzero, \rmOmega),
  \qquad
  \rmOmega \sim \mathrm{LKJ}(\zeta),
  \qquad
  \zeta \sim \mathcal{U}(0.5, 5),
\end{equation}
of dimension $R_{\mathrm{cont}} + \sum_r C_r$, with one coordinate per continuous predictor and one score per level of each categorical predictor. Continuous predictors take their coordinates directly. Each categorical predictor is sampled through a softmax over its block of scores $\vh_{n,r} \in \mathbb{R}^{C_r}$,
\begin{equation}
\label{eq:design-categorical}
  c_{nr} \sim \mathrm{Categorical}\bigl(\mathrm{softmax}(\bm{\alpha}_r + \vh_{n,r})\bigr),
  \qquad
  \bm{\alpha}_r \sim \mathcal{N}(\vzero, \rmI_{C_r}),
\end{equation}
where the dataset-level intercepts $\bm{\alpha}_r$ set the level proportions, which range from nearly balanced when the intercepts are similar to strongly uneven when they are dispersed. Because the continuous coordinates and the categorical scores share one correlation matrix, a single LKJ draw induces both the correlations among continuous predictors and their association with the categorical factors. Small $\zeta$ yields strongly dependent designs, and large $\zeta$ yields nearly independent ones. Each categorical predictor receives a full block of $C_r$ scores with exchangeable intercept priors, so no level is privileged during generation. Levels that receive no trials are dropped, and the remaining levels are relabelled in random order before dummy coding against the first, which makes the reference level arbitrary. Pairwise interactions are added as products of the corresponding predictor columns.

Regression coefficients are sampled on the pre-link scale. Intercepts are drawn from the corresponding family-level parameter priors, and slopes for continuous predictors are drawn from $\mathcal{N}(0,1)$. For categorical predictors, we use a cell-means construction rather than assigning independent priors directly to reference-coded contrasts. Latent level means are drawn symmetrically and converted into contrasts relative to the reference level, with categorical interactions defined by the corresponding double differences. This construction ensures that the prior does not privilege the arbitrarily chosen reference level.

Finally, because model families may share some parameters, we embed all model families in a common parameter space of dimension $D_{\mathcal C}$, defined as the ordered union of the intrinsic parameters across the model class $\mathcal C$. Parameters absent from the sampled family are set to zero and marked inactive. The design matrix, coefficient matrix, and mask are likewise zero-padded to a common maximum of $I_{\max}$ design columns, so that $\rmX \in \mathbb{R}^{N \times I_{\max}}$ and $\rmB, \rmM \in \mathbb{R}^{I_{\max} \times D_{\mathcal{C}}}$. This common representation allows a single network to perform inference across datasets that differ in model family, experimental design, active parameters, and regression structure. The number of trials $N$ is shared within a training batch, whereas the remaining quantities are sampled independently for each dataset.

\subsection*{CogFormerV2 architecture}
\label{sec:backbone}

\begin{figure}[t]
    \centering
    \includegraphics[width=\linewidth]{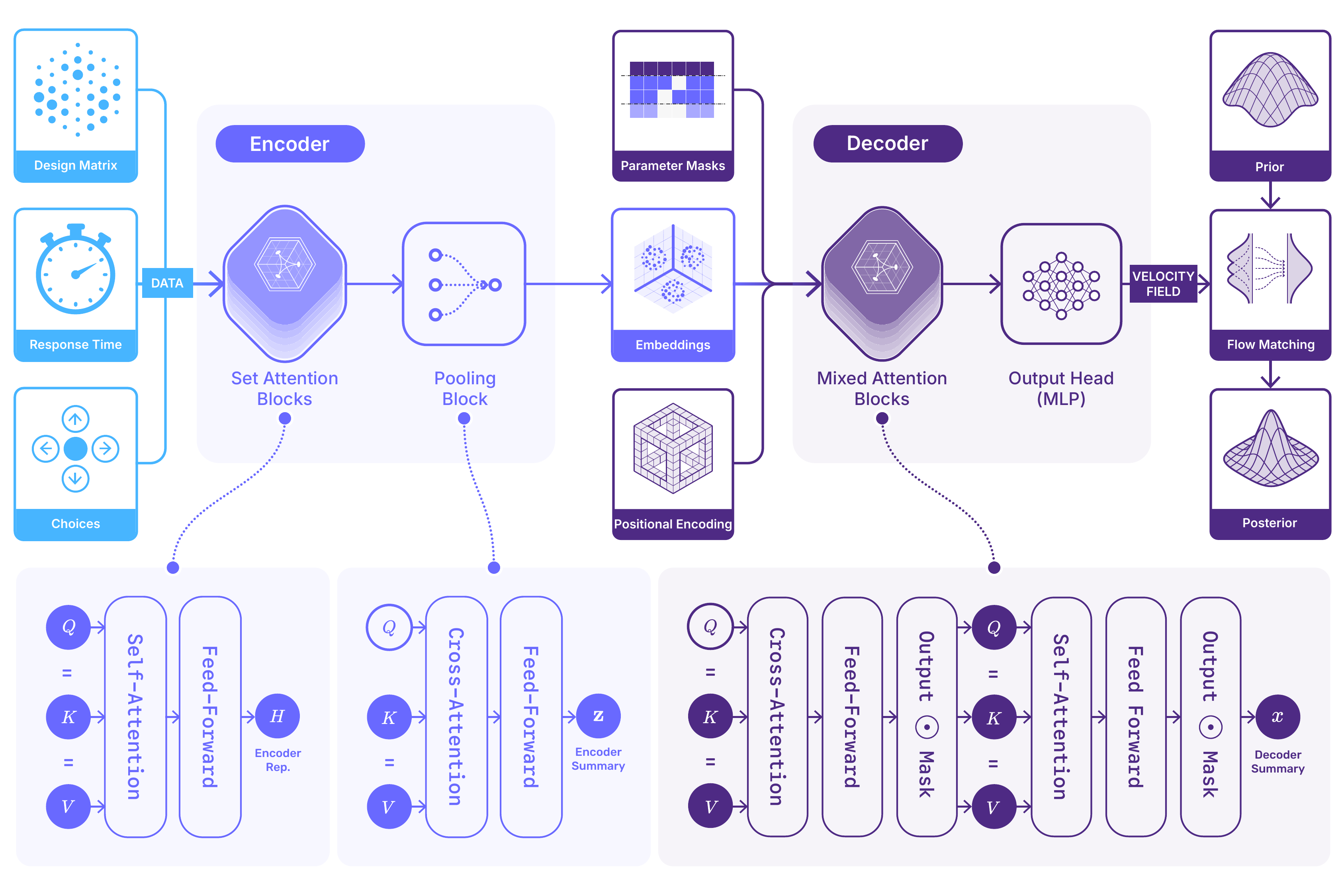}
    \caption{\textbf{CogFormer architecture}. CogFormer is an encoder-decoder architecture where the encoder is a set transformer that converts a design matrix and model observables $(\rmX, \rmY)$ into summary representations $\mathbf{S}$, and the decoder is a stack of mixed-attention layers that outputs a velocity field $\hat{\mathbf{u}}$. The decoder is trained with a conditional flow-matching objective, and integrating $\hat{\mathbf{u}}$ yields posterior samples.}
    \label{fig:transformer}
\end{figure}

Let $\mathcal{A} = (\rmY, \rmX, \rmM, \mathcal{F})$ collect all conditioning variables that define the amortization scope. Our goal is to approximate the full joint posterior $p(\rmB \mid \mathcal{A})$ without restrictive parametric or autoregressive assumptions. To this end, we represent the posterior implicitly through a conditional velocity field $\hat{u}(\rmZ_t; \mathcal{A}, t)$, realized by an encoder-decoder transformer (\autoref{fig:transformer}) and trained with conditional flow matching \citep{lipman2023flow, liu2023flow}. 

The encoder is an enhanced version of the Set Transformer \citep{lee2019set}. The trial-wise rows of the design matrix and the observed responses $(\rmX, \rmY)$ pass through a stack of set attention blocks (self-attention across trials) and are pooled into a sequence of $L$ learned summary vectors $\rmS = (\rvs_1, \dots, \rvs_L)$, with $\rvs_\ell \in \mathbb{R}^{d}$. Because it operates on sets of trials, the encoder is invariant to trial order and accommodates any number of trials and any experimental design without architectural changes.

The decoder maps these summaries to the velocity of the parameters $\rmB$. Its distinguishing feature is that it treats the coefficient matrix $\rmB \in \mathbb{R}^{I \times D_{\mathcal{F}}}$ not as a fixed-length vector but as a sequence of $I \times D_{\mathcal{F}}$ tokens, one per entry $\beta_{ij}$. This allows for the free variation of the number of predictors (rows) and intrinsic parameters (columns) across models. Each token carries sinusoidal embeddings of its predictor index $i$, its intrinsic-parameter index $j$, and its model-family index $f$, which retain the spatial structure of the GLM, as well as of the flow time $t$. The binary parameter mask $\rmM$ enters separately as a query mask that distinguishes active from fixed parameters.
The tokens then pass through a stack of mixed-attention layers that alternate between two operations. Cross-attention to the encoder summaries $\rmS$ brings in information from the data, and self-attention among the current parameter states $\rmZ_t$ and their embeddings models dependencies across the entire coefficient matrix. Within each block, FiLM layers \citep{perez2018film} modulate the residual stream by the time embedding, a form of conditioning found to improve conditional flow matching \citep{wildberger2023flow, arruda2025diffusion}.

The velocity field produced by the decoder defines a flow that transports a standard Gaussian latent $\rmZ_1 \sim \mathcal{N}(\mathbf{0}, \mathbf{I})$ into the posterior, $\rmZ_0 \equiv \rmB \sim p(\rmB \mid \mathcal{A})$, along the linear probability path $\rmZ_t = (1-t)\rmB + t\rmZ_1$ \citep{liu2023flow}. We use flow matching because recent SBI benchmarks identify it as a highly competitive free-form sampler \citep{arruda2025diffusion}, although the architecture is equally compatible with other free-form methods such as score-based diffusion. Along the linear path, the target velocity is simply $\rmZ_1 - \rmB$, and the network is trained to regress it by minimizing
\begin{equation}\label{eq:flow-matching-loss}
\mathcal{L}_{\text{FM}} =
\mathbb{E}_{p(t)\,p(\rmB, \mathcal{A})\,p(\rmZ_1)}
\left[ \frac{1}{\Vert \rmM \Vert_1} \left\Vert \rmM \odot \bigl(\hat{u}(\rmZ_t; \mathcal{A}, t) - (\rmZ_1 - \rmB)\bigr) \right\Vert_2^2 \right],
\end{equation}
where the expectation is taken over flow times $t$, draws from the meta-simulator, and the Gaussian latent. The mask $\rmM$ restricts the loss to active parameters, and normalizing by their number $\Vert \rmM \Vert_1$ weights models with many and few free parameters equally. Flow times are drawn as $t = 1 - u^{3/2}$ with $u \sim \mathcal{U}(0, 1)$, which places more training mass near the noise end of the path. At inference time, we draw samples from the approximate posterior $q(\rmB \mid \mathcal{A}, \hat{u})$ by integrating the learned velocity field from $t = 1$ to $t = 0$ with an off-the-shelf ODE solver. In our experiments, we use a fourth-order Runge--Kutta scheme with 50 steps ($\Delta t = 0.02$).

\subsection*{Experiment 1: CogFormerV2 Training and evaluation}
\label{ssec:exp1-protocol}

CogFormerV2 is trained using a single NVIDIA RTX 6000 Ada Generation GPU with 48 GB of GDDR6 memory on a workstation equipped with an AMD Ryzen Threadripper PRO 5965WX processor (24 cores, 48 threads) and 256 GiB of usable system RAM. It is trained on simulated data from the DDM, RDM, and CDM families using the settings in \autoref{app:hypers}. It is trained for 15,000 epochs, with 100 steps per epoch and 128 simulations (i.e., batch size) per step, totaling 192,000,000 simulations as the final training budget. Excluding intermediate evaluation time during training, the total training time is approximately 24.5 hours. 

We evaluate 1000 randomly sampled model configurations, using 200 datasets of 500 observations per configuration (200,000 datasets in total). For each sampled model configuration, we assess parameter recovery and simulation-based calibration (SBC) for all active inference targets, including the contamination probability $\pi$. All evaluation metrics are defined in \nameref{ssec:metrics}. For our static design optimization study, we use model configurations involving two interacting predictors that affect DDM drift rate $\vartheta$ and boundary separation $a$. Each predictor is continuous (C) or binary (B), yielding 4 possible predictor combinations: CC, CB, BC, and BB. We vary the number of observations from 200 to 1,000 and compare EIG and NRMSE. To quantify the meta-amortization gap, we also compare CogFormerV2 with 15 design-specific neural estimators on identical test datasets from the benchmark configurations (see \autoref{tab:benchmark-configs} for details about parameter-predictor mappings for the benchmark cases). 

\subsection*{Experiment 2: Empirical datasets}
\label{ssec:exp2-protocol}

We evaluate CogFormerV2 with three empirical datasets from three different experiments. First, we fit the DDM to data from a truth judgment task \cite{stump2026cognitive}, which includes 72 participants with 100--116 observations each. There are two interacting predictors: stimulus type (\textit{novel} vs. familiar) and factual truth (true vs. false). These predictors influence drift rate, boundary separation, initial bias, NDT, and its variability. Then, we fit the RDM with a three-alternative Stroop dataset from \cite{stevensonBayesianHierarchical2026}, which includes 149 participants with 171--192 observations each. We configure two interacting predictors: congruency (congruent vs. incongruent) and session (0 vs. 1). Responses are stimulus-coded, with the accumulator matching the ink color receiving the additional drift rate. Finally, we fit CDM with the continuous orientation-judgment dataset \cite{kvam2019geometric}, which includes 12 participants with 456--599 observations each. The predictors are speed conditions (speeded vs. unspeeded) and difficulty (easy, medium, or hard), which are dummy-coded with \textit{easy} as the reference level. Recorded angular deviations on $[-\pi/2, \pi/2]$ are doubled to match the model's response domain of $[-\pi, \pi]$. Detailed predictor-parameter mappings can be found in \autoref{tab:e2-conditions}.

For all empirical applications, we perform posterior predictive checks by generating replicated datasets using parameter draws from the posterior distribution. These replications preserve each participant’s number of observations, experimental conditions, and stimulus coding and incorporate the contamination process. Simulated observations are pooled across participants within each draw and compared with observed data by experimental condition. We examine RT distributions and the 0.1, 0.3, 0.5, 0.7, and 0.9 quantiles, summarizing predictions with medians and 95\% predictive intervals, alongside comparisons of observed and predicted response distributions.

\subsection*{Experiment 3: Zero-shot inference and fine-tuning procedures}
\label{ssec:exp3-protocol}

We evaluate the pre-trained CogFormerV2's adaptability to previously unseen model families in two stages. First, without fine-tuning, we examine parameter recovery and calibration with 200 simulated datasets for the LBA model. This is possible because the LBA shares the same intrinsic parameters as the RDM, which makes it possible to use the RDM's model embedding directly. The LBA threshold is parameterized so that $a$ represents the mean distance from the starting point to the threshold, matching the interpretation of $a$ in the RDM. 

After zero-shot inference, we continue training the pretrained CogFormerV2 on an expanded model class that includes the LBA for an additional 1,000 epochs, using an initial learning rate of $10^{-5}$ and the same AdamW optimizer and cosine decay schedule. This allows the LBA to have its own model embedding in CogFormer, initialized from the RDM's. The pretrained and fine-tuned networks are evaluated on the same benchmark datasets with two stimulus-coded alternatives. 

To visualize changes during LBA fine-tuning, we simulate DDM, RDM, CDM, aDDM, and LBA under baseline model configurations without predictors, drawing 200 parameter sets per family and generating four independent datasets of 500 trials per set. Identical datasets are passed through the original encoder and the encoder fine-tuned in Experiment~3. Summaries are averaged across repetitions, and each summary dimension is standardized using its pooled mean and standard deviation across all families and both encoders. We then fit a single two-dimensional metric multidimensional scaling (MDS) solution to the Euclidean distances between these standardized summaries, using four initializations. The resulting coordinates are displayed separately by encoder in a shared coordinate system. 

Afterwards, we apply the fine-tuned network to the lexical-decision dataset from \cite{wagenmakers2008diffusion}, comprising 17 participants with 792–1,920 observations each. Models are fitted separately for each participant, with speed versus accuracy instruction (effect-coded as \(\pm 0.5\)) predicting baseline drift rate, drift advantage, threshold distance, and NDT. Drift variability, NDT variability, and contamination probability are estimated as intercept-only parameters, with the starting-point range fixed at \(A=0.5\). Responses are stimulus-coded as word versus non-word, with the stimulus-appropriate accumulator receiving the additional drift rate. 

Following the procedure described in \autoref{ssec:exp2-protocol}, we perform posterior predictive checks using 500 posterior draws per participant, grouping predictions by instruction condition, word-frequency band, and lexicality. Frequency and lexicality are not included as regression predictors; lexicality is entered through stimulus coding.

\subsection*{Experiment 4: Encoder representation transfer procedures}
\label{ssec:exp4-protocol}

We choose the aDDM to study CogFormerV2's encoder representation, with five parameters: $(\vartheta_{\text{diff}}, \omega, a, \tau, s_\tau)$. Baseline drift rate $\vartheta$ and its variability $s_\vartheta$ are fixed. We compare CogFormerV2's frozen encoder with a 128-dimensional summary. The baseline uses a 16-dimensional Set Transformer summary trained jointly with a flow matching inference network \cite{lipman2023flow} implemented in BayesFlow \cite{kuhmichel2026bayesflow}. For validation, both neural estimators receive the same trial-level inputs and are trained for 500 epochs, each with 100 batches of 64 simulated datasets of 500 observations. We assess parameter recovery on 500 validation datasets per estimator, drawing 500 posterior samples per dataset.

To compare the encoder representations (\nameref{par:rsa}), we draw 200 aDDM parameter sets and simulate four independent datasets of 500 observations per set, passing identical datasets through both encoders. We compute linear centered kernel alignment (CKA) \cite{kornblith2019similarity} on summaries averaged across repetitions for each parameter set. CKA compares the representations’ centered linear Gram matrices, allowing comparison across different summary dimensions. For both measures, we construct a dimensionality-matched reference using 30 random Gaussian linear projections of the frozen summaries from 128 to 16 dimensions. This reference captures similarity retained after dimensionality reduction, rather than an absence of shared structure. As a complementary measure, we also compute representational dissimilarity matrices \cite{kriegeskorte2008representational} using crossnobis distances, cross-validate across the four repetitions, and quantify their correspondence using Pearson correlation between off-diagonal entries. 

\section*{Code availability}
\label{sec:code}

The code containing the implementation of this work can be found as a GitHub repository: \href{https://github.com/bayesflow-org/MACoM}{https://github.com/bayesflow-org/MACoM}.

\backmatter

\bmhead{Acknowledgments}

STR and JMH were funded by the National Science Foundation under Grant No.~2448380. We would like to thank Niels L. Bracher, Paul-Christian Bürkner, and Simon Kucharsky for their helpful feedback.

\section*{Declarations}

No conflict of interest.



\bibliography{sn-bibliography}

\newpage

\begin{appendices}

{\renewcommand{\contentsname}{Supplemental Materials}
\setcounter{tocdepth}{2}
\tableofcontents}
\newpage

\section{Notation}
\label{app:notation}

{\small
\renewcommand{\arraystretch}{1.1}
\begin{longtable}{@{}p{2.4cm} p{3.6cm} >{\raggedright\arraybackslash}p{\dimexpr\linewidth-6cm-4\tabcolsep\relax}@{}}
\caption{Table of notation.}\label{tab:notation}\\
\toprule
\textbf{Category} & \textbf{Symbol} & \textbf{Meaning} \\
\midrule
\endfirsthead
\multicolumn{3}{@{}l}{\textit{\tablename~\thetable{} continued}}\\
\toprule
\textbf{Category} & \textbf{Symbol} & \textbf{Meaning} \\
\midrule
\endhead
\midrule
\multicolumn{3}{r@{}}{\textit{continued on next page}}\\
\endfoot
\bottomrule
\endlastfoot
\multirow{4}{=}{\textit{Amortization scopes}}
 & $\mathcal{C}$ & Model class (e.g.\ evidence accumulation models) \\
 & $\mathcal{F}$ & Model family (shared computational structure) \\
 & $\mathcal{M}$ & Model species (fully specified observation model) \\
 & $\mathcal{D}$ & Design configuration over the intrinsic parameters \\
\midrule
\multirow{8}{=}{\textit{Generative objects}}
 & $\rvtheta_{\mathcal{F}}$ & Intrinsic parameters of family $\mathcal{F}$ \\
 & $\rmX$ & Design matrix, $\rmX \in \mathbb{R}^{N \times I}$ \\
 & $\rmB$ & Coefficient matrix, $\rmB \in \mathbb{R}^{I \times D_{\mathcal{F}}}$ (inference target); entries $\beta_{ij}$ \\
 & $\rmM$ & Binary parameter mask, $\rmM \in \{0,1\}^{I \times D_{\mathcal{F}}}$; entries $M_{ij}$ \\
 & $\Theta$ & Trial-level parameters, $\Theta = g(\rmX(\rmM \odot \rmB)) \in \mathbb{R}^{N \times D_{\mathcal{F}}}$; rows $\rvtheta_n$ \\
 & $\rmY$ & Model observables, with rows $\rvy_n$ \\
 & $\pi$ & Contamination probability (global, not regressed) \\
 & $\mathcal{Y}_{\mathcal{F}}$ & Response domain of family $\mathcal{F}$ \\
\midrule
\multirow{4}{=}{\textit{Design generation}}
 & $R_{\mathrm{cont}},\, R_{\mathrm{cat}}$ & Numbers of continuous and categorical predictors \\
 & $C_r,\, C_{\max}$ & Number of levels of categorical predictor $r$; its maximum \\
 & $\vh_n,\, \rmOmega$ & Latent design vector of trial $n$; its LKJ-distributed correlation matrix \\
 & $\bm{\alpha}_r$ & Level intercepts of categorical predictor $r$ \\
\midrule
\multirow{8}{=}{\textit{Dimensions and indices}}
 & $N$ & Number of trials per dataset; index $n$ \\
 & $R$ & Number of predictors, $R = R_{\mathrm{cont}} + R_{\mathrm{cat}}$ \\
 & $I$ & Number of design-matrix columns (intercept, main effects, interactions) \\
 & $I_{\max}$ & Maximum design capacity (zero-padding target) \\
 & $D_{\mathcal{F}}$ & Number of intrinsic parameters of family $\mathcal{F}$ \\
 & $D_{\mathcal{C}}$ & Size of the shared parameter space of class $\mathcal{C}$ \\
 & $i,\, j$ & Row (design column) and column (parameter) indices of $\rmB, \rmM$ \\
 & $f$ & Integer index of family $\mathcal{F}$ (model-family embedding) \\
\midrule
\multirow{1}{=}{\textit{Link functions}}
 & $g = (g_1, \dots, g_{D_{\mathcal{F}}})$ & Per-parameter link functions, applied column-wise \\
\midrule
\multirow{7}{=}{\textit{Flow matching and inference}}
 & $\mathcal{A}$ & Conditioning set, $\mathcal{A} = (\rmY, \rmX, \rmM, \mathcal{F})$ \\
 & $t$ & Flow time, $t \in [0, 1]$ \\
 & $\rmZ_t$ & Flow state, $\rmZ_t = (1-t)\rmB + t\rmZ_1$ \\
 & $\rmZ_0,\, \rmZ_1$ & Target endpoint $\rmZ_0 \equiv \rmB$; Gaussian latent $\rmZ_1 \sim \mathcal{N}(\vzero, \rmI)$ \\
 & $\hat{u}$ & Learned conditional velocity field, $\hat{u}(\rmZ_t; \mathcal{A}, t)$ \\
 & $\mathcal{L}_{\mathrm{FM}}$ & Masked conditional flow matching loss \\
 & $q(\rmB \mid \mathcal{A}, \hat{u})$ & Approximate posterior induced by $\hat{u}$ \\
\midrule
\multirow{3}{=}{\textit{Architecture}}
 & $\rmS$ & Encoder summaries, $\rmS = (\rvs_1, \dots, \rvs_L)$ \\
 & $\rvs_\ell$ & Summary vector, $\rvs_\ell \in \mathbb{R}^{d}$ \\
 & $L,\, d$ & Number of summary tokens (encoder seeds); summary dimension \\
\midrule
\multirow{9}{=}{\textit{Evaluation}}
 & $S$ & Number of simulated test datasets; index $s$ \\
 & $K$ & Number of posterior draws per dataset; index $k$ \\
 & $\beta^{(s)}$ & Ground truth of an active coefficient in dataset $s$ \\
 & $\tilde{\beta}^{(s)}_k,\, \bar{\beta}^{(s)}$ & Posterior draws of that coefficient; their mean \\
 & $\Delta_\beta$ & NRMSE normalizer (range, or standard deviation, of the ground truths) \\
 & $\mathrm{rank}^{(s)},\, \hat{F}$ & SBC rank statistic; empirical CDF of fractional ranks \\
 & $\gamma,\, \hat{c}(\gamma)$ & Nominal credible level; its empirical coverage \\
 & $\Gamma$ & Set of 20 nominal levels used for ECE \\
 & $\boldsymbol{\Sigma}_{\mathrm{prior}},\, \boldsymbol{\Sigma}_{\mathrm{post}}$ & Prior and posterior covariances used for EIG \\
\end{longtable}
}

\newpage
\section{Specifications of evidence accumulation models}
\label{app:decision-models}

\subsection{Cognitive model families}
\paragraph{Diffusion decision model (DDM)}

The DDM describes binary choices as a process of noisy evidence accumulation toward one of two decision boundaries.
Let $x(t)$ denote the accumulated evidence at time $t$.
The evidence accumulation process follows the stochastic differential equation
\begin{equation}
dx(t)=\vartheta\,dt+s\,dW_t,
\end{equation}
where $\vartheta$ denotes the drift rate, reflecting the average rate of evidence accumulation, $s$ is the diffusion coefficient (fixed at 1 for scaling constraints), and $W_t$ denotes a Wiener process.
Across trials, drift rates are normally distributed around $\vartheta$ with standard deviation $s_\vartheta$.
Accumulation starts at an initial state $x(0)=az$, where $z\in[-1,1]$.

A decision is triggered when the accumulation process reaches one of the boundaries.
Let $T_{\text{dec}}$ denote the corresponding first-passage time.
The observed decision $D$ is determined by the boundary reached,
\begin{equation}
D =
\begin{cases}
1 & \text{if } x(T_{\text{dec}})=a,\\
0 & \text{if } x(T_{\text{dec}})=-a.
\end{cases}
\end{equation}

\paragraph{Attentional drift-diffusion model (aDDM)}

The aDDM extends the DDM by making the drift rate depend on which alternative is fixated.
Let $\vartheta_{0i}$ and $\vartheta_{1i}$ denote the evidence accumulation rates for responses $0$ and $1$ on trial $i$, before attentional weighting.
Their means are parameterized as
\begin{equation}
\vartheta_0=\vartheta,
\qquad
\vartheta_1=\vartheta+\vartheta_{\text{diff}},
\end{equation}
where $\vartheta_{\text{diff}}$ captures the difference between the mean rates. 

The evidence accumulates according to
\begin{equation}
dx_i(t)=\mu_i(t)\,dt+s\,dW_{i,t},
\end{equation}
where the drift is the weighted rate for response $0$ minus that for response $1$,
\begin{equation}
\mu_i(t)=
\begin{cases}
\vartheta_{0i}-\omega\vartheta_{1i}
& \text{if alternative }0\text{ is fixated},\\
\omega\vartheta_{0i}-\vartheta_{1i}
& \text{if alternative }1\text{ is fixated}.
\end{cases}
\end{equation}
The fixated alternative receives full weight, while the other receives weight $\omega\in(0,1)$. Smaller values of $\omega$ imply stronger attentional discounting, whereas values approaching one reduce the influence of fixation.

Fixations alternate between the two alternatives, with either alternative equally likely to be fixated first. Fixation durations are independently drawn from an exponential distribution with a mean of $0.3$ seconds. We fixed $\vartheta$ and $s_\vartheta$ at $\operatorname{softplus}(0.6)$ and $\operatorname{softplus}(-1.0)$, respectively.

\paragraph{Circular diffusion model (CDM)}

The CDM extends the diffusion framework to continuous response spaces.
Evidence accumulation occurs in two dimensions,
\begin{equation}
d\rmX(t)=
\begin{pmatrix}
dx_1(t)\\
dx_2(t)
\end{pmatrix}
=
\boldsymbol{\mu}\,dt+s\,d\mathbf{W}_t,
\end{equation}
where $\boldsymbol{\mu}$ is a two-dimensional drift vector and $\mathbf{W}_t$ is a two-dimensional Wiener process.

We express the drift in polar coordinates using a drift magnitude $\vartheta$ and a drift angle $\vartheta_\theta$.
The Cartesian drift components are obtained via
\begin{equation}
\vartheta_x=\vartheta\cos(\vartheta_\theta),
\qquad
\vartheta_y=\vartheta\sin(\vartheta_\theta).
\end{equation}
Both drift components are subject to across-trial noise with standard deviation $s_\vartheta$.

The accumulation starts at the origin and evolves within a circular absorbing boundary with radius $a$.
A response is generated when the diffusion trajectory first reaches the boundary.
The angular position of the boundary crossing determines the response direction, while the hitting time determines the decision time.

Here we express responses as angular deviations from the correct response. Unbiased responding corresponds to a circular mean error of zero. The drift angle $\vartheta_\theta$ reflects systematic error bias. The drift magnitude $\vartheta$ controls the mean rate of evidence accumulation, similar to the drift rate in the DDM.

\paragraph{Racing diffusion model (RDM)}

The RDM posits multiple independent noisy accumulators that race toward a decision threshold $a$. For $K$ alternatives, each option $k$ is associated with a diffusion process
\begin{equation}
dx_k(t)=\vartheta_k\,dt+s\,dW_{k,t},
\end{equation}
where $\vartheta_k$ denotes the drift rate of accumulator $k$ and $W_{k,t}$ are independent Wiener processes. During training, the number of alternatives varies over $K\in\{2,3,4\}$.

The response is determined by the first accumulator to reach the threshold.
Let $T_k$ denote the first-passage time of accumulator $k$.
The chosen alternative and decision time are
\begin{equation}
k^*=\arg\min_k T_k,
\qquad
T_{\text{dec}}=\min_k T_k.
\end{equation}

Drift rates are parameterized using stimulus coding.
On each trial, let $c$ denote the accumulator matching the correct response.
The mean drift rates are defined as
\begin{equation}
\vartheta_k =
\begin{cases}
\vartheta+\vartheta_{\text{diff}} & \text{if } k=c,\\
\vartheta & \text{otherwise},
\end{cases}
\label{eq:stim_coding}
\end{equation}
where $\vartheta$ denotes the base drift rate shared by all accumulators and $\vartheta_{\text{diff}}$ denotes the additional drift toward the correct response.
Larger values of $\vartheta_{\text{diff}}$ therefore increase the drift advantage of the correct accumulator over the competing accumulators.
Across trials, drift rates are sampled from a normal distribution with standard deviation $s_\vartheta$.

\paragraph{Linear ballistic accumulator (LBA)}

Like the RDM, the LBA describes choices as a race between accumulators.
Evidence evolves linearly within each trial, with starting points and drift rates varying across trials.
For accumulator $k$ on trial $i$, the accumulated evidence is given by
\begin{equation}
x_{ki}(t)=z_{ki}+\vartheta_{ki}t,
\end{equation}
where the starting point is drawn from $z_{ki}\sim U(0,A)$ and $\vartheta_{ki}$ denotes the trial-specific drift rate. Across trials, drift rates are sampled independently from a normal distribution with standard deviation $s_\vartheta$. Unlike the other models, there is no within-trial variation of drift rates.

In the present work, we used two alternatives and the same stimulus-coding scheme as the RDM (Equation \ref{eq:stim_coding}). We set $A=\min(0.5,2a)$ and the common decision threshold to $a+A/2$, so that $a$ represents the mean distance from the starting point to the threshold.

\paragraph{Shared model assumptions}

Across all five models, the observed RT is the sum of a decision time and a non-decision component,
\begin{equation}
RT_i=T_{\text{dec},i}+\tau_i,
\qquad
\tau_i=\tau+U(0,s_\tau),
\end{equation}
where $\tau$ denotes the minimum non-decision time and $s_\tau$ determines its range of variability.
Non-decision time captures processes such as stimulus encoding and motor execution.

\subsection{Priors and link functions}

A table of prior and link function specifications for each model family can be found in \autoref{tab:model-priors}.

\begin{table}[htbp]
\centering
\footnotesize
\renewcommand{\arraystretch}{1.2}
\setlength{\tabcolsep}{2pt}
\caption{Priors and link functions. $^{*}$Identity link for the aDDM.}
\label{tab:model-priors}
\begin{tabular*}{\textwidth}{@{\extracolsep{\fill}}lccccccl@{}}
\toprule
Parameter & DDM & RDM & CDM & LBA & aDDM & & Link \\
\midrule
$\vartheta$             & $\mathcal{N}(1.0,0.8)$   & $\mathcal{N}(0.6,0.5)$   & $\mathcal{N}(1.0,0.8)$   & $\mathcal{N}(0.6,0.5)$   & --                        & & Softplus \\
$a$                     & $\mathcal{N}(0.5,0.5)$   & $\mathcal{N}(0.5,0.5)$   & $\mathcal{N}(0.5,0.5)$   & $\mathcal{N}(0.5,0.5)$   & $\mathcal{N}(0.25,0.5)$  & & Softplus \\
$z$                     & $\mathcal{N}(0.0,0.8)$   & --                        & --                        & --                        & --                        & & Scaled sigmoid \\
$\tau$                  & $\mathcal{N}(-1.2,0.5)$  & $\mathcal{N}(-1.2,0.5)$  & $\mathcal{N}(-1.2,0.5)$  & $\mathcal{N}(-1.2,0.5)$  & $\mathcal{N}(-1.2,0.5)$  & & Softplus \\
$s_{\vartheta}$         & $\mathcal{N}(-1.0,0.6)$  & $\mathcal{N}(-1.0,0.6)$  & $\mathcal{N}(-1.0,0.6)$  & $\mathcal{N}(-1.0,0.6)$  & --                        & & Softplus \\
$s_{\tau}$              & $\mathcal{N}(-1.5,0.7)$  & $\mathcal{N}(-1.5,0.7)$  & $\mathcal{N}(-1.5,0.7)$  & $\mathcal{N}(-1.5,0.7)$  & $\mathcal{N}(-1.5,0.7)$  & & Softplus \\
$\vartheta_{\mathrm{diff}}$ & --                    & $\mathcal{N}(0.6,0.5)$   & --                        & $\mathcal{N}(0.6,0.5)$   & $\mathcal{N}(0.0,0.5)$   & & Softplus$^{*}$ \\
$\vartheta_{\theta}$    & --                        & --                        & $\mathcal{N}(0.0,0.5)$   & --                        & --                        & & Scaled sigmoid \\
$\omega$                & --                        & --                        & --                        & --                        & $\mathcal{N}(0.0,1.2)$   & & Sigmoid \\
$\pi$                   & $\mathrm{Beta}(1,50)$    & $\mathrm{Beta}(1,50)$    & $\mathrm{Beta}(1,50)$    & $\mathrm{Beta}(1,50)$    & --                        & & Identity \\
\bottomrule
\end{tabular*}
\end{table}

\subsection{Benchmarked model configuration}

A table of benchmarked model configurations for validating CogFormerV2 can be found in \autoref{tab:benchmark-configs}.

\begin{table}[ht]
\centering
\small
\caption{Benchmarked model configurations. Each cell lists the parameters assigned to the corresponding effect type. A dash ($-$) indicates no parameters are assigned. The LBA uses the same parameter names and configurations as the RDM.}
\label{tab:benchmark-configs}
\begin{tabular}{lllll}
\toprule
\textbf{Configuration} & \makecell[l]{\textbf{Intercept}\\($\mathbf{1}$)} & \makecell[l]{\textbf{Main Effect}\\($u_1$)} & \makecell[l]{\textbf{Main Effect}\\($u_2$)} & \makecell[l]{\textbf{Interaction}\\($u_1{\times}u_2$)} \\
\midrule
\multicolumn{5}{l}{\textbf{DDM}} \\
Intercept Only & $\vartheta, a, z, \tau, s_{\vartheta}, s_\tau$ & $-$ & $-$ & $-$ \\
Fixed Variability & $\vartheta, a, z, \tau$ & $-$ & $-$ & $-$ \\
Regressed & $\vartheta, a, z, \tau, s_{\vartheta}, s_\tau$ & $\vartheta, a, z$ & $\vartheta, a, z$ & $-$ \\
Fixed + Regressed & $\vartheta, a, z, \tau$ & $\vartheta, a, z$ & $\vartheta, a, z$ & $-$ \\
With Interaction & $\vartheta, a, z, \tau, s_{\vartheta}, s_\tau$ & $\vartheta, a, z, \tau, s_{\vartheta}$ & $\vartheta, a, z, \tau$ & $\vartheta, a, z$ \\
\midrule
\multicolumn{5}{l}{\textbf{RDM}} \\
Intercept Only & $\vartheta, \vartheta_\Delta, a, \tau, s_{\vartheta}, s_\tau$ & $-$ & $-$ & $-$ \\
Fixed Variability & $\vartheta, \vartheta_\Delta, a, \tau$ & $-$ & $-$ & $-$ \\
Regressed & $\vartheta, \vartheta_\Delta, a, \tau, s_{\vartheta}, s_\tau$ & $\vartheta_\Delta, a$ & $\vartheta_\Delta, a$ & $-$ \\
Fixed + Regressed & $\vartheta, \vartheta_\Delta, a, \tau$ & $\vartheta_\Delta, a$ & $\vartheta_\Delta, a$ & $-$ \\
With Interaction & $\vartheta, \vartheta_\Delta, a, \tau, s_{\vartheta}, s_\tau$ & $\vartheta, \vartheta_\Delta, a, \tau, s_{\vartheta}$ & $\vartheta, \vartheta_\Delta, a, \tau$ & $\vartheta, \vartheta_\Delta, a$ \\
\midrule
\multicolumn{5}{l}{\textbf{CDM}} \\
Intercept Only & $\vartheta, \vartheta_\theta, a, \tau, s_{\vartheta}, s_\tau$ & $-$ & $-$ & $-$ \\
Fixed Variability & $\vartheta, \vartheta_\theta, a, \tau$ & $-$ & $-$ & $-$ \\
Regressed & $\vartheta, \vartheta_\theta, a, \tau, s_{\vartheta}, s_\tau$ & $\vartheta, a$ & $\vartheta, a$ & $-$ \\
Fixed + Regressed & $\vartheta, \vartheta_\theta, a, \tau$ & $\vartheta, a$ & $\vartheta, a$ & $-$ \\
With Interaction & $\vartheta, \vartheta_\theta, a, \tau, s_{\vartheta}, s_\tau$ & $\vartheta, \vartheta_\theta, a, \tau, s_{\vartheta}$ & $\vartheta, \vartheta_\theta, a, \tau$ & $\vartheta, \vartheta_\theta, a$ \\
\midrule
\multicolumn{5}{l}{\textbf{LBA}: same configurations as RDM} \\
\bottomrule
\end{tabular}
\end{table}

\section{Meta-simulator settings}
\label{app:meta-sim-settings}

\autoref{tab:meta-sim-hyper} lists the distributions and constants of the meta-simulator used to train the model-class network. Symbols follow the main text.

\begin{table}[ht]
\centering
\caption{Meta-simulator settings used for training.}
\label{tab:meta-sim-hyper}
\begin{tabular}{lll}
  \toprule
  Quantity & Symbol & Value \\
  \midrule
  Model families               & $\mathcal{C}$          & DDM, RDM, CDM (uniform) \\
  Shared parameter space       & $D_{\mathcal{C}}$      & $9$ ($v, a, z, \tau, s_v, s_\tau, \pi, v_{\mathrm{diff}}, v_\theta$) \\
  Trials per dataset           & $N$                    & $\mathcal{U}\{200, \dots, 1000\}$ \\
  predictors                   & $R$                    & $\mathcal{U}\{0, \dots, 4\}$ \\
  Levels per categorical predictor & $k_r$              & $\mathcal{U}\{2, 3, 4\}$ \\
  Design capacity              & $I_{\max}$             & $67$ \\
  Inclusion rates              & $p_0, p_1$             & $\mathrm{Beta}(2, 2)$ \\
  Contamination                & $\pi$                  & $\mathrm{Beta}(1, 50)$ \\
  \midrule
  Categorical probability      & $q_{\mathrm{cat}}$     & $0.5$ \\
  Balanced-level probability   & $q_{\mathrm{bal}}$     & $0.3$ \\
  Level concentration          & $\alpha$               & $\mathrm{LogUniform}(0.5, 50)$ \\
  Categorical association      & $\lambda$              & $0$ w.p.\ $0.5$, else $\mathcal{U}(0, 1)$ \\
  Coupling concentration       & $\kappa$               & $\mathrm{LogUniform}(0.3, 10)$ \\
  Categorical--continuous association & $\eta^2$        & $\min\{\mathrm{Beta}(1.2, 3),\, 0.85\}$ \\
  LKJ shape                    & $\zeta$                & $\mathcal{U}(0.5, 5)$ \\
  \bottomrule
\end{tabular} 
\end{table}

\section{CogFormer hyperparameters}
\label{app:hypers}

CogFormerV2's encoder has four self-attention layers with four heads and a width of 128. It pools the trial-level tokens into 32 learned seed tokens of the same width. Each trial is embedded together with its design row, padded to at most four predictors and four categories per predictor. The encoder is also conditioned on the coefficient structure through a two-layer MLP embedding. The flow-matching decoder has eight attention layers with eight heads and a projection dimension of 256 (32 per head). It uses 32-dimensional sinusoidal time and positional embeddings and an 8-dimensional learned embedding of the model family. The network has 7.6 million parameters. It was trained with AdamW at a learning rate of $10^{-4}$, annealed with a cosine schedule. Training ran for 15,000 epochs of 100 steps with batches of 128 simulated data sets (192 million in total). Each data set had between 200 and 1,000 trials. Dropout and layer dropout were both 0.05, and flow-matching times were drawn as $t = u^{1.5}$ with $u \sim \mathcal{U}(0,1)$. Posterior samples were obtained by integrating the learned flow with a fourth-order ge–Kutta solver over 50 steps. For the LBA, the network was fine-tuned for a further 1,000 epochs at a learning rate of $10^{-5}$.

\pagebreak
\section*{Evaluation metrics}
\label{ssec:metrics}

All simulation-based evaluations follow a common protocol. For a fixed design, we draw $S$ ground-truth coefficient matrices $\rmB^{(s)}$, $s = 1, \dots, S$, from the meta-simulator prior, simulate one dataset from each, and draw $K$ posterior samples
\begin{equation}
\label{eq:eval-draws}
  \tilde{\rmB}^{(s)}_k \sim q\bigl(\rmB \mid \mathcal{A}^{(s)}, \hat{u}\bigr),
  \qquad k = 1, \dots, K.
\end{equation}
Unless stated otherwise, $S = 200$ datasets of $N = 500$ trials and $K = 200$. Every metric is computed separately for each active coefficient $\beta = \beta_{ij}$ with $M_{ij} = 1$, on the unconstrained (link) scale on which the network is trained. For a given coefficient, we write $\beta^{(s)}$ for its ground truth in dataset $s$, $\tilde{\beta}^{(s)}_k$ for its posterior draws, and $\bar{\beta}^{(s)} = K^{-1} \sum_k \tilde{\beta}^{(s)}_k$ for its posterior mean. Values reported for a whole design average the per-coefficient metrics over its active coefficients. The contamination probability $\pi$ is estimated jointly with $\rmB$ but scored separately.

\paragraph{Parameter recovery and accuracy}
Recovery plots show the posterior means $\bar{\beta}^{(s)}$ against the ground truths $\beta^{(s)}$, together with their Pearson correlation $r$ across datasets. As a scalar summary of accuracy, we use the normalized root mean squared error (NRMSE) of the posterior draws. For each draw index $k$,
\begin{equation}
\label{eq:nrmse}
  \mathrm{NRMSE}_k = \frac{1}{\Delta_\beta}
  \sqrt{\frac{1}{S}\sum_{s=1}^{S} \bigl(\tilde{\beta}^{(s)}_k - \beta^{(s)}\bigr)^2},
  \qquad
  \Delta_\beta = \max_s \beta^{(s)} - \min_s \beta^{(s)}.
\end{equation}
NRMSE penalizes both bias and posterior spread and, through $\Delta_\beta$, is comparable across coefficients with different prior scales (lower is better). For the design comparison in Experiment~1, coefficients of binary and continuous predictors have different prior widths, so there $\Delta_\beta$ is the standard deviation of the ground truths instead of their range.

\paragraph{Posterior contraction}
Contraction measures how much the data narrow the posterior relative to the prior. For each dataset $s$,
\begin{equation}
\label{eq:contraction}
  \mathrm{PC}^{(s)} = 1 - \frac{\mathrm{Var}_k\bigl[\tilde{\beta}^{(s)}_k\bigr]}{\mathrm{Var}_s\bigl[\beta^{(s)}\bigr]},
\end{equation}
clipped to $[0, 1]$, where the variance of the ground truths across datasets estimates the prior variance. Values near 1 indicate a parameter that is well identified by the data; values near 0 indicate a posterior that remains close to the prior. Contraction is informative only alongside calibration, since an overconfident posterior also contracts strongly.

\paragraph{Calibration}
A calibrated approximator produces credible intervals whose empirical coverage matches their nominal level over the prior predictive distribution. We assess this in two ways. First, following simulation-based calibration \citep[SBC;][]{talts2018validating}, we compute the rank of each ground truth among its posterior draws and the empirical CDF of the fractional ranks,
\begin{equation}
\label{eq:sbc}
  \mathrm{rank}^{(s)} = \sum_{k=1}^{K} \mathbf{1}\bigl[\tilde{\beta}^{(s)}_k < \beta^{(s)}\bigr],
  \qquad
  \hat{F}(u) = \frac{1}{S} \sum_{s=1}^{S} \mathbf{1}\Bigl[\frac{\mathrm{rank}^{(s)}}{K} \leq u\Bigr],
  \quad u \in [0, 1].
\end{equation}
Under exact calibration, $\mathrm{rank}^{(s)}$ is uniform on $\{0, \dots, K\}$ and $\hat{F}(u) \approx u$. We display $\hat{F}(u) - u$ together with 95\% confidence bands for uniformity \citep{sailynoja2022graphical}. Excursions above or below the band indicate systematic under- or overestimation, and U-shaped or inverted-U-shaped deviations indicate posteriors that are too narrow or too wide. Second, as a scalar summary, we compute the empirical calibration error (ECE) over a set $\Gamma$ of 20 nominal levels equally spaced on $[0.005, 0.995]$. Let $\mathrm{CI}^{(s)}_{\gamma}$ be the central $\gamma$ credible interval of dataset $s$, bounded by the $(1 - \gamma)/2$ and $(1 + \gamma)/2$ quantiles of its posterior draws. For each level $\gamma \in \Gamma$,
\begin{equation}
\label{eq:ece}
  \mathrm{ECE}(\gamma) = \bigl| \hat{c}(\gamma) - \gamma \bigr|,
  \qquad
  \hat{c}(\gamma) = \frac{1}{S} \sum_{s=1}^{S} \mathbf{1}\bigl[\beta^{(s)} \in \mathrm{CI}^{(s)}_{\gamma}\bigr],
\end{equation}
where $\hat{c}(\gamma)$ is the empirical coverage at level $\gamma$ (lower ECE is better; 0 indicates perfect coverage at every level).

\paragraph{Aggregation of scalar metrics}
Eqs.~\eqref{eq:nrmse}, \eqref{eq:contraction}, and \eqref{eq:ece} yield one value per posterior draw, per dataset, and per nominal level, respectively. Following the BayesFlow implementation \citep{kuhmichel2026bayesflow}, we summarize each by its median over these units, which is robust to the occasional outlying draw or dataset; the NRMSE, contraction, and ECE reported throughout are these medians.

\paragraph{Posterior discrepancy between approximators}
To quantify the meta-amortization gap, we compare CogFormerV2's posteriors with those of single-model baselines on identical datasets using classifier two-sample tests \citep[C2ST;][]{lopez2016revisiting}. A classifier is trained to tell the draws of the two approximators apart, and its 5-fold cross-validated accuracy is reported: 0.5 means the two sets of draws are indistinguishable, and 1.0 means they are perfectly separable. The joint test over all active coefficients uses a multilayer perceptron with two hidden layers of width $2^{\lceil \log_2 (10\Vert \rmM \Vert_1) \rceil}$ for $\Vert \rmM \Vert_1$ active coefficients, trained with early stopping on standardized inputs. The per-coefficient test uses quadratic discriminant analysis, which is more stable than the MLP with the few draws available per dataset.

The level at which draws are compared matters. Pooling draws over all datasets compares the two data-averaged posteriors. For any calibrated approximator, these approach the prior, so the pooled C2ST stays near chance even when the two approximators disagree on each dataset. Our primary measure is therefore the per-dataset C2ST, which fits one classifier per dataset and averages accuracies over datasets. With only $K/2 = 100$ draws per side, the accuracy of a finite classifier exceeds 0.5 even for identical distributions. We thus report each accuracy next to a draw-split null floor: the first half of each approximator's draws is used for the real comparison; the null compares the first half of the same approximator's draws with its second half; and the floor is the larger of the two nulls. Only the excess over the floor is evidence of a discrepancy. The same procedure compares CogFormerV2 before and after fine-tuning in Experiment~3.

\paragraph{Expected information gain}
\label{par:eig}
For the design comparison in Experiment~1, we measure how informative a design is about its coefficients by the expected information gain, the mutual information between coefficients and data,
\begin{equation}
\label{eq:eig}
  \mathrm{EIG} = \mathrm{H}\bigl[p(\rmB)\bigr] - \mathbb{E}_{p(\mathcal{A})}\Bigl[\mathrm{H}\bigl[p(\rmB \mid \mathcal{A})\bigr]\Bigr]
  \approx \tfrac{1}{2} \log\det \boldsymbol{\Sigma}_{\rmB} - \tfrac{1}{2}\, \mathbb{E}_{p(\mathcal{A})}\bigl[\log\det \boldsymbol{\Sigma}_{\rmB \mid \mathcal{A}}\bigr],
\end{equation}
in nats, where $\mathrm{H}$ denotes differential entropy and the approximation replaces both distributions with Gaussians. Here, $\boldsymbol{\Sigma}_{\rmB}$ is the sample covariance of the active coefficients over 1,500 prior draws, $\boldsymbol{\Sigma}_{\rmB \mid \mathcal{A}}$ is the sample covariance of the posterior draws for one dataset, and the expectation is estimated by averaging over the $S$ simulated datasets. To place binary and continuous predictors on a common estimand, each slope is multiplied by the realized standard deviation of its predictor before both covariances are computed. Because EIG is measured relative to each design's own prior, designs with wider priors can gain more information; we therefore report it together with NRMSE.

\paragraph{Posterior predictive checks}
For empirical data, we draw 500 posterior samples for each participant. Each draw simulates that participant's own trials under their own design, including the contamination process, and the simulated trials are pooled over participants, giving one replicated dataset per draw. We compare summary statistics of the replicated and observed data: the 0.1, 0.3, 0.5, 0.7, and 0.9 quantiles of RT (and, for the CDM, of the response angle), plus response proportions for discrete choices, computed per level of each experimental condition. We report the posterior predictive median and its central 95\% interval across draws.

\paragraph{Representational similarity}
\label{par:rsa}
In Experiment~4, we compare how the frozen CogFormerV2 encoder and a learned single-model summary network organize the same aDDM datasets. To obtain unbiased dissimilarity estimates, we simulate several independent datasets for each ground-truth parameter setting and encode each with both networks. Representational similarity analysis \citep[RSA;][]{kriegeskorte2008representational} correlates the two networks' representational dissimilarity matrices, which we estimate with the cross-validated Mahalanobis (crossnobis) distance \citep{walther2016reliability}. Unlike plain distance correlations, it is not inflated by measurement noise. As a cross-check, we compute linear centered kernel alignment \citep[CKA;][]{kornblith2019similarity} with the unbiased HSIC estimator \citep{song2022fast}, which compares the centered Gram matrices of the two representations. CKA is invariant to orthogonal transformations and isotropic scaling, but it is dominated by a few high-variance directions.

Both scores depend on the dimensionality gap between the two representations: even a random linear readout of one representation attains a nonzero similarity with it. We therefore compare each score with a dimensionality-matched null, formed by random Gaussian projections of the frozen summary to the dimensionality of the learned one. We report the observed score together with its $z$-score relative to this null.

\pagebreak
\section{Additional Results}

\subsection{Experiment 1: Generalization across model space}

\begin{figure}[t]
  \centering
    \includegraphics[width=\textwidth]{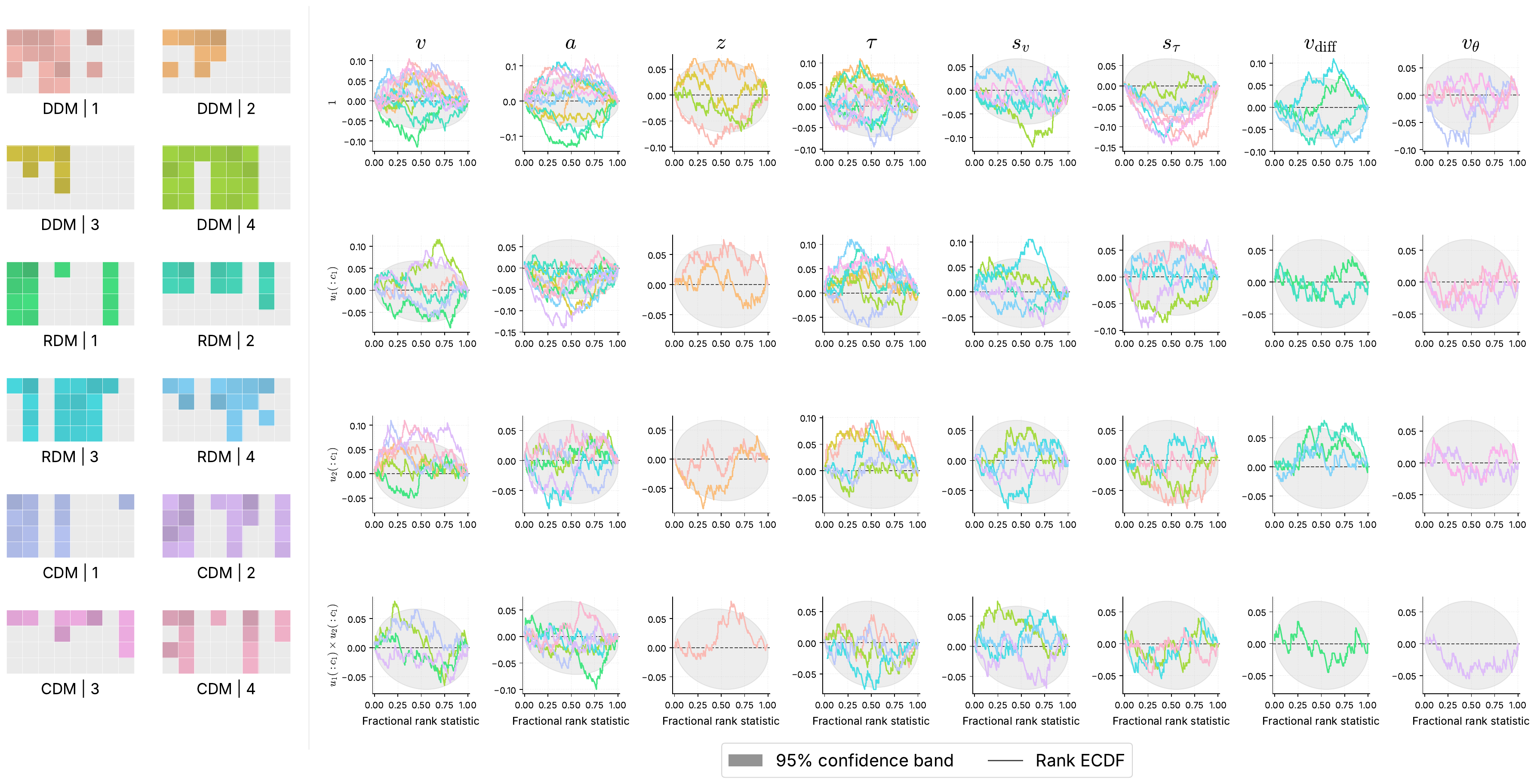}
  \caption{\textbf{Calibration across random model configurations.} \textit{Left}: the 12 designs of \autoref{fig:exp1-main}; brighter hues indicate lower calibration error. \textit{Right}: rank ECDF minus the uniform CDF for each coefficient (rows) and parameter (columns), one line per design, from 200 datasets of 500 trials per design. Curves inside the gray 95\% band are consistent with calibration.}
    \label{fig:exp1-random-ensemble-ecdf} 
\end{figure}

\begin{figure}[t]
    \centering
    \includegraphics[width=\linewidth]{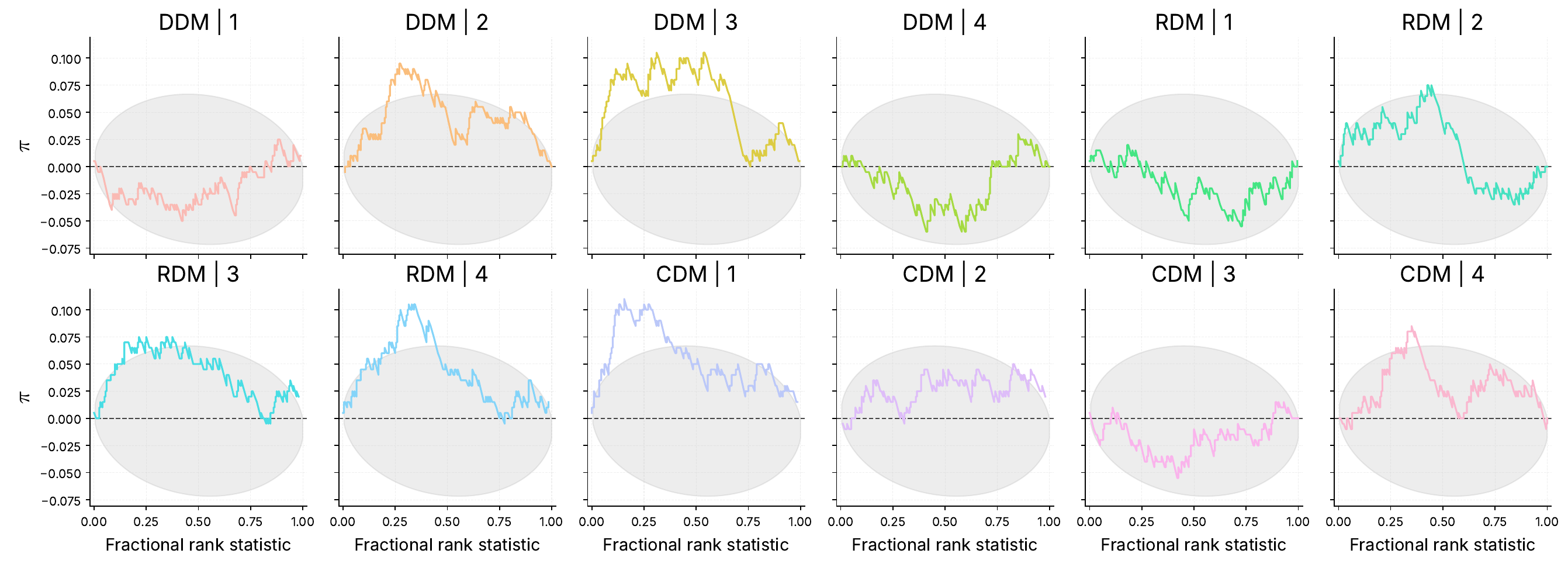}
    \caption{\textbf{Calibration of the contamination probability $\pi$} for the 12 designs in \autoref{fig:exp1-random-ensemble-ecdf}. Curves above zero indicate overestimation; the gray band is the 95\% confidence band.}
    \label{fig:exp1-random-ensemble-ecdf-pi}
\end{figure}

\begin{figure}[t]
  \centering
    \includegraphics[width=\textwidth]{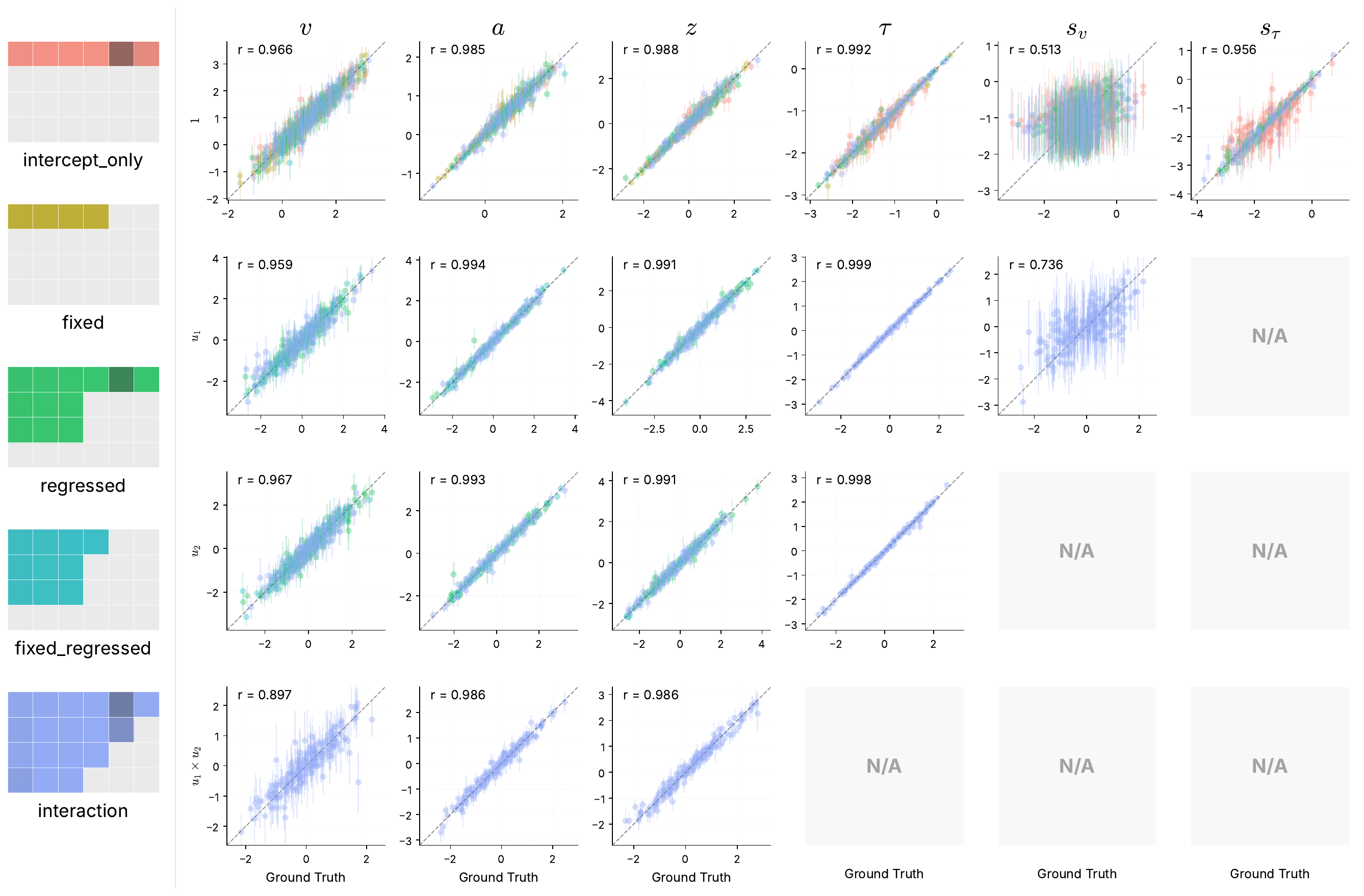}
  \caption{\textbf{DDM benchmark designs: parameter recovery, BayesFlow baselines.} \textit{Left}: the five benchmark configurations (\autoref{tab:benchmark-configs}); colored cells mark estimated coefficients, brighter hues indicating better recovery. \textit{Right}: posterior means against ground truth with credible intervals, 200 test datasets of 500 trials per configuration (colors); $r$ is the pooled Pearson correlation. Each configuration has its own BayesFlow estimator.}
    \label{fig:exp1-ddm-bench-ensemble-recovery}
\end{figure}

\begin{figure}[t]
  \centering
    \includegraphics[width=\textwidth]{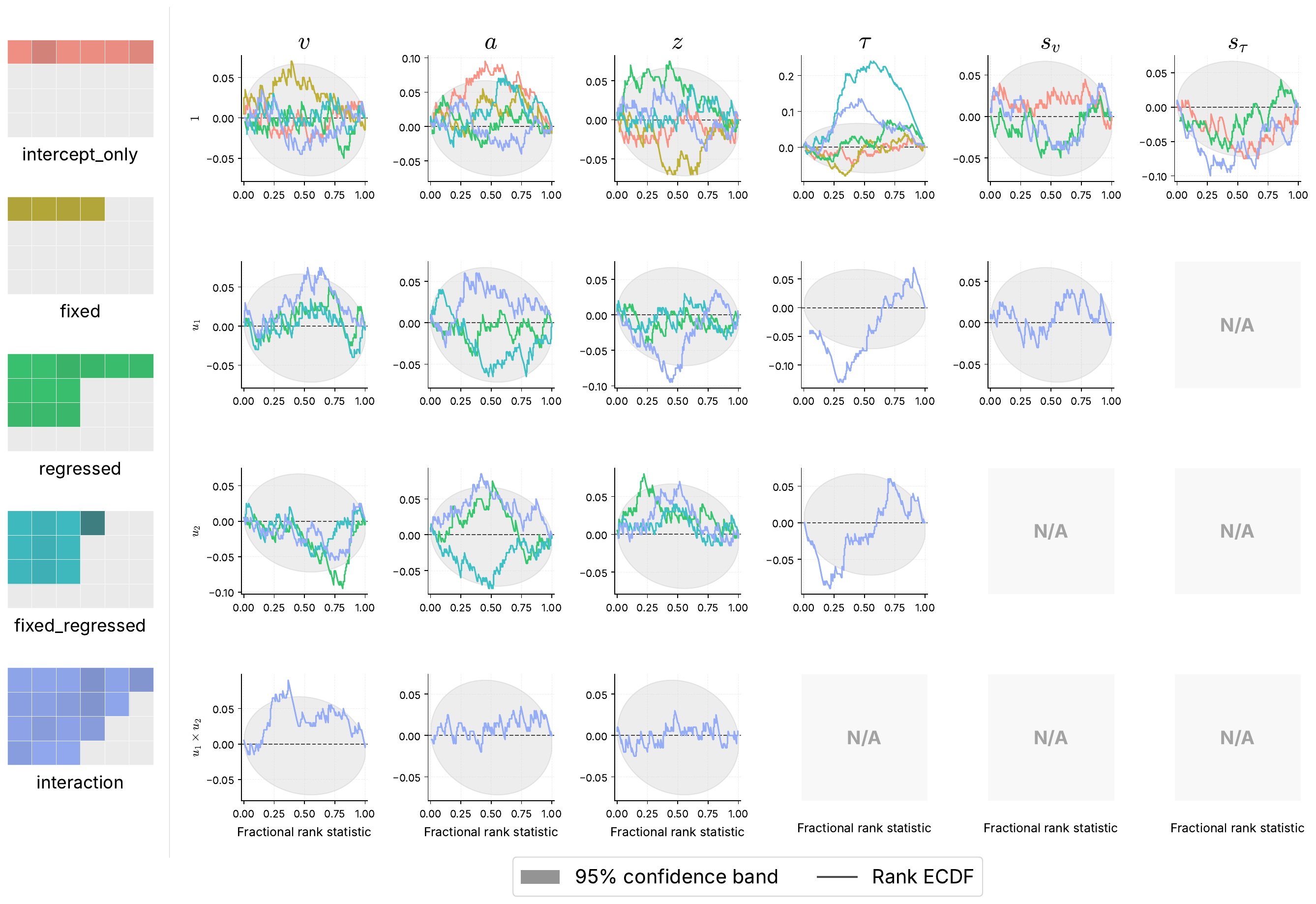}
  \caption{\textbf{DDM benchmark designs: calibration, BayesFlow baselines.} As \autoref{fig:exp1-ddm-bench-ensemble-recovery}, showing rank ECDF minus the uniform CDF; curves inside the gray 95\% band are consistent with calibration. Brighter hues indicate lower calibration error.}
    \label{fig:exp1-ddm-bench-ensemble-ecdf}
\end{figure}

\begin{figure}[t]
  \centering
    \includegraphics[width=\textwidth]{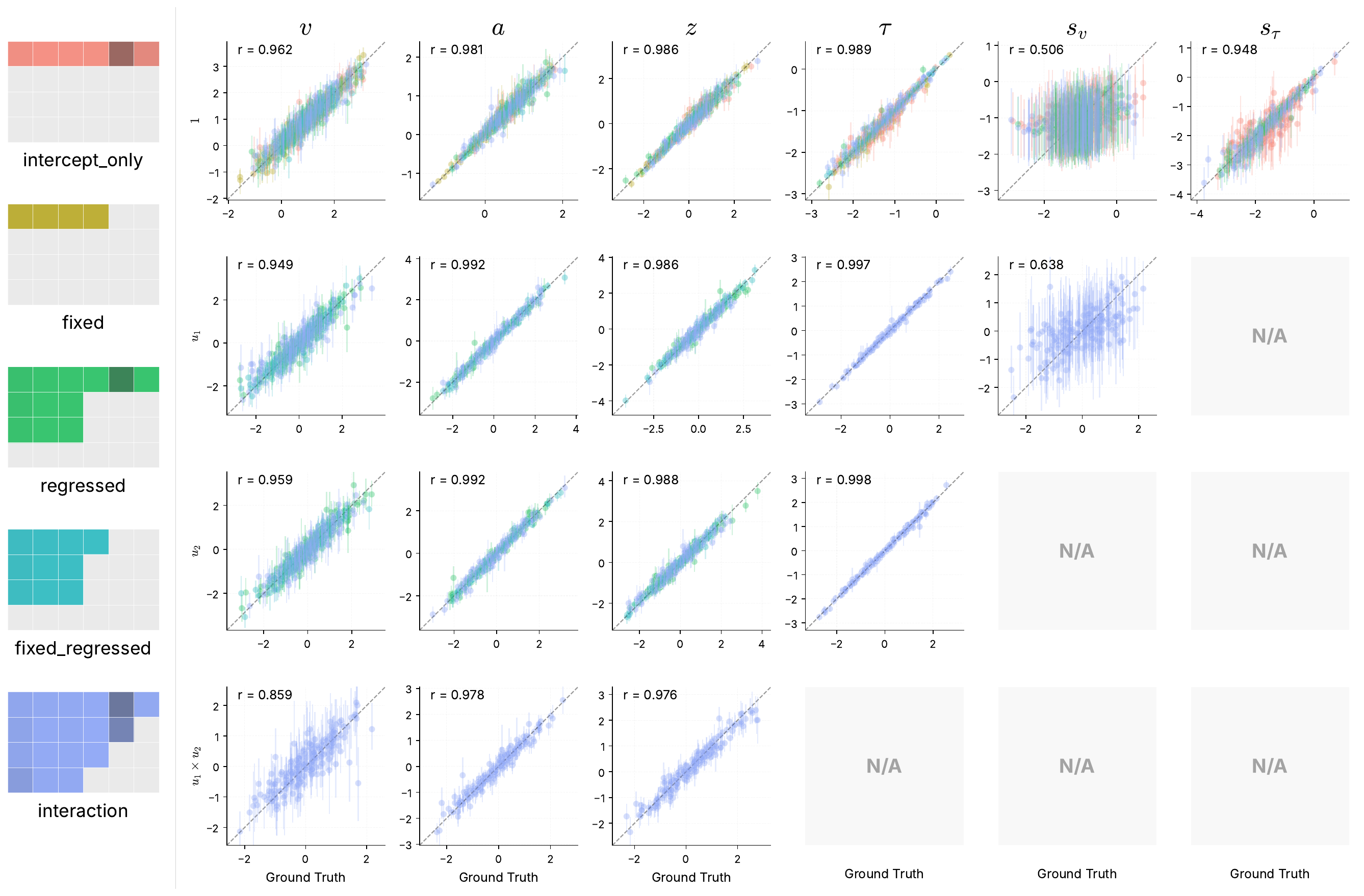}
    \includegraphics[width=\textwidth]{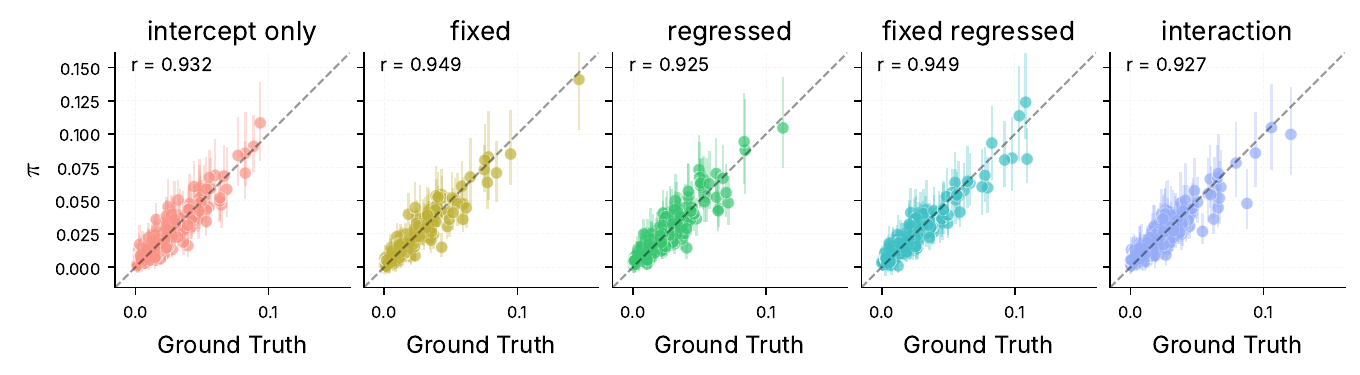}
  \caption{\textbf{DDM benchmark designs: parameter recovery, CogFormerV2.} Same test datasets and layout as \autoref{fig:exp1-ddm-bench-ensemble-recovery}.}
    \label{fig:exp1-ddm-bench-ensemble-recovery-cf}
\end{figure}

\begin{figure}[t]
  \centering
    \includegraphics[width=\textwidth]{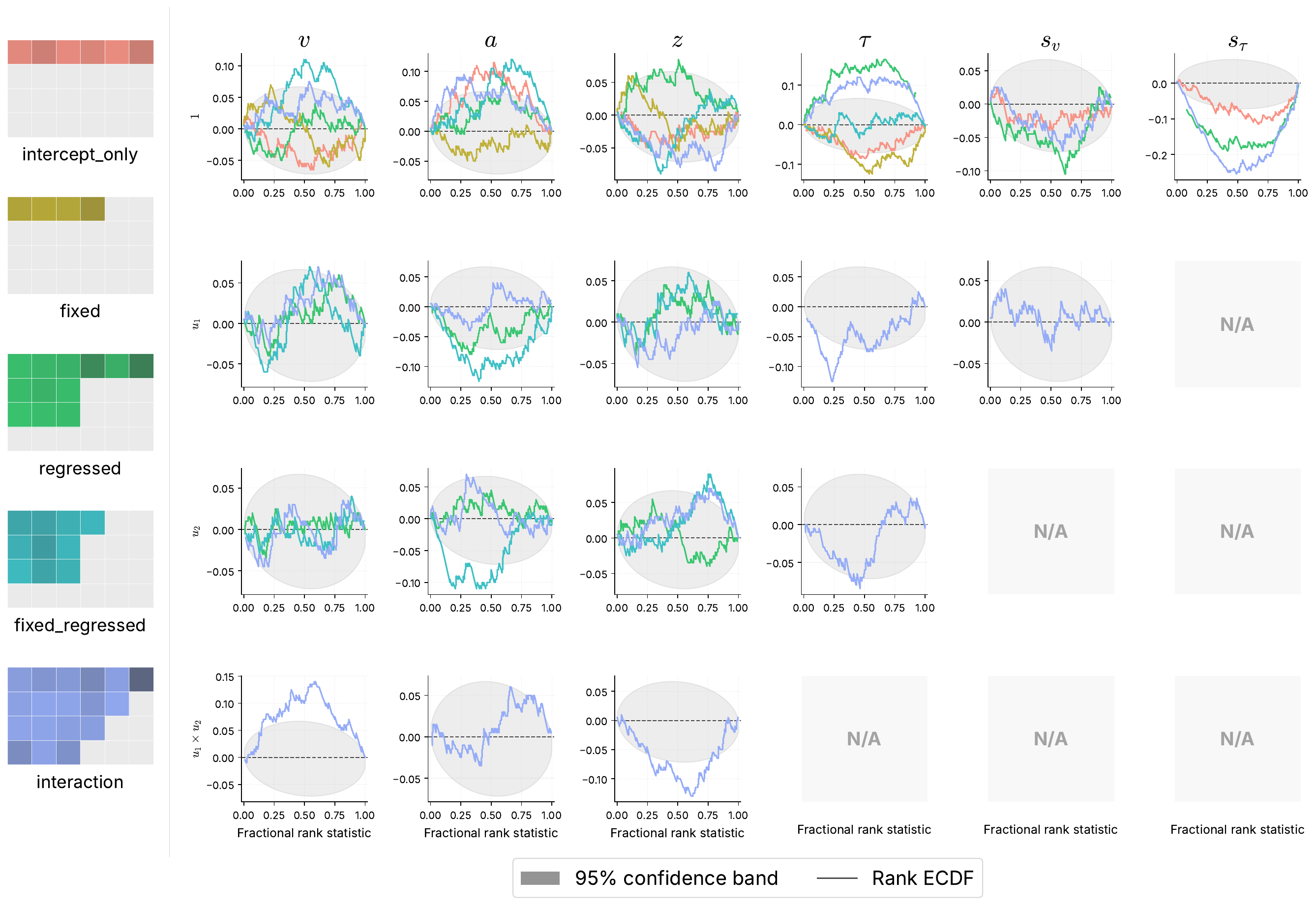}
    \includegraphics[width=\textwidth]{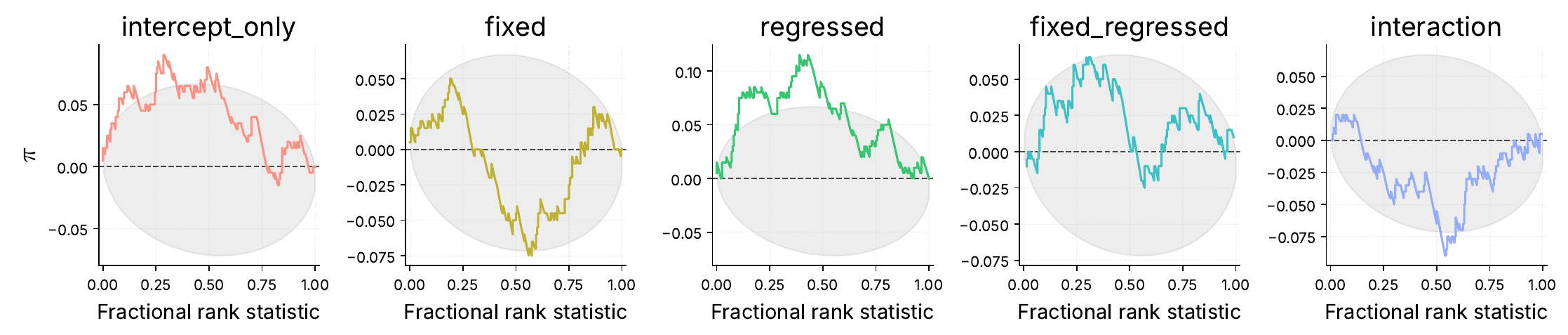}
  \caption{\textbf{DDM benchmark designs: calibration, CogFormerV2.} Same test datasets and layout as \autoref{fig:exp1-ddm-bench-ensemble-ecdf}.}
    \label{fig:exp1-ddm-bench-ensemble-ecdf-cf}
\end{figure}

\begin{table}[t]
\centering
\small
\caption{\textbf{DDM benchmark metrics.} Median $\pm$ SEM over each configuration's parameters for the BayesFlow baselines and CogFormerV2, on the same test datasets. Metrics are defined in \autoref{ssec:metrics}.}
\label{tab:benchmark-metrics-ddm}
\begin{tabularx}{\textwidth}{llCCC}
\toprule
\textbf{Configuration} & \textbf{Method} & \textbf{NRMSE} ($\downarrow$) & \textbf{Cal. Error} ($\downarrow$) & \textbf{Post. Contr.} ($\uparrow$) \\
\midrule
Intercept only & \textbf{Baseline} & $0.075 \pm 0.022$ & $0.012 \pm 0.002$ & $0.942 \pm 0.094$ \\
 & \textbf{CogFormerV2} & $0.074 \pm 0.022$ & $0.012 \pm 0.006$ & $0.949 \pm 0.087$ \\
\midrule
Fixed & \textbf{Baseline} & $0.040 \pm 0.007$ & $0.023 \pm 0.007$ & $0.993 \pm 0.012$ \\
 & \textbf{CogFormerV2} & $0.040 \pm 0.007$ & $0.035 \pm 0.007$ & $0.993 \pm 0.011$ \\
\midrule
Regressed & \textbf{Baseline} & $0.042 \pm 0.014$ & $0.017 \pm 0.003$ & $0.984 \pm 0.061$ \\
 & \textbf{CogFormerV2} & $0.057 \pm 0.015$ & $0.022 \pm 0.003$ & $0.977 \pm 0.066$ \\
\midrule
Fixed + regressed & \textbf{Baseline} & $0.030 \pm 0.007$ & $0.016 \pm 0.005$ & $0.989 \pm 0.007$ \\
 & \textbf{CogFormerV2} & $0.038 \pm 0.007$ & $0.026 \pm 0.005$ & $0.979 \pm 0.008$ \\
\midrule
Interaction & \textbf{Baseline} & $0.045 \pm 0.013$ & $0.018 \pm 0.004$ & $0.975 \pm 0.038$ \\
 & \textbf{CogFormerV2} & $0.058 \pm 0.013$ & $0.023 \pm 0.002$ & $0.962 \pm 0.043$ \\
\midrule
All configurations & \textbf{Baseline} & $0.046 \pm 0.007$ & $0.017 \pm 0.002$ & $0.978 \pm 0.023$ \\
 & \textbf{CogFormerV2} & $0.054 \pm 0.007$ & $0.023 \pm 0.002$ & $0.968 \pm 0.024$ \\
\bottomrule
\end{tabularx}
\end{table}

\begin{figure}[t]
  \centering
    \includegraphics[width=\textwidth]{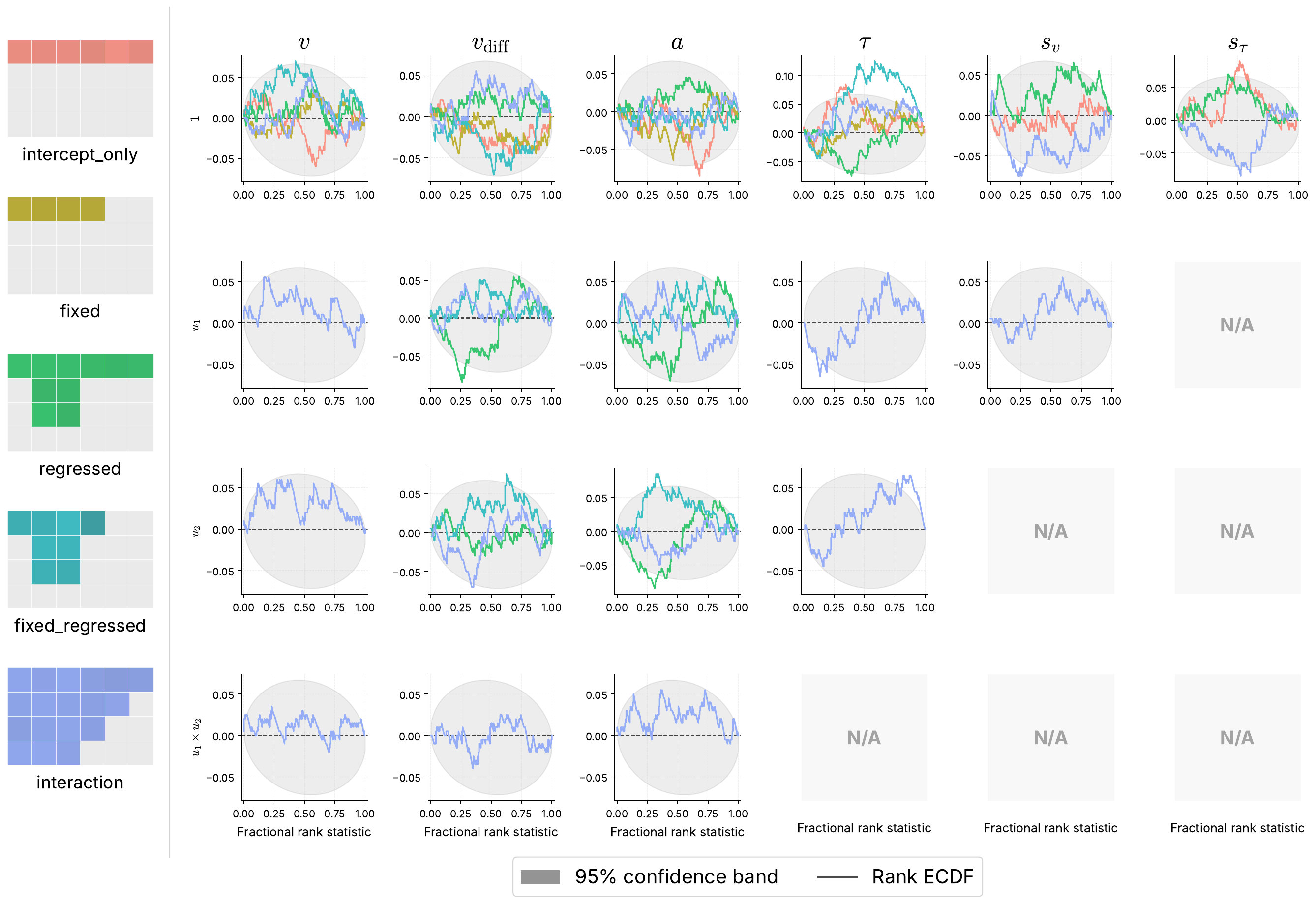}
  \caption{\textbf{RDM benchmark designs: calibration, BayesFlow baselines.} Layout as in \autoref{fig:exp1-ddm-bench-ensemble-ecdf}.}
    \label{fig:exp1-rdm-bench-ensemble-ecdf}
\end{figure}

\begin{figure}[t]
  \centering
    \includegraphics[width=\textwidth]{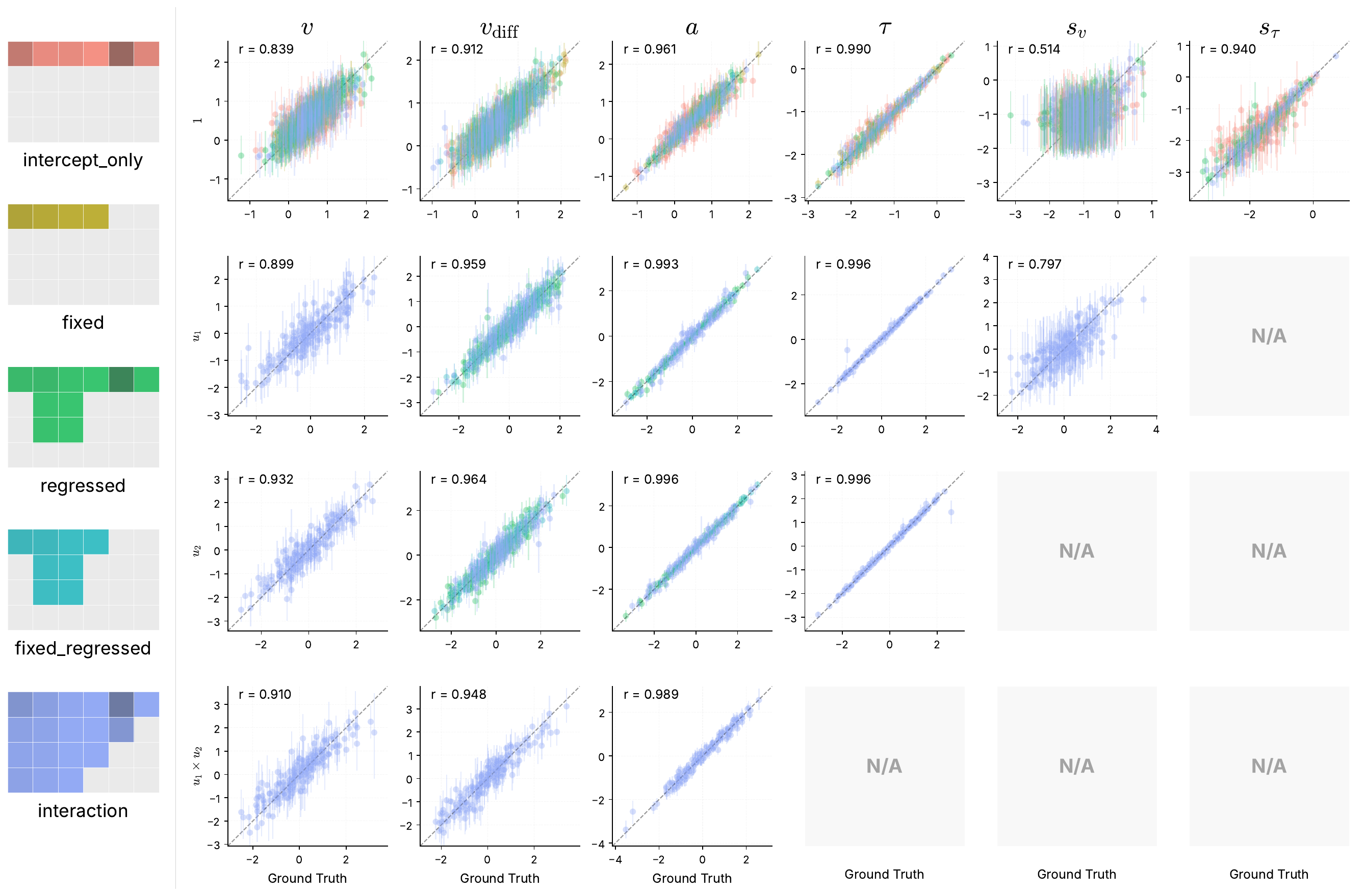}
  \caption{\textbf{RDM benchmark designs: parameter recovery, BayesFlow baselines.} Layout as in \autoref{fig:exp1-ddm-bench-ensemble-recovery}.}
    \label{fig:exp1-rdm-bench-ensemble-recovery}
\end{figure}

\begin{figure}[t]
  \centering
    \includegraphics[width=\textwidth]{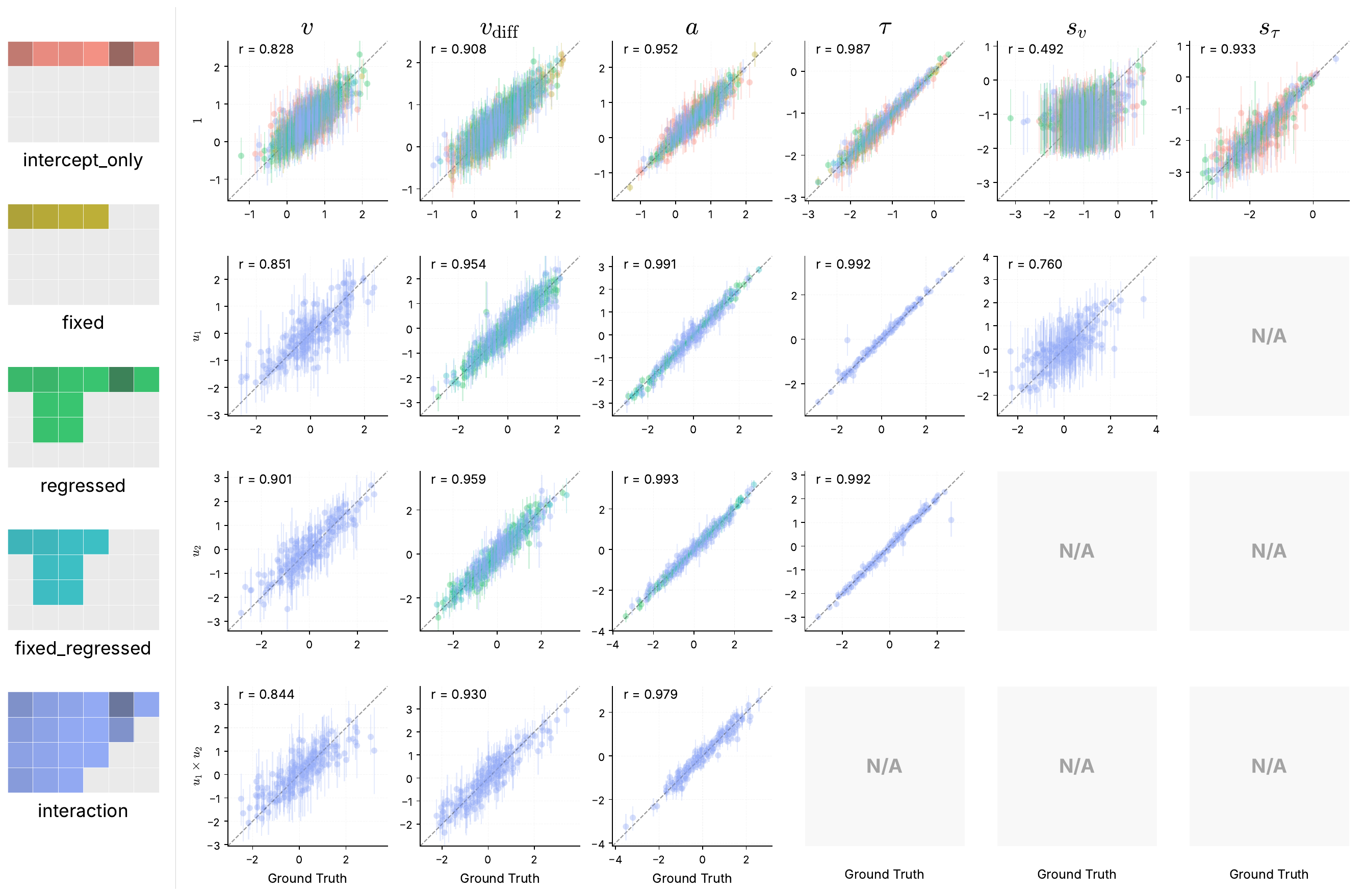}
    \includegraphics[width=\textwidth]{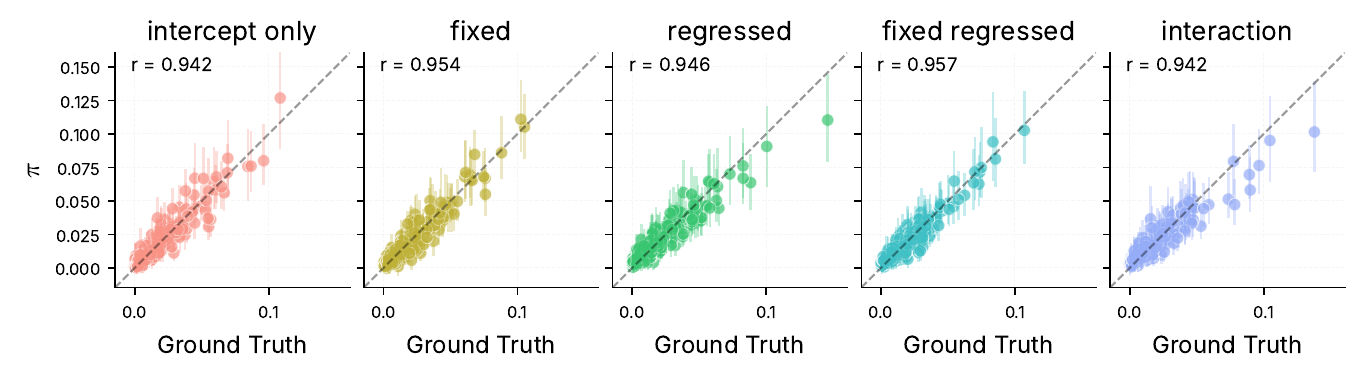}
  \caption{\textbf{RDM benchmark designs: parameter recovery, CogFormerV2.} Same test datasets and layout as \autoref{fig:exp1-rdm-bench-ensemble-recovery}.}
    \label{fig:exp1-rdm-bench-ensemble-recovery-cf}
\end{figure}

\begin{figure}[t]
  \centering
    \includegraphics[width=\textwidth]{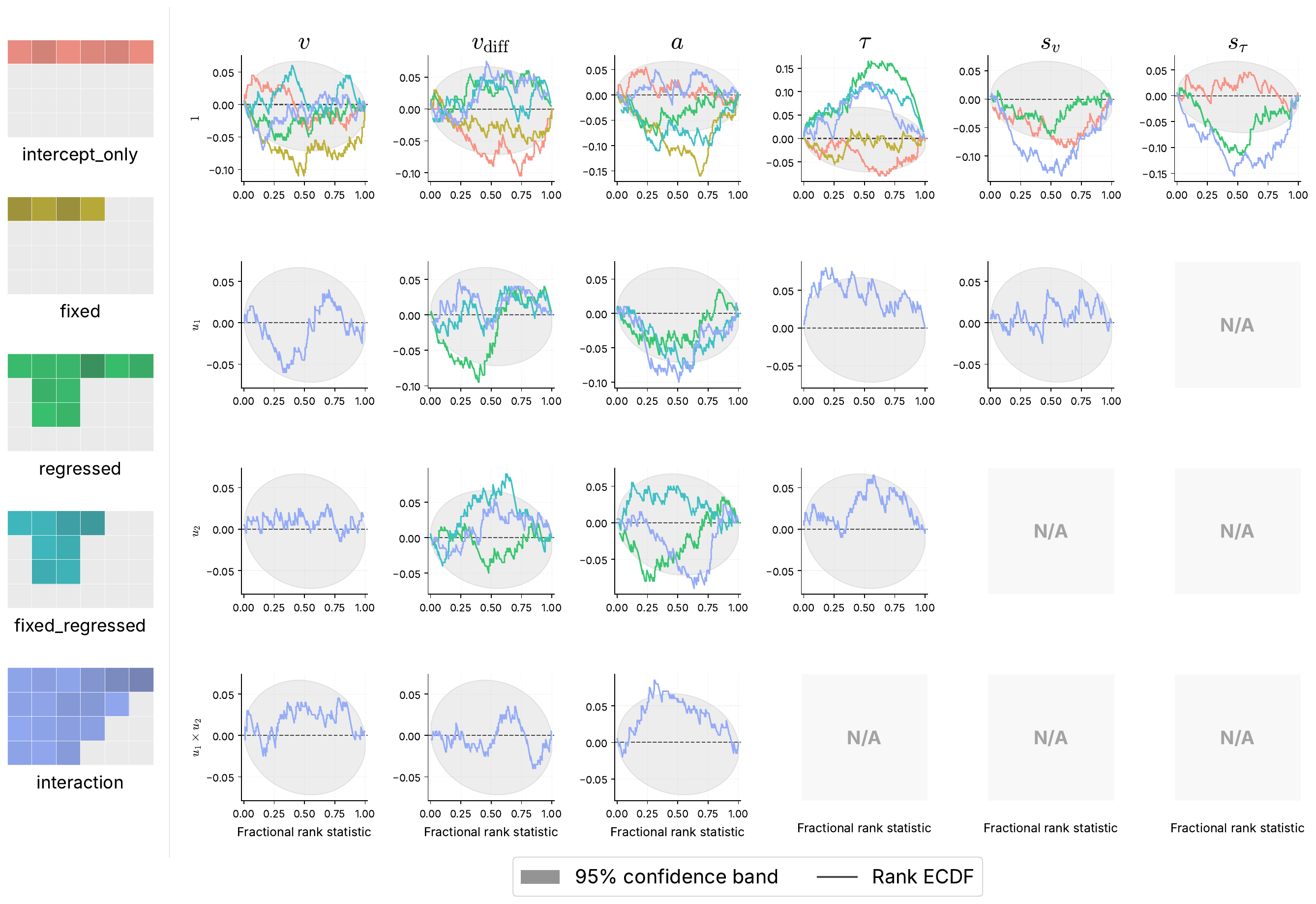}
    \includegraphics[width=\textwidth]{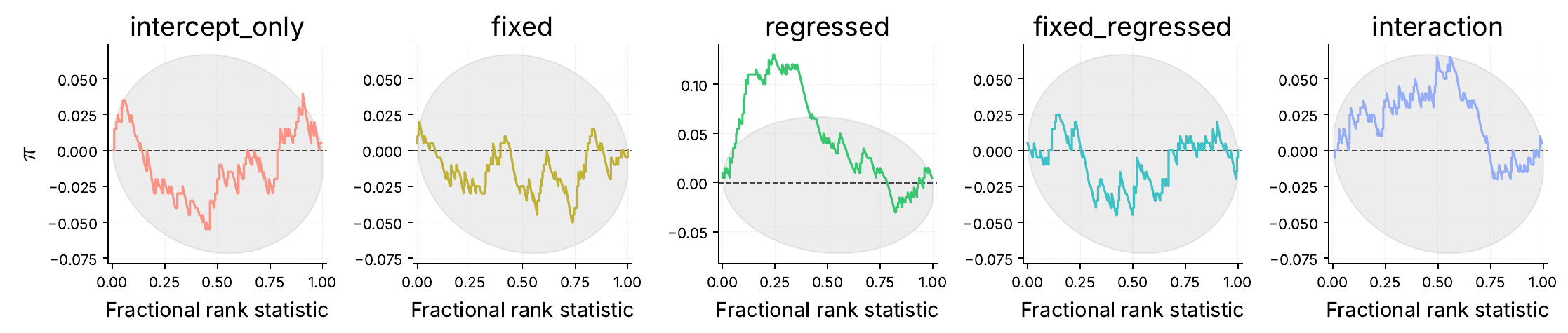}
  \caption{\textbf{RDM benchmark designs: calibration, CogFormerV2.} Same test datasets and layout as \autoref{fig:exp1-rdm-bench-ensemble-ecdf}.}
    \label{fig:exp1-rdm-bench-ensemble-ecdf-cf}
\end{figure}

\begin{table}[t]
\centering
\small
\caption{\textbf{RDM benchmark metrics.} Median $\pm$ SEM over each configuration's parameters for the BayesFlow baselines and CogFormerV2, on the same test datasets. Metrics are defined in \autoref{ssec:metrics}.}
\label{tab:benchmark-metrics-rdm}
\begin{tabularx}{\textwidth}{llCCC}
\toprule
\textbf{Configuration} & \textbf{Method} & \textbf{NRMSE} ($\downarrow$) & \textbf{Cal. Error} ($\downarrow$) & \textbf{Post. Contr.} ($\uparrow$) \\
\midrule
Intercept only & \textbf{Baseline} & $0.109 \pm 0.023$ & $0.024 \pm 0.003$ & $0.844 \pm 0.115$ \\
 & \textbf{CogFormerV2} & $0.110 \pm 0.023$ & $0.018 \pm 0.004$ & $0.842 \pm 0.106$ \\
\midrule
Fixed & \textbf{Baseline} & $0.081 \pm 0.018$ & $0.017 \pm 0.003$ & $0.928 \pm 0.040$ \\
 & \textbf{CogFormerV2} & $0.082 \pm 0.019$ & $0.017 \pm 0.001$ & $0.945 \pm 0.039$ \\
\midrule
Regressed & \textbf{Baseline} & $0.068 \pm 0.014$ & $0.019 \pm 0.006$ & $0.953 \pm 0.064$ \\
 & \textbf{CogFormerV2} & $0.073 \pm 0.014$ & $0.018 \pm 0.008$ & $0.937 \pm 0.068$ \\
\midrule
Fixed + regressed & \textbf{Baseline} & $0.048 \pm 0.013$ & $0.018 \pm 0.005$ & $0.969 \pm 0.019$ \\
 & \textbf{CogFormerV2} & $0.056 \pm 0.013$ & $0.015 \pm 0.002$ & $0.958 \pm 0.023$ \\
\midrule
Interaction & \textbf{Baseline} & $0.088 \pm 0.012$ & $0.018 \pm 0.003$ & $0.911 \pm 0.040$ \\
 & \textbf{CogFormerV2} & $0.100 \pm 0.012$ & $0.020 \pm 0.002$ & $0.869 \pm 0.044$ \\
\midrule
All configurations & \textbf{Baseline} & $0.075 \pm 0.007$ & $0.019 \pm 0.002$ & $0.928 \pm 0.027$ \\
 & \textbf{CogFormerV2} & $0.082 \pm 0.007$ & $0.018 \pm 0.002$ & $0.936 \pm 0.027$ \\
\bottomrule
\end{tabularx}
\end{table}

\begin{figure}[t]
  \centering
    \includegraphics[width=\textwidth]{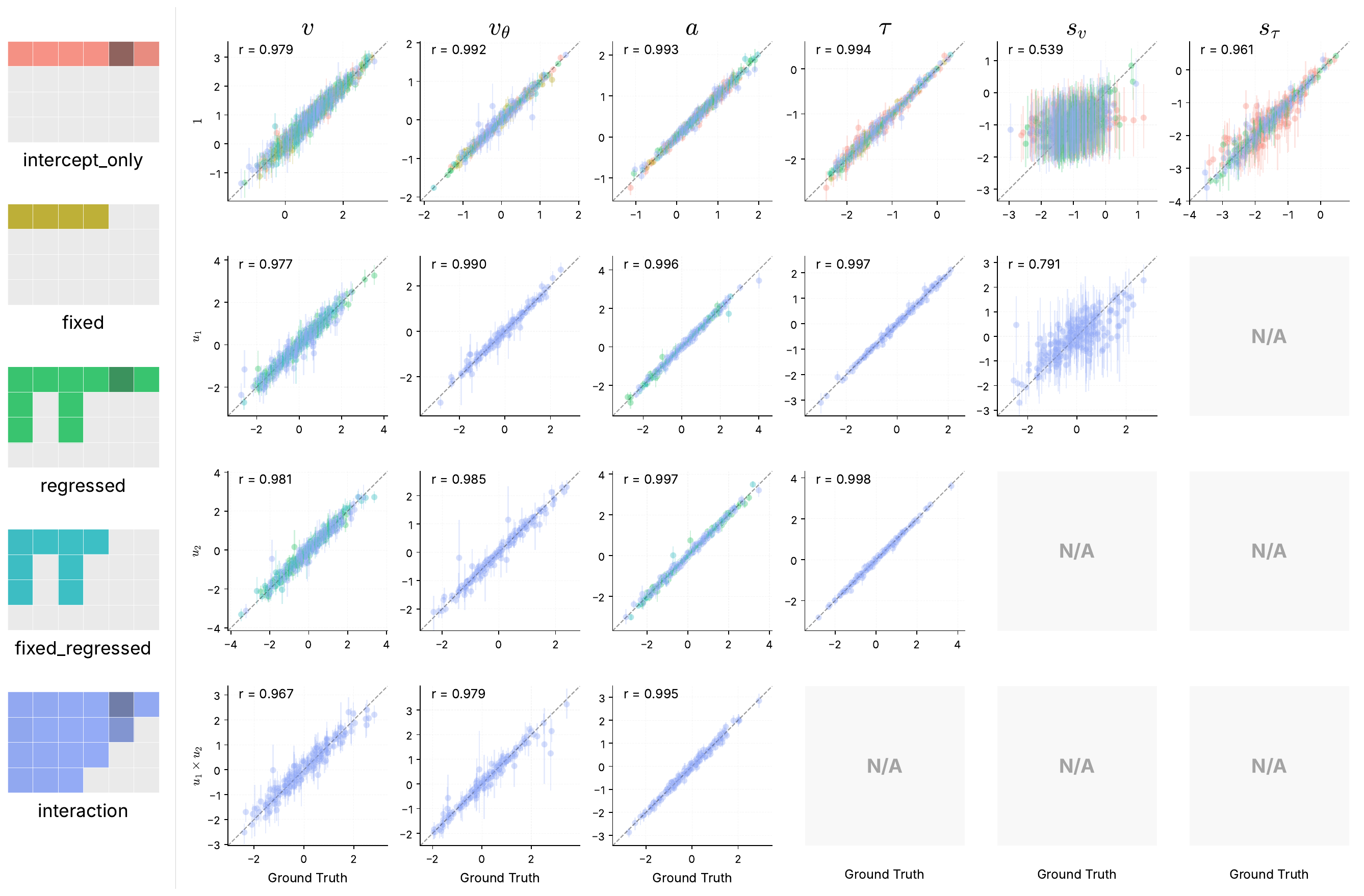}
  \caption{\textbf{CDM benchmark designs: parameter recovery, BayesFlow baselines.} Layout as in \autoref{fig:exp1-ddm-bench-ensemble-recovery}.}
    \label{fig:exp1-cdm-bench-ensemble-recovery}
\end{figure}

\begin{figure}[t]
  \centering
    \includegraphics[width=\textwidth]{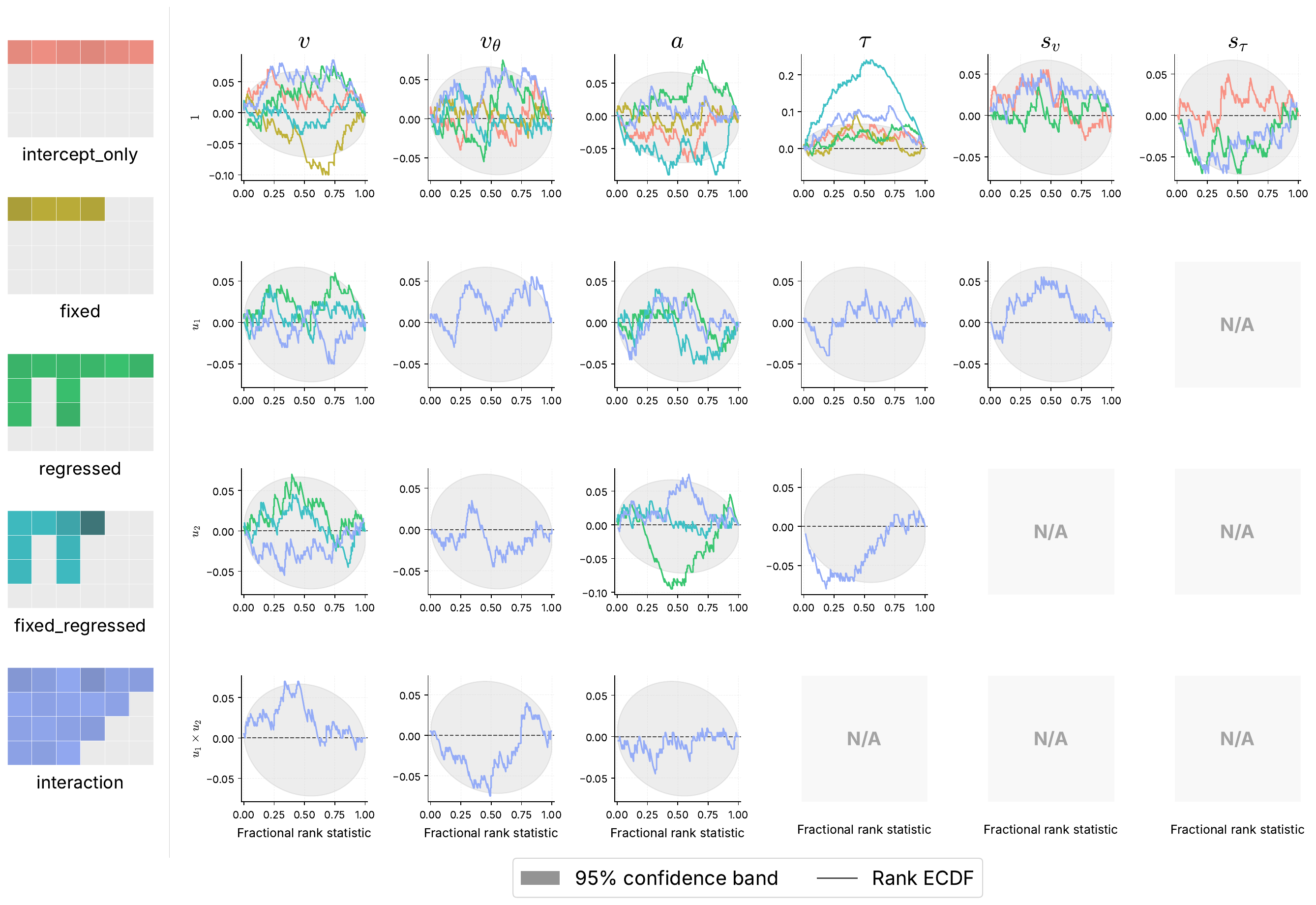}
  \caption{\textbf{CDM benchmark designs: calibration, BayesFlow baselines.} Layout as in \autoref{fig:exp1-ddm-bench-ensemble-ecdf}.}
    \label{fig:exp1-cdm-bench-ensemble-ecdf}
\end{figure}

\begin{figure}[t]
  \centering
    \includegraphics[width=\textwidth]{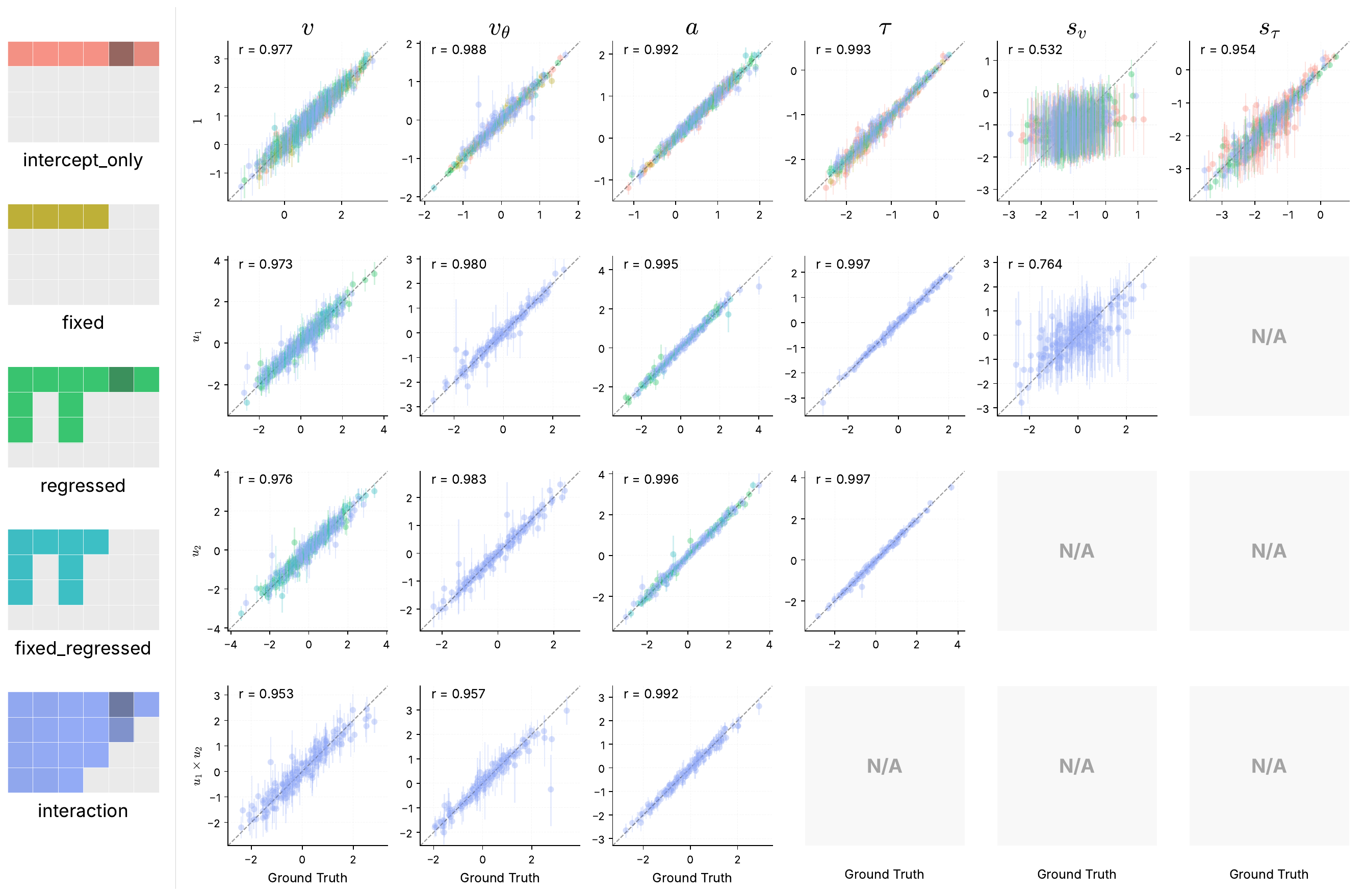}
    \includegraphics[width=\textwidth]{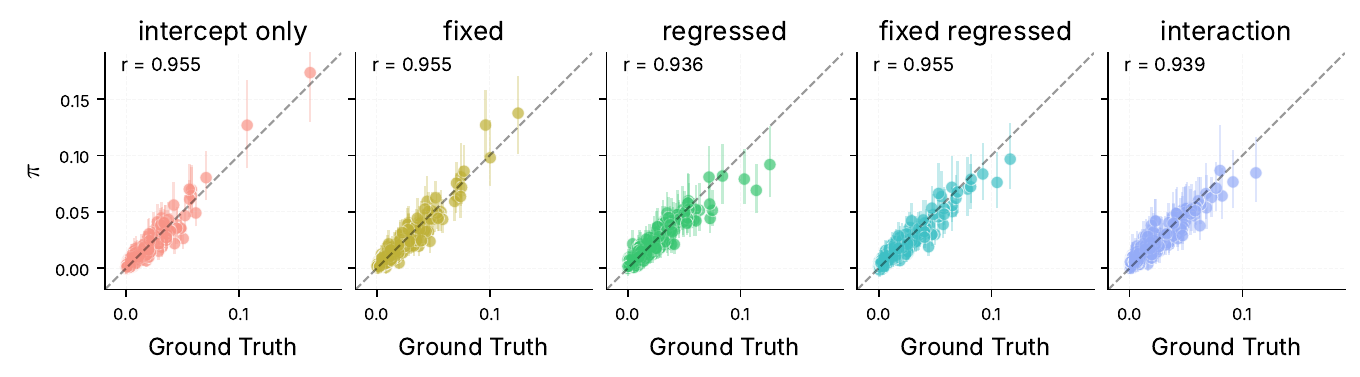}
  \caption{\textbf{CDM benchmark designs: parameter recovery, CogFormerV2.} Same test datasets and layout as \autoref{fig:exp1-cdm-bench-ensemble-recovery}.}
    \label{fig:exp1-cdm-bench-ensemble-recovery-cf}
\end{figure}

\begin{figure}[t]
  \centering
    \includegraphics[width=\textwidth]{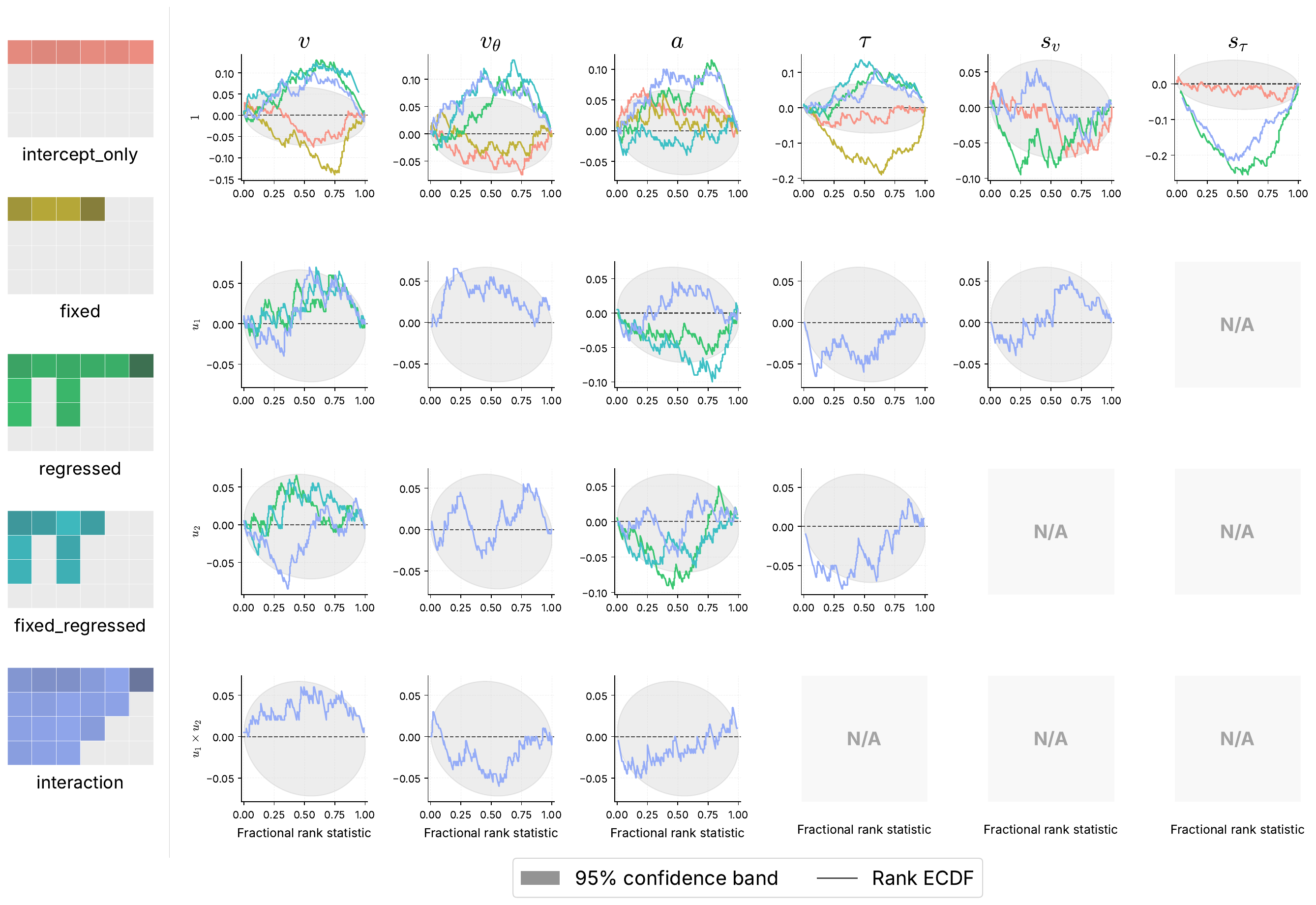}
    \includegraphics[width=\textwidth]{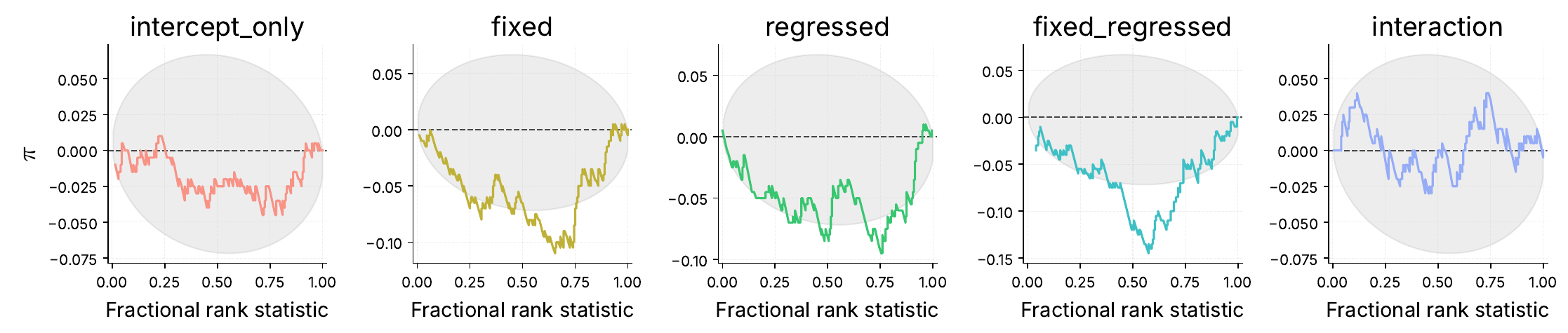}
  \caption{\textbf{CDM benchmark designs: calibration, CogFormerV2.} Same test datasets and layout as \autoref{fig:exp1-cdm-bench-ensemble-ecdf}.}
    \label{fig:exp1-cdm-bench-ensemble-ecdf-cf}
\end{figure}

\begin{table}[t]
\centering
\small
\caption{\textbf{CDM benchmark metrics.} Median $\pm$ SEM over each configuration's parameters for the BayesFlow baselines and CogFormerV2, on the same test datasets. Metrics are defined in \autoref{ssec:metrics}.}
\label{tab:benchmark-metrics-cdm}
\begin{tabularx}{\textwidth}{llCCC}
\toprule
\textbf{Configuration} & \textbf{Method} & \textbf{NRMSE} ($\downarrow$) & \textbf{Cal. Error} ($\downarrow$) & \textbf{Post. Contr.} ($\uparrow$) \\
\midrule
Intercept only & \textbf{Baseline} & $0.050 \pm 0.026$ & $0.017 \pm 0.003$ & $0.965 \pm 0.091$ \\
 & \textbf{CogFormerV2} & $0.053 \pm 0.025$ & $0.016 \pm 0.006$ & $0.969 \pm 0.083$ \\
\midrule
Fixed & \textbf{Baseline} & $0.030 \pm 0.009$ & $0.011 \pm 0.006$ & $0.993 \pm 0.013$ \\
 & \textbf{CogFormerV2} & $0.031 \pm 0.009$ & $0.016 \pm 0.007$ & $0.992 \pm 0.011$ \\
\midrule
Regressed & \textbf{Baseline} & $0.043 \pm 0.016$ & $0.012 \pm 0.003$ & $0.986 \pm 0.059$ \\
 & \textbf{CogFormerV2} & $0.049 \pm 0.016$ & $0.024 \pm 0.005$ & $0.978 \pm 0.062$ \\
\midrule
Fixed + regressed & \textbf{Baseline} & $0.019 \pm 0.008$ & $0.018 \pm 0.003$ & $0.995 \pm 0.007$ \\
 & \textbf{CogFormerV2} & $0.024 \pm 0.008$ & $0.020 \pm 0.003$ & $0.991 \pm 0.007$ \\
\midrule
Interaction & \textbf{Baseline} & $0.053 \pm 0.010$ & $0.012 \pm 0.002$ & $0.984 \pm 0.038$ \\
 & \textbf{CogFormerV2} & $0.064 \pm 0.011$ & $0.018 \pm 0.003$ & $0.970 \pm 0.041$ \\
\midrule
All configurations & \textbf{Baseline} & $0.043 \pm 0.006$ & $0.013 \pm 0.001$ & $0.985 \pm 0.022$ \\
 & \textbf{CogFormerV2} & $0.049 \pm 0.007$ & $0.020 \pm 0.002$ & $0.978 \pm 0.023$ \\
\bottomrule
\end{tabularx}
\end{table}

\pagebreak

\subsection{Experiment 2: Fitting empirical data}
\label{app:experiment-2-figures}

This section reports posterior predictive checks for the three empirical datasets of Experiment~2 (\autoref{ssec:exp2-protocol}). For each family, we show the predictive RT density under each of two grouping variables (\autoref{tab:e2-conditions}). For the CDM, whose response is a continuous angle, we also show the predictive density of the response angle. For the DDM and RDM, whose responses are discrete, choice proportions are compared in \autoref{fig:exp2-main}. All predictive bands are 95\% intervals.

\begin{table}[htbp]
  \centering
  \small
  \begin{tabular}{@{}llll@{}}
    \toprule
    \textbf{Model Family} & \textbf{Predictors}       & \textbf{Conditions}            & \textbf{Regressed Parameters} \\
    \midrule
    DDM & Stimulus type      & Novel, Familiar        & $v, a, z, \tau, s_\tau$ \\
        & Factual truth      & False, True            & $v, a, z, \tau, s_\tau$ \\
    \addlinespace
    RDM & Congruency         & Incongruent, Congruent & $v, v_{\mathrm{diff}}, a, \tau$ \\
        & Session            & Session 0, Session 1   & $v, v_{\mathrm{diff}}, a, \tau$ \\
    \addlinespace
    CDM & Speed instruction  & Unspeeded, Speeded     & $v, a, \tau$ \\
        & Difficulty         & Easy, Medium, Hard     & $v, a, \tau$ \\
    \bottomrule
  \end{tabular}
  \caption{Parameter regressions in the Experiment~2 real-data fits.
    Each interaction is between the two predictors listed for that family.
    All parameters have an intercept. Intercept-only parameters are shared
    across conditions within each participant. Posterior predictive checks
    are grouped by the levels of each predictor.}
  \label{tab:e2-conditions}
\end{table}


\begin{figure}[htbp]
  \centering
  \includegraphics[width=\linewidth]{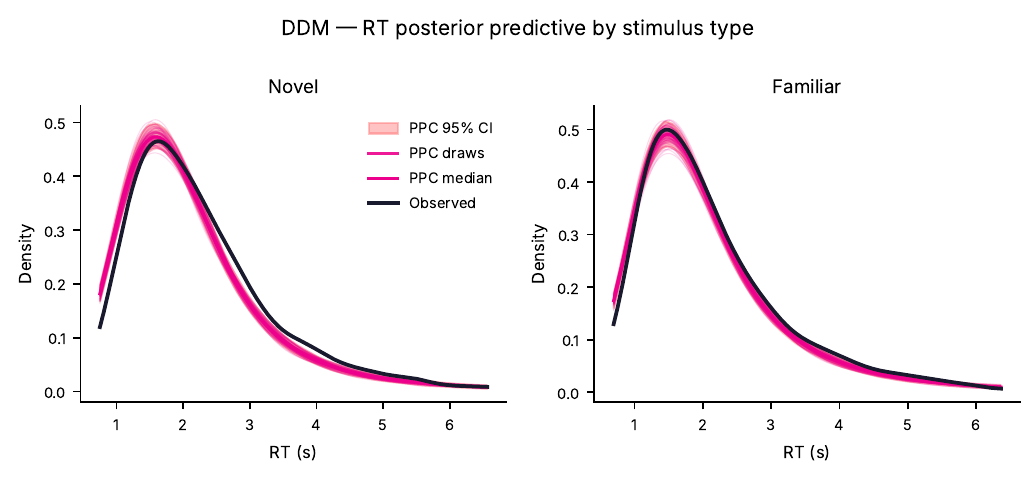}
  \caption[DDM, stimulus type: RT posterior predictive]{\textbf{DDM: RT posterior predictive by stimulus type.} Observed RT density (black) against the posterior predictive: 95\% interval (band) and individual draws (thin lines).}
  \label{fig:e2-ddm-rt_ppc-stim_type}
\end{figure}

\begin{figure}[htbp]
  \centering
  \includegraphics[width=\linewidth]{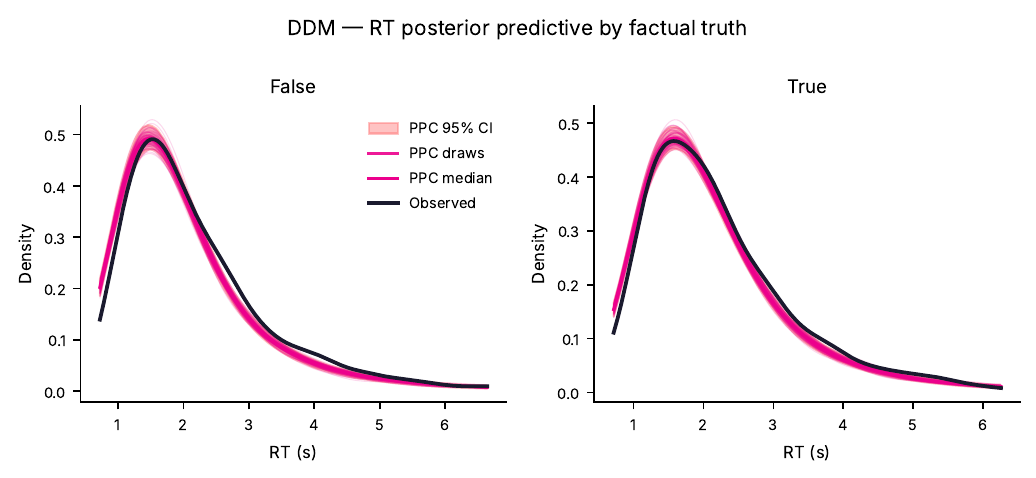}
  \caption[DDM, factual truth: RT posterior predictive]{\textbf{DDM: RT posterior predictive by factual truth.} As \autoref{fig:e2-ddm-rt_ppc-stim_type}.}
  \label{fig:e2-ddm-rt_ppc-truth}
\end{figure}

\begin{figure}[htbp]
  \centering
  \includegraphics[width=\linewidth]{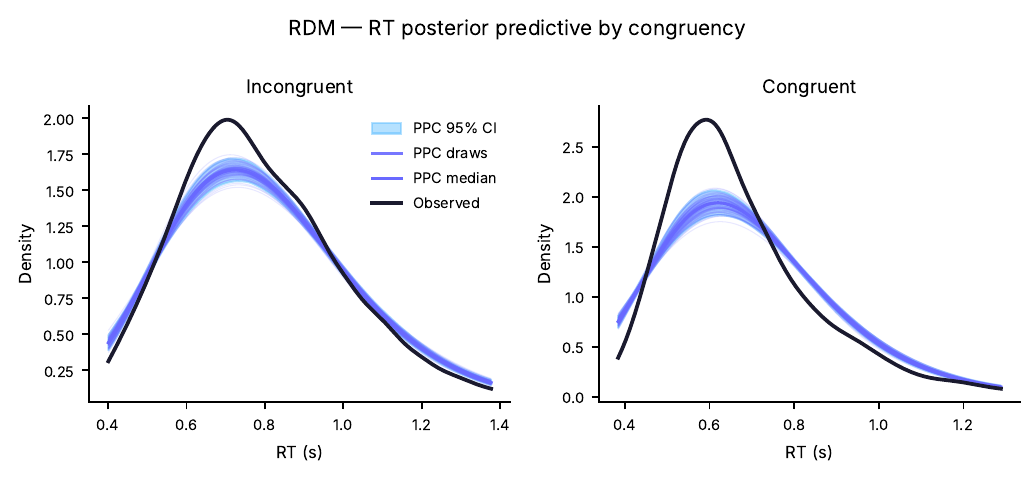}
  \caption[RDM, congruency: RT posterior predictive]{\textbf{RDM: RT posterior predictive by congruency.} As \autoref{fig:e2-ddm-rt_ppc-stim_type}.}
  \label{fig:e2-rdm-rt_ppc-congruency}
\end{figure}

\begin{figure}[htbp]
  \centering
  \includegraphics[width=\linewidth]{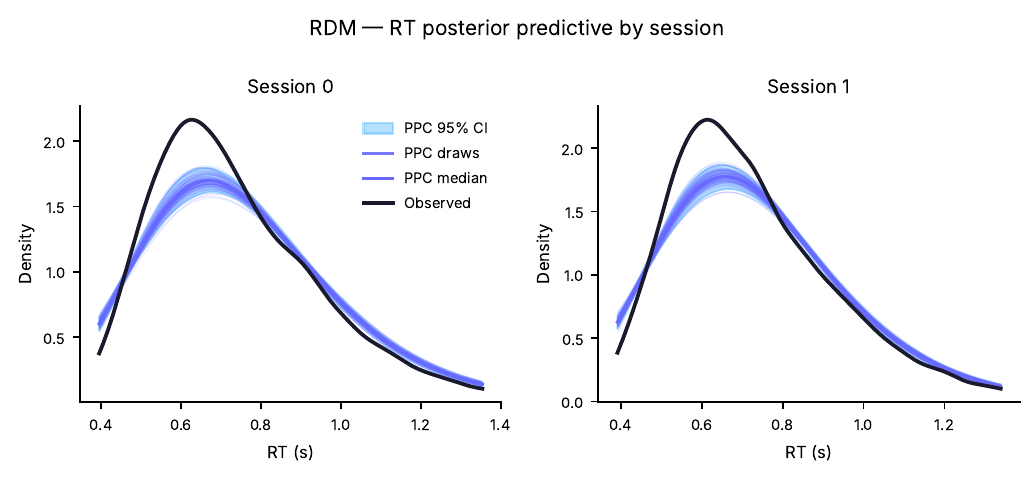}
  \caption[RDM, session: RT posterior predictive]{\textbf{RDM: RT posterior predictive by session.} As \autoref{fig:e2-ddm-rt_ppc-stim_type}.}
  \label{fig:e2-rdm-rt_ppc-session}
\end{figure}

\begin{figure}[htbp]
  \centering
  \includegraphics[width=\linewidth]{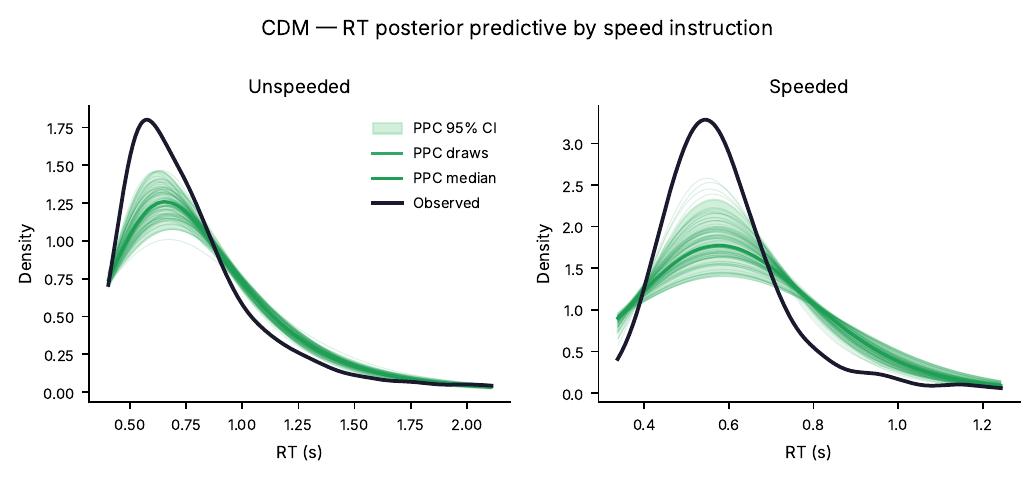}
  \caption[CDM, speed instruction: RT posterior predictive]{\textbf{CDM: RT posterior predictive by speed instruction.} As \autoref{fig:e2-ddm-rt_ppc-stim_type}.}
  \label{fig:e2-cdm-rt_ppc-speed}
\end{figure}

\begin{figure}[htbp]
  \centering
  \includegraphics[width=\linewidth]{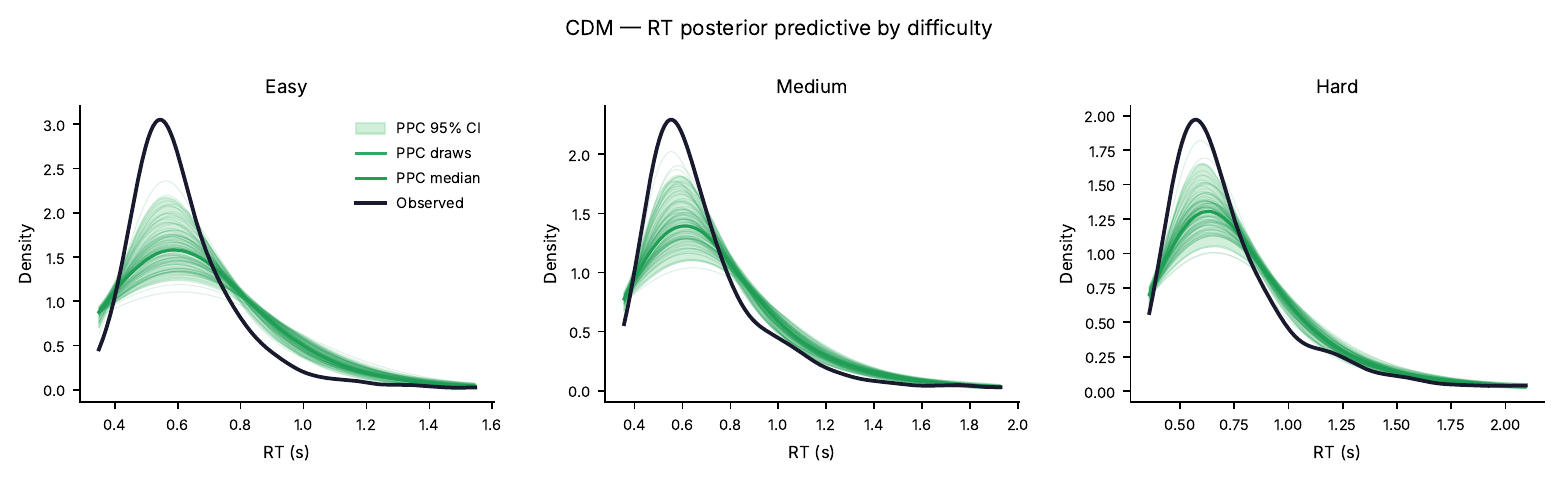}
  \caption[CDM, difficulty: RT posterior predictive]{\textbf{CDM: RT posterior predictive by difficulty.} As \autoref{fig:e2-ddm-rt_ppc-stim_type}.}
  \label{fig:e2-cdm-rt_ppc-difficulty}
\end{figure}

\begin{figure}[htbp]
  \centering
  \includegraphics[width=\linewidth]{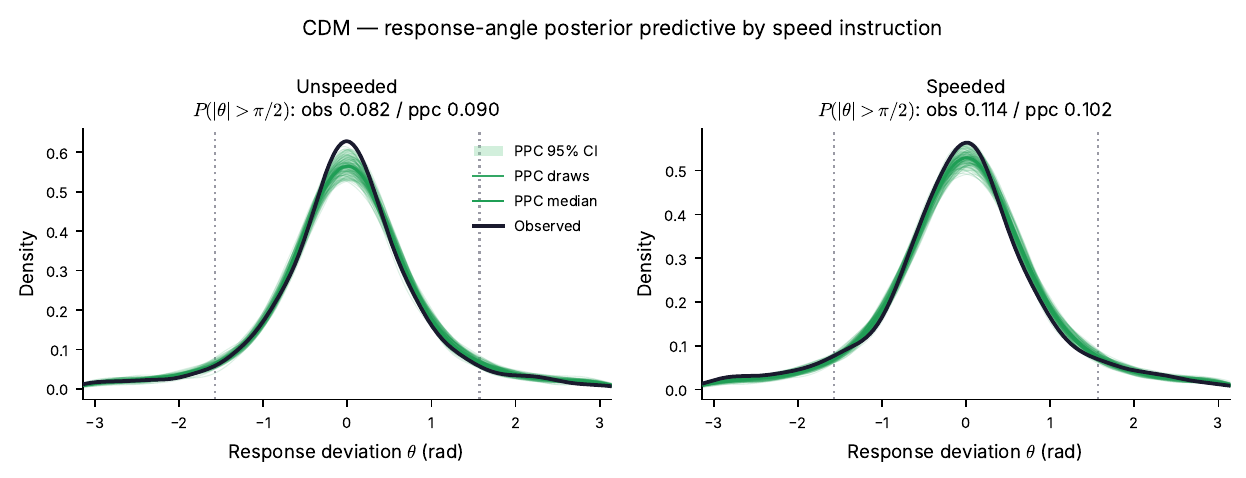}
  \caption[CDM, speed instruction: response-dimension posterior predictive]{\textbf{CDM: response-angle posterior predictive by speed instruction.} Observed density of the response angle $\theta$ against the posterior predictive (95\% band); annotations give observed and predicted proportions beyond $\pm\pi/2$ (dotted lines).}
  \label{fig:e2-cdm-choice_ppc-speed}
\end{figure}

\begin{figure}[htbp]
  \centering
  \includegraphics[width=\linewidth]{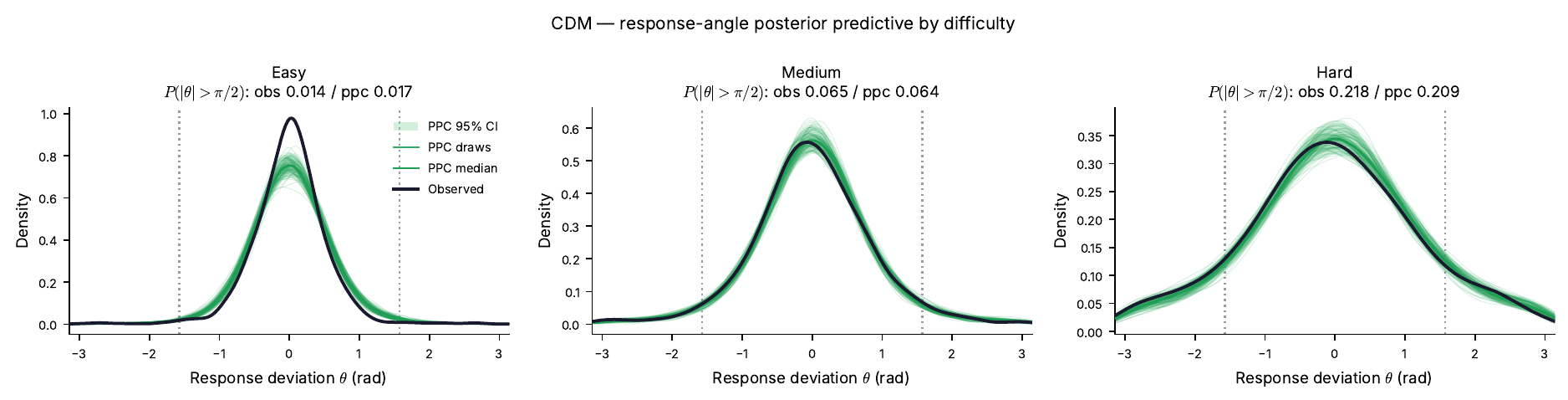}
  \caption[CDM, difficulty: response-dimension posterior predictive]{\textbf{CDM: response-angle posterior predictive by difficulty.} As \autoref{fig:e2-cdm-choice_ppc-speed}.}
  \label{fig:e2-cdm-choice_ppc-difficulty}
\end{figure}

\pagebreak

\subsection{Experiment 3: Model adaptation via fine-tuning}

\begin{figure}
    \centering
    \includegraphics[width=0.49\linewidth]{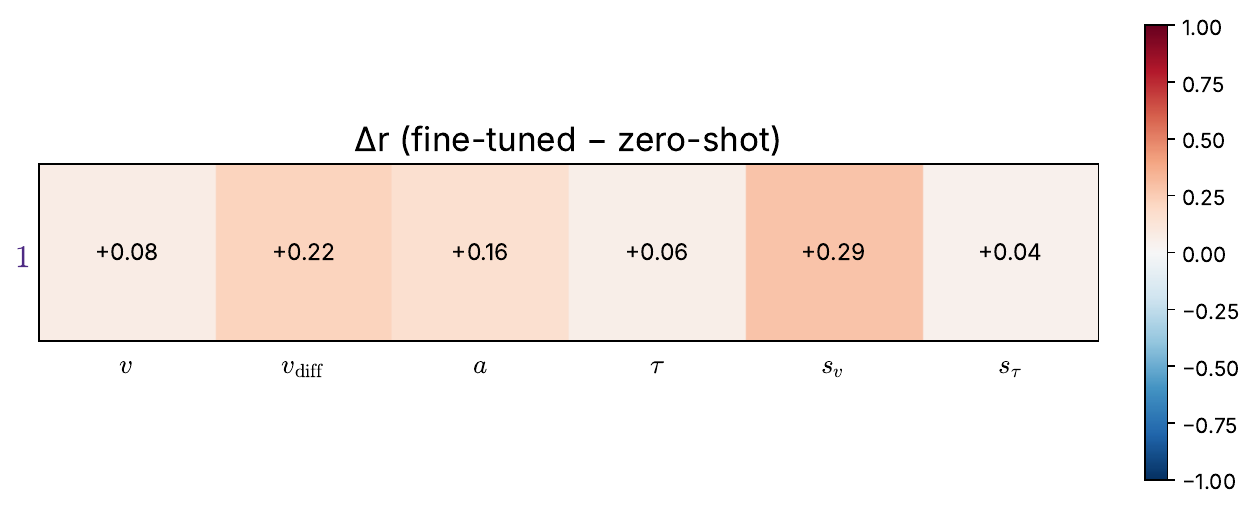}
    \includegraphics[width=0.49\linewidth]{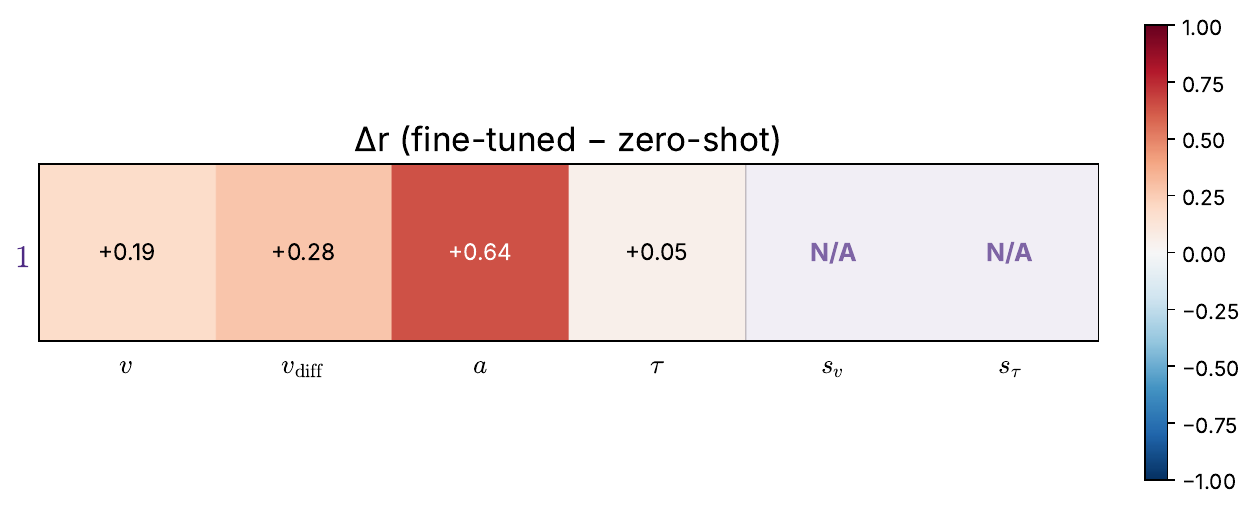}
    \includegraphics[width=0.49\linewidth]{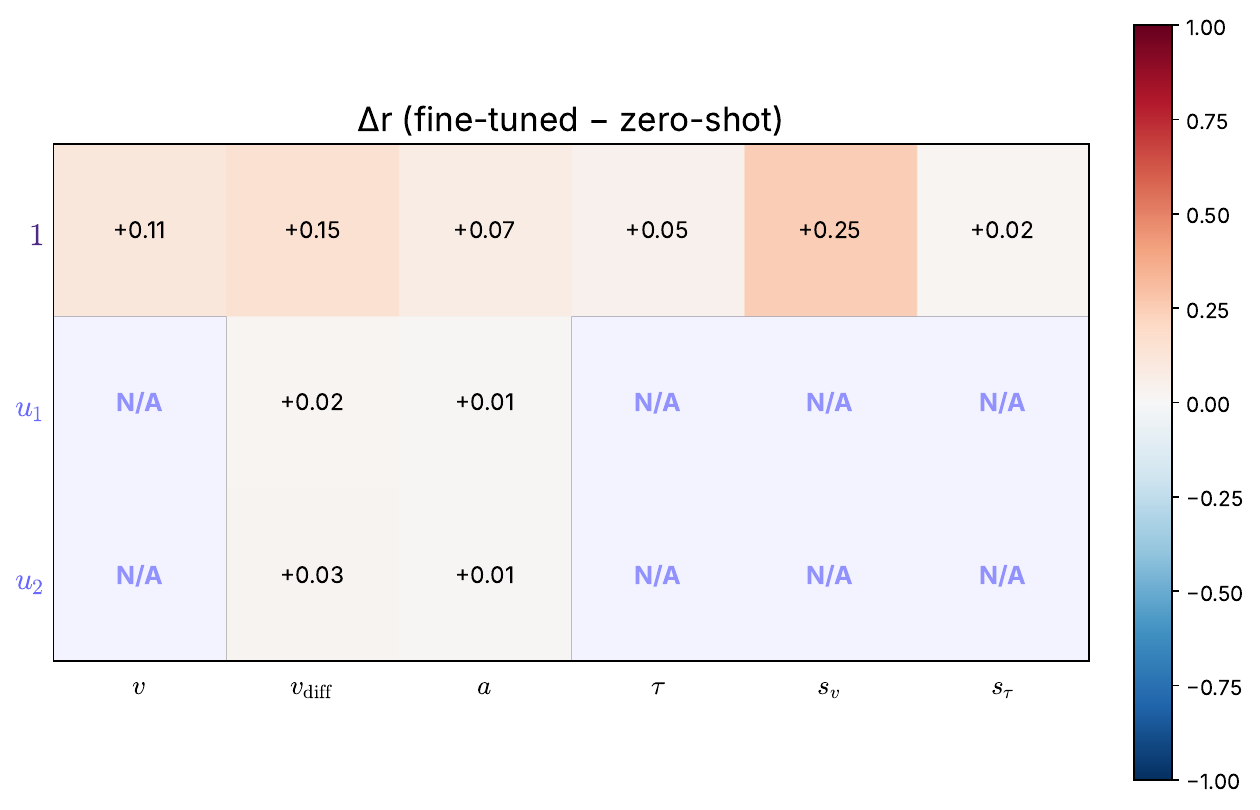}
    \includegraphics[width=0.49\linewidth]{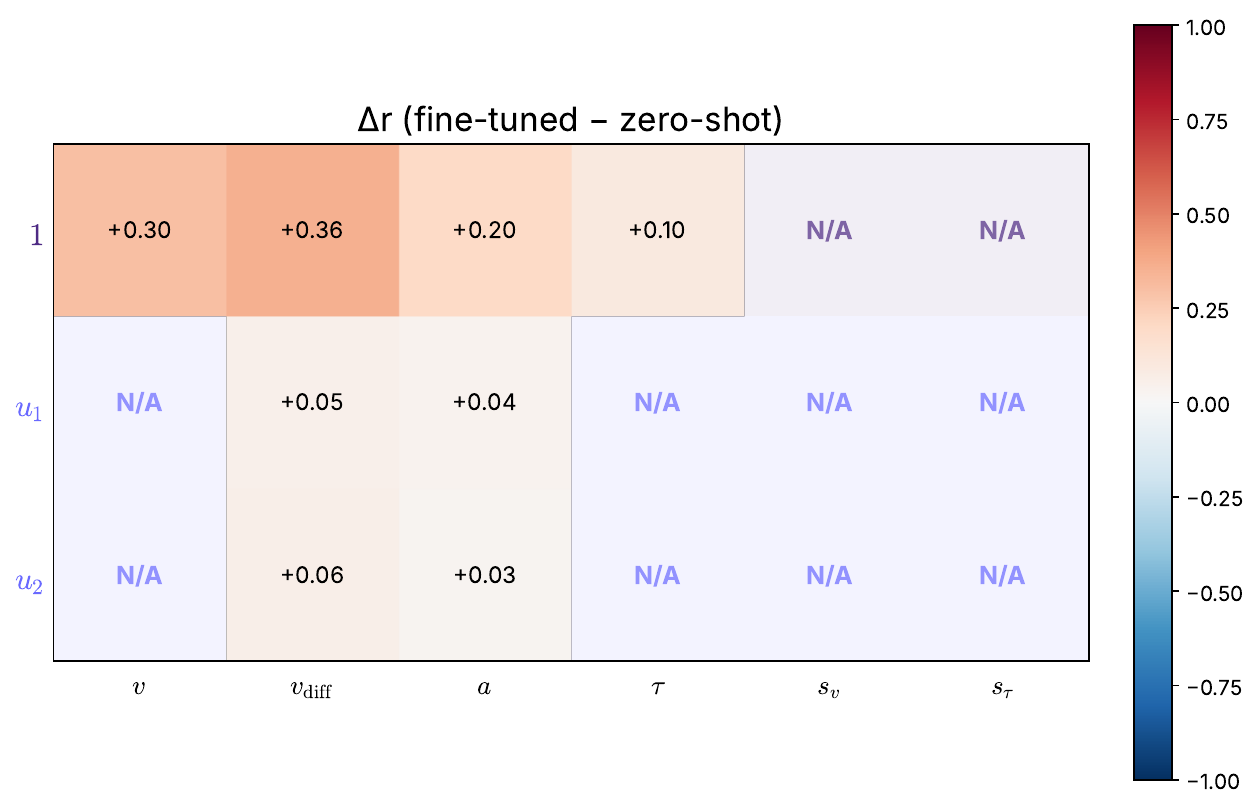}
    \caption{\textbf{Change in LBA recovery with fine-tuning.} $\Delta r$ (fine-tuned minus zero-shot correlation of posterior means with ground truth) per coefficient for the intercept-only, fixed, regressed, and fixed + regressed designs (left to right, top to bottom); the interaction design is in \autoref{fig:exp3-main}.}
    \label{fig:exp3-lba-deltas}
\end{figure}

This section reports the LBA benchmark results behind \autoref{fig:exp3-main}. The LBA was not among CogFormerV2's training families. We compare three estimators on the same test datasets for each benchmark configuration (\autoref{tab:benchmark-configs}): single-model BayesFlow baselines, CogFormerV2 before fine-tuning (zero-shot, with the LBA  under the RDM's model embedding), and CogFormerV2 after fine-tuning (\autoref{ssec:exp3-protocol}). \autoref{tab:benchmark-metrics-lba} lists their metrics. Fine-tuning reduced NRMSE in every configuration (median across configurations: 0.155 zero-shot, 0.103 fine-tuned) and brought calibration error close to that of the baselines (0.024 against 0.018). We also show the change in recovery with fine-tuning (\autoref{fig:exp3-lba-deltas}) and posterior predictive checks for the lexical-decision data of \cite{wagenmakers2008diffusion}.

\begin{figure}[t]
  \centering
    \includegraphics[width=\textwidth]{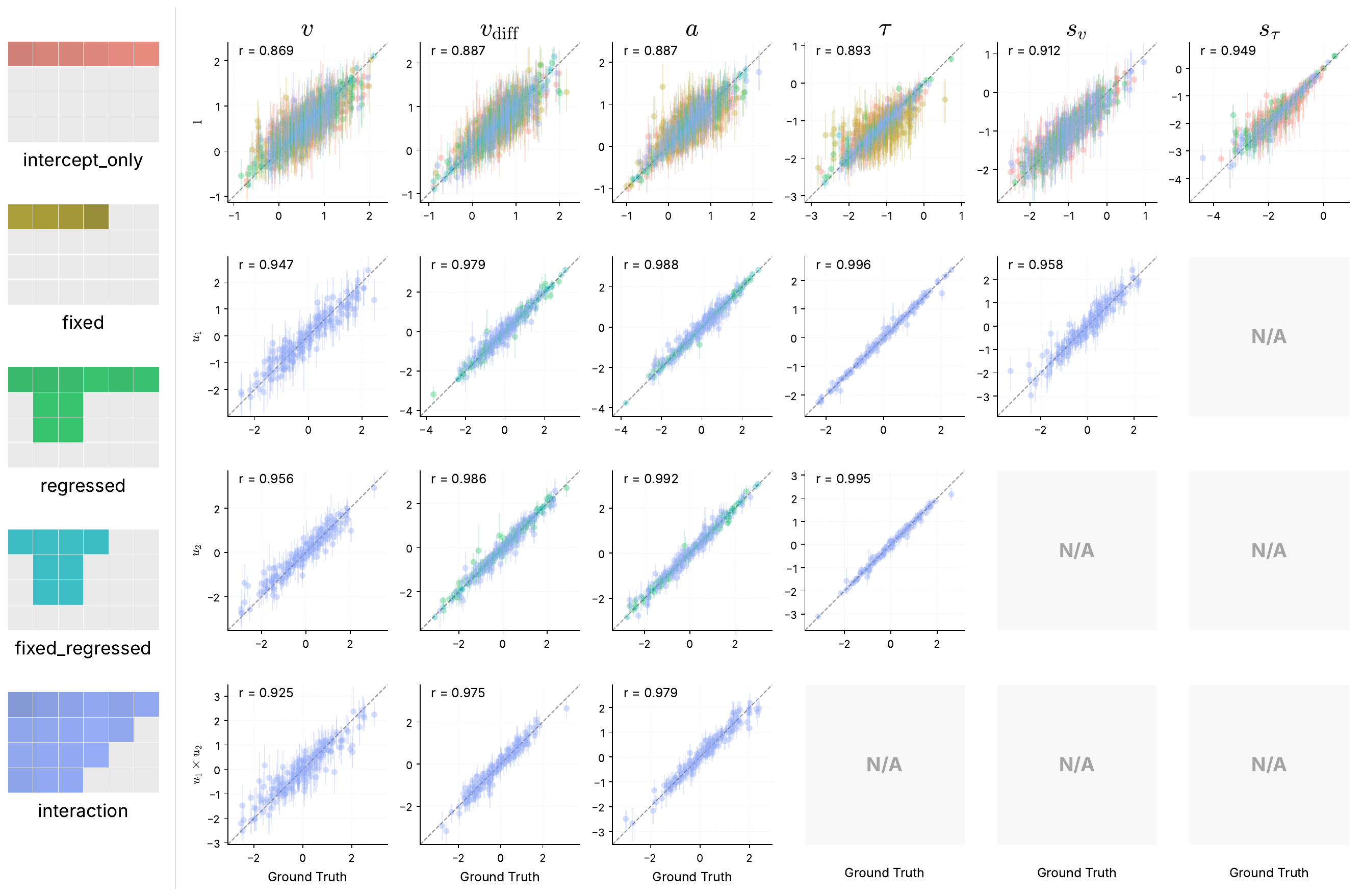}
  \caption{\textbf{LBA benchmark designs: parameter recovery, BayesFlow baselines.} Layout as in \autoref{fig:exp1-ddm-bench-ensemble-recovery}.}
    \label{fig:exp3-lba-bench-ensemble-recovery-bf}
\end{figure}

\begin{figure}[t]
  \centering
    \includegraphics[width=\textwidth]{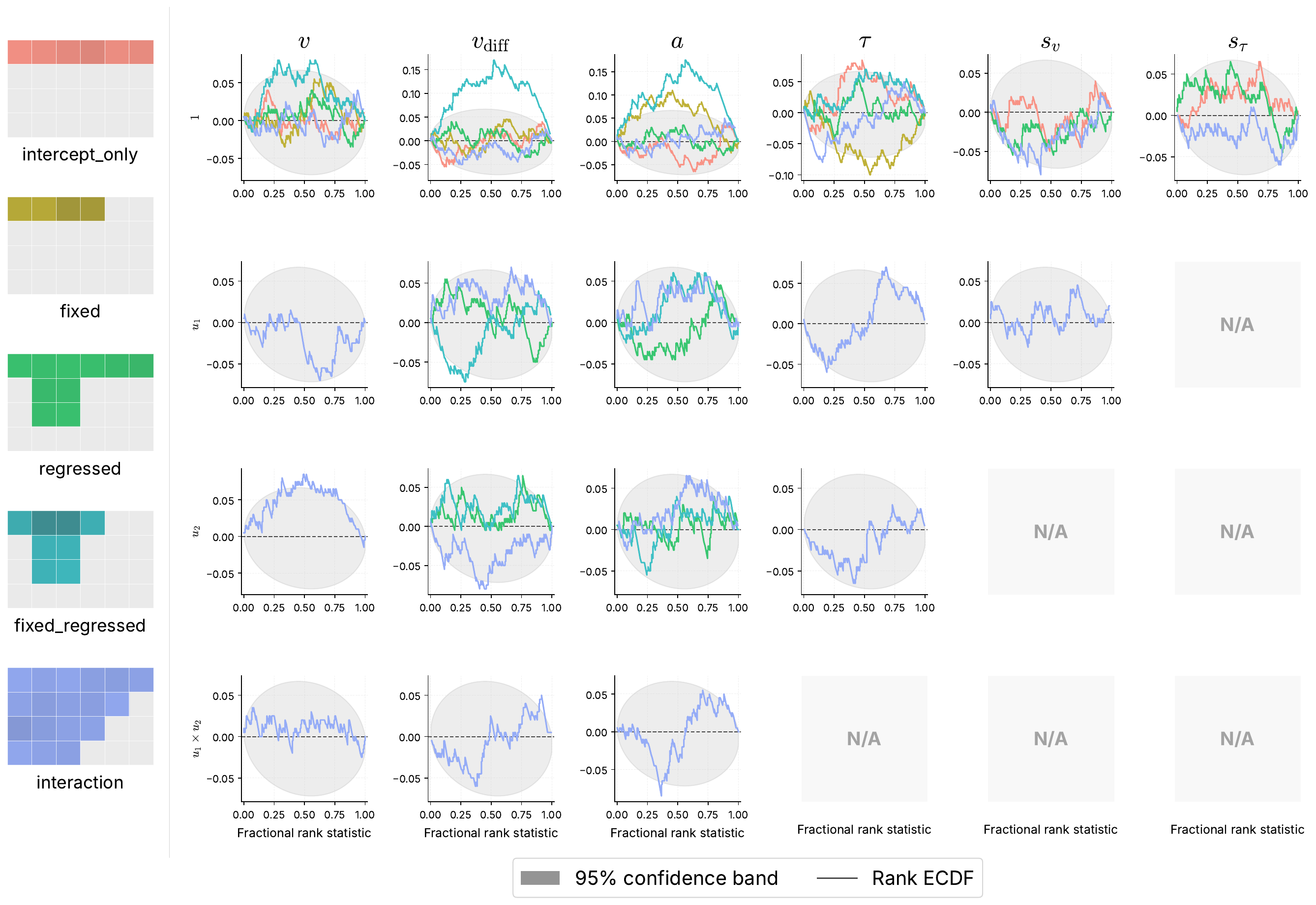}
  \caption{\textbf{LBA benchmark designs: calibration, BayesFlow baselines.} Layout as in \autoref{fig:exp1-ddm-bench-ensemble-ecdf}.}
    \label{fig:exp3-lba-bench-ensemble-ecdf-bf}
\end{figure}

\begin{figure}[t]
  \centering
    \includegraphics[width=\textwidth]{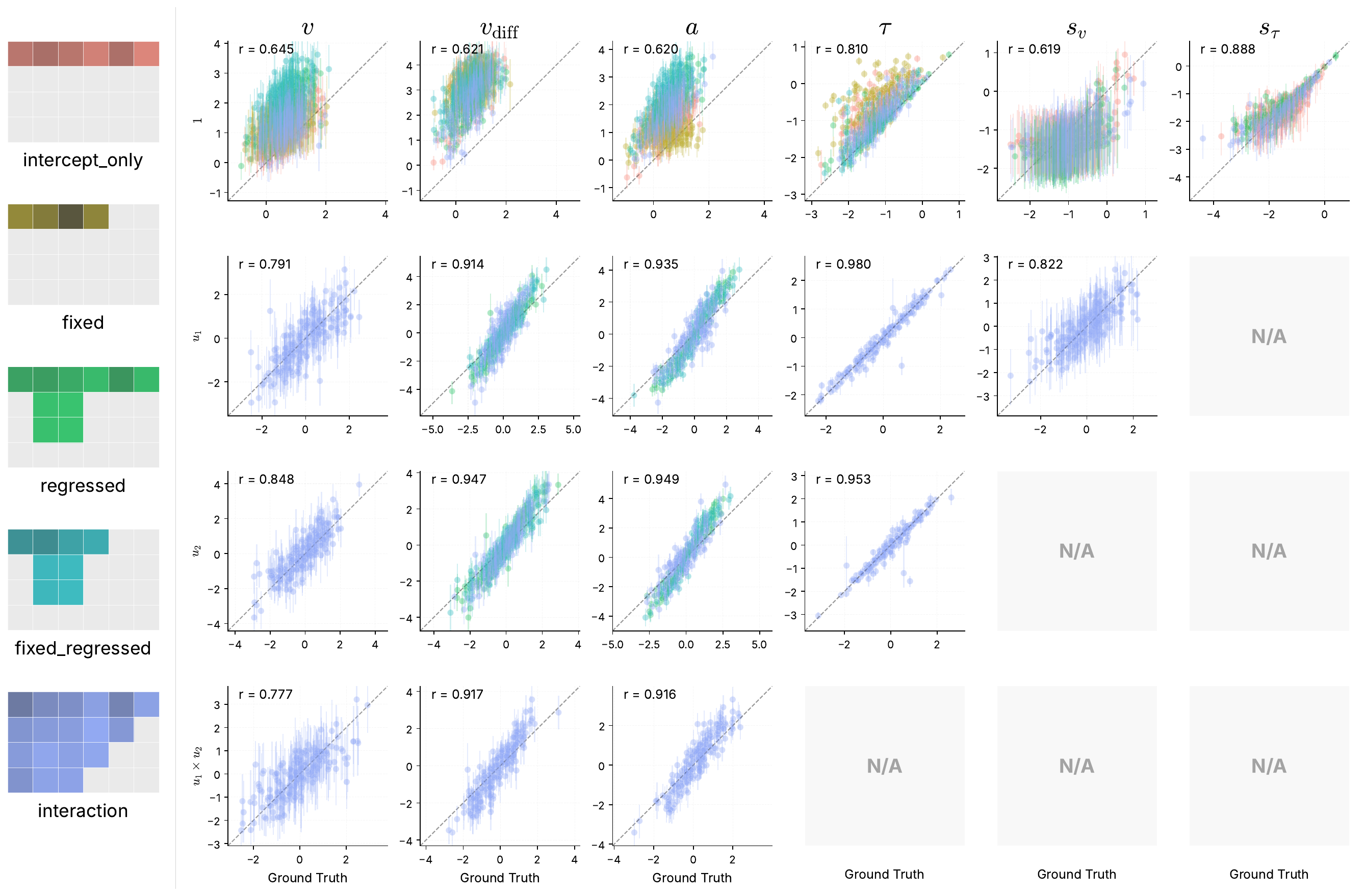}
  \caption{\textbf{LBA benchmark designs: parameter recovery, zero-shot CogFormerV2.} Before fine-tuning, with the LBA  under the RDM's model embedding. Layout as in \autoref{fig:exp1-ddm-bench-ensemble-recovery}.}
    \label{fig:exp3-lba-bench-ensemble-recovery-zs}
\end{figure}

\begin{figure}[t]
  \centering
    \includegraphics[width=\textwidth]{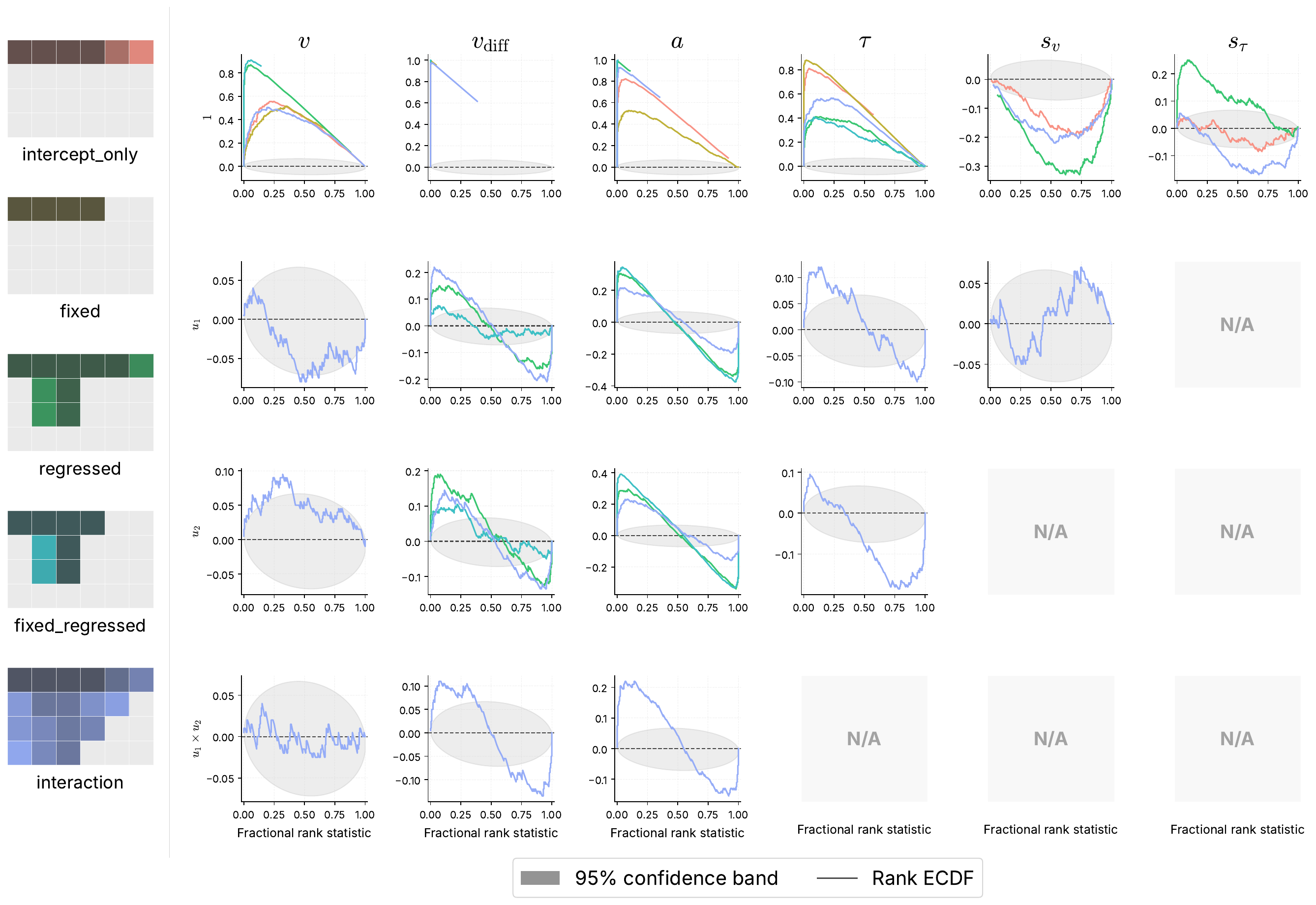}
  \caption{\textbf{LBA benchmark designs: calibration, zero-shot CogFormerV2.} Before fine-tuning, with the LBA  under the RDM's model embedding. Layout as in \autoref{fig:exp1-ddm-bench-ensemble-ecdf}.}
    \label{fig:exp3-lba-bench-ensemble-ecdf-zs}
\end{figure}

\begin{figure}[t]
  \centering
    \includegraphics[width=\textwidth]{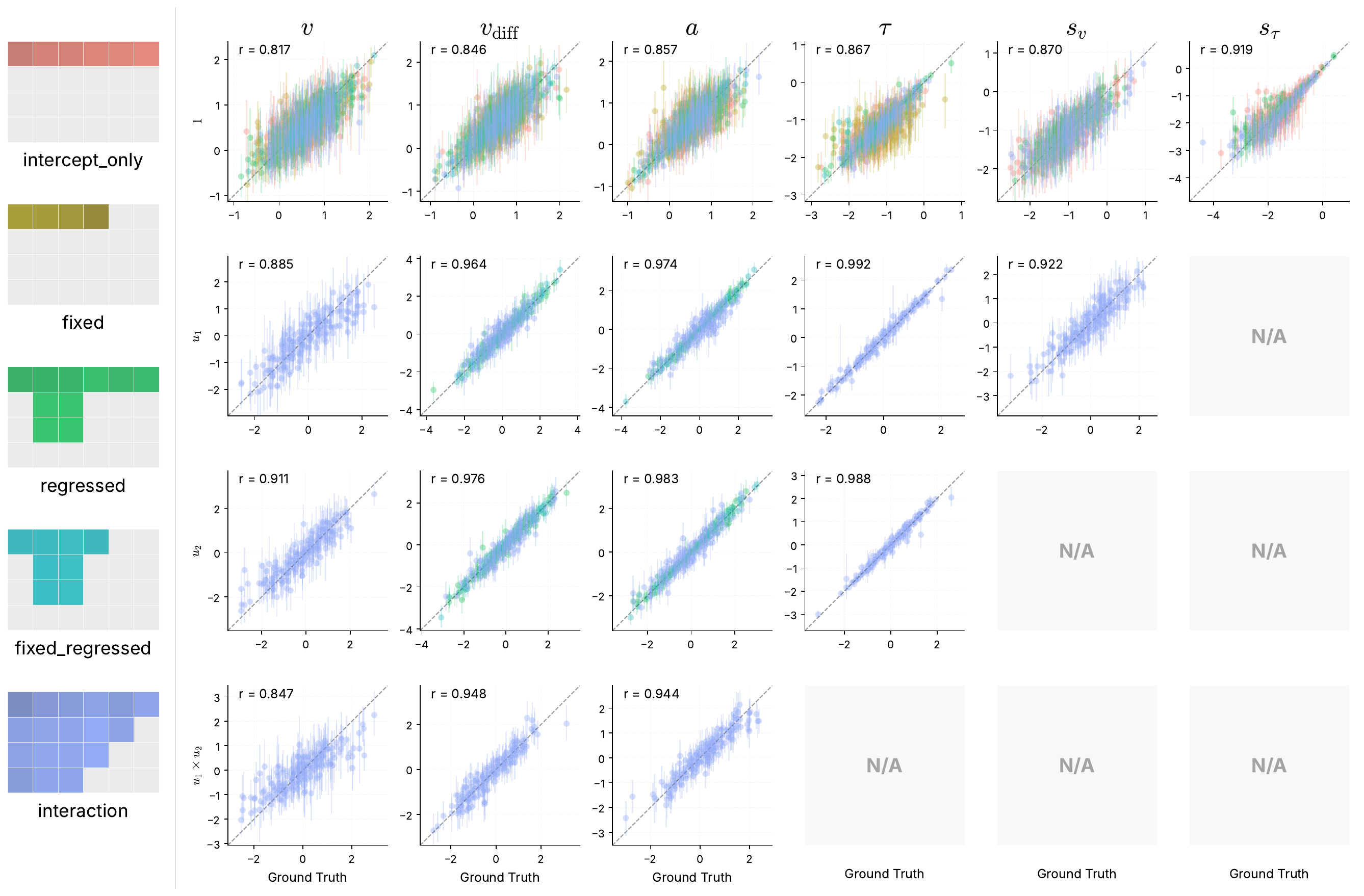}
  \caption{\textbf{LBA benchmark designs: parameter recovery, fine-tuned CogFormerV2.} After fine-tuning on all four families. Layout as in \autoref{fig:exp1-ddm-bench-ensemble-recovery}.}
    \label{fig:exp3-lba-bench-ensemble-recovery-ft}
\end{figure}

\begin{figure}[t]
  \centering
    \includegraphics[width=\textwidth]{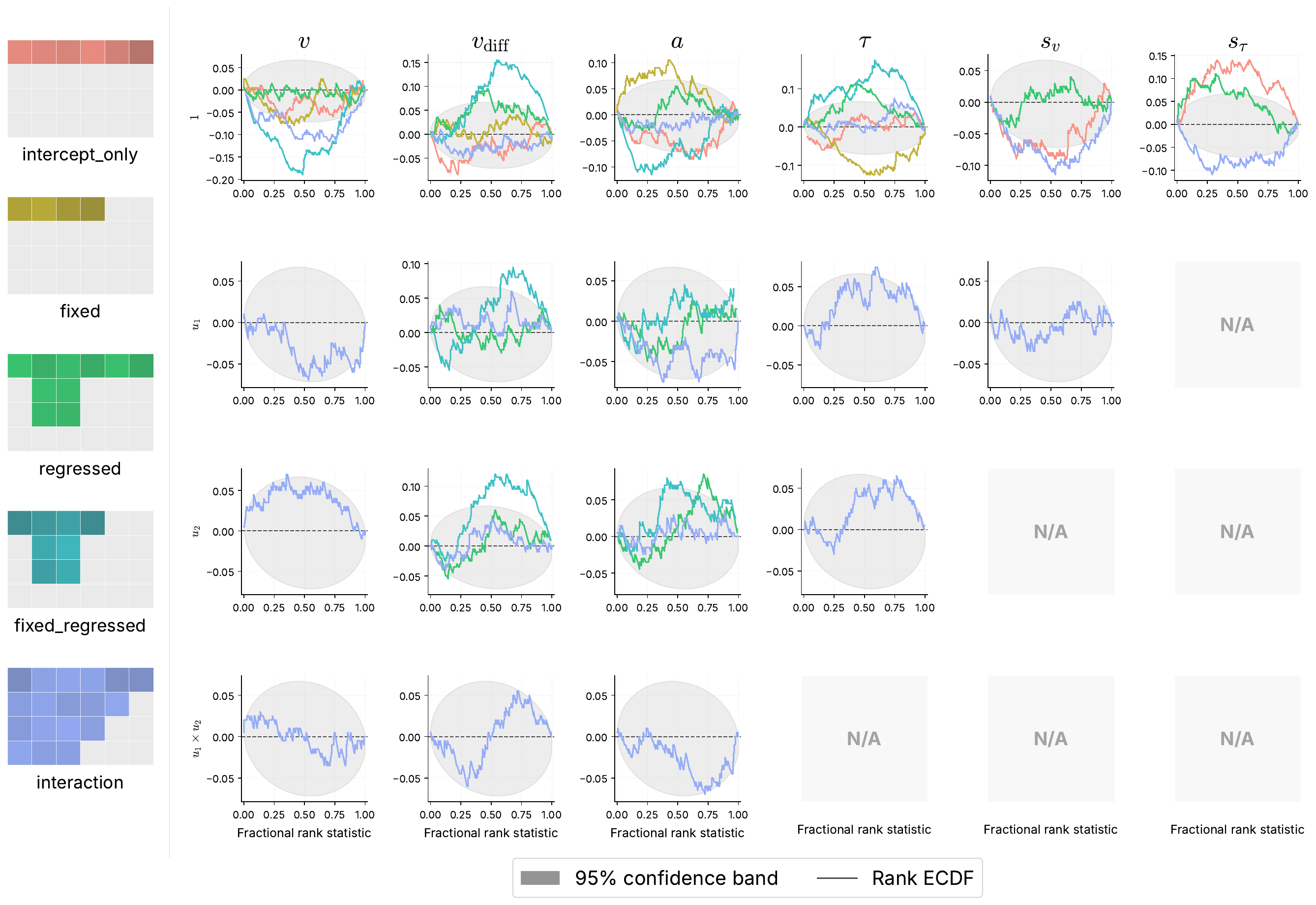}
  \caption{\textbf{LBA benchmark designs: calibration, fine-tuned CogFormerV2.} After fine-tuning on all four families. Layout as in \autoref{fig:exp1-ddm-bench-ensemble-ecdf}.}
    \label{fig:exp3-lba-bench-ensemble-ecdf-ft}
\end{figure}

\begin{table}[t]
\centering
\small
\caption{\textbf{LBA benchmark metrics.} Median $\pm$ SEM over each configuration's parameters for the BayesFlow baselines and for CogFormerV2 before (zero-shot) and after fine-tuning, on the same test datasets. Zero-shot, the LBA's model embedding is copied from the RDM's. Metrics are defined in \autoref{ssec:metrics}.}
\label{tab:benchmark-metrics-lba}
\begin{tabularx}{\textwidth}{llCCC}
\toprule
\textbf{Configuration} & \textbf{Method} & \textbf{NRMSE} ($\downarrow$) & \textbf{Cal. Error} ($\downarrow$) & \textbf{Post. Contr.} ($\uparrow$) \\
\midrule
Intercept only & \textbf{Baseline} & $0.137 \pm 0.010$ & $0.018 \pm 0.003$ & $0.648 \pm 0.047$ \\
 & \textbf{CogFormerV2} (zero-shot) & $0.232 \pm 0.102$ & $0.263 \pm 0.078$ & $0.654 \pm 0.089$ \\
 & \textbf{CogFormerV2} (fine-tuned) & $0.144 \pm 0.009$ & $0.024 \pm 0.003$ & $0.608 \pm 0.045$ \\
\midrule
Fixed & \textbf{Baseline} & $0.149 \pm 0.021$ & $0.023 \pm 0.002$ & $0.697 \pm 0.065$ \\
 & \textbf{CogFormerV2} (zero-shot) & $0.298 \pm 0.136$ & $0.324 \pm 0.061$ & $0.608 \pm 0.184$ \\
 & \textbf{CogFormerV2} (fine-tuned) & $0.149 \pm 0.019$ & $0.041 \pm 0.010$ & $0.697 \pm 0.055$ \\
\midrule
Regressed & \textbf{Baseline} & $0.061 \pm 0.010$ & $0.016 \pm 0.002$ & $0.979 \pm 0.019$ \\
 & \textbf{CogFormerV2} (zero-shot) & $0.128 \pm 0.075$ & $0.235 \pm 0.052$ & $0.923 \pm 0.063$ \\
 & \textbf{CogFormerV2} (fine-tuned) & $0.073 \pm 0.013$ & $0.017 \pm 0.006$ & $0.951 \pm 0.034$ \\
\midrule
Fixed + regressed & \textbf{Baseline} & $0.026 \pm 0.006$ & $0.023 \pm 0.003$ & $0.998 \pm 0.007$ \\
 & \textbf{CogFormerV2} (zero-shot) & $0.135 \pm 0.113$ & $0.415 \pm 0.063$ & $0.853 \pm 0.098$ \\
 & \textbf{CogFormerV2} (fine-tuned) & $0.063 \pm 0.009$ & $0.036 \pm 0.007$ & $0.970 \pm 0.016$ \\
\midrule
Interaction & \textbf{Baseline} & $0.075 \pm 0.008$ & $0.017 \pm 0.003$ & $0.938 \pm 0.020$ \\
 & \textbf{CogFormerV2} (zero-shot) & $0.155 \pm 0.038$ & $0.190 \pm 0.034$ & $0.817 \pm 0.050$ \\
 & \textbf{CogFormerV2} (fine-tuned) & $0.105 \pm 0.009$ & $0.018 \pm 0.003$ & $0.866 \pm 0.027$ \\
\midrule
All configurations & \textbf{Baseline} & $0.074 \pm 0.006$ & $0.018 \pm 0.002$ & $0.938 \pm 0.018$ \\
 & \textbf{CogFormerV2} (zero-shot) & $0.155 \pm 0.035$ & $0.235 \pm 0.024$ & $0.817 \pm 0.037$ \\
 & \textbf{CogFormerV2} (fine-tuned) & $0.103 \pm 0.006$ & $0.024 \pm 0.003$ & $0.870 \pm 0.020$ \\
\bottomrule
\end{tabularx}
\end{table}

\begin{figure}
    \centering
    \includegraphics[width=\linewidth]{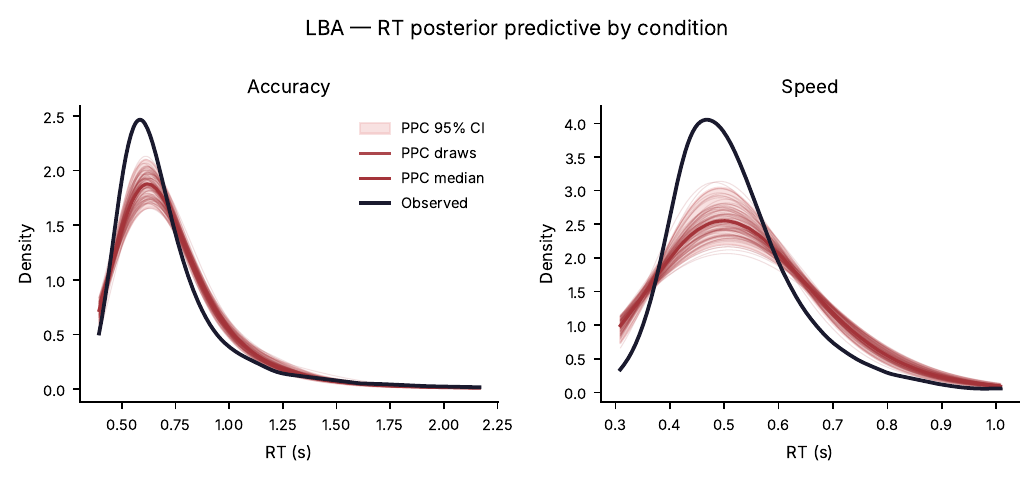}
    \caption{\textbf{LBA: RT posterior predictive by instruction.} Observed RT density (black) in the lexical-decision data of \cite{wagenmakers2008diffusion} against the fine-tuned CogFormerV2's posterior predictive: median (line), 95\% interval (band), and individual draws (thin lines). Panels show the accuracy and speed instructions.}
    \label{fig:exp3-ppc-rt-condition}
\end{figure}

\begin{figure}
    \centering
    \includegraphics[width=\linewidth]{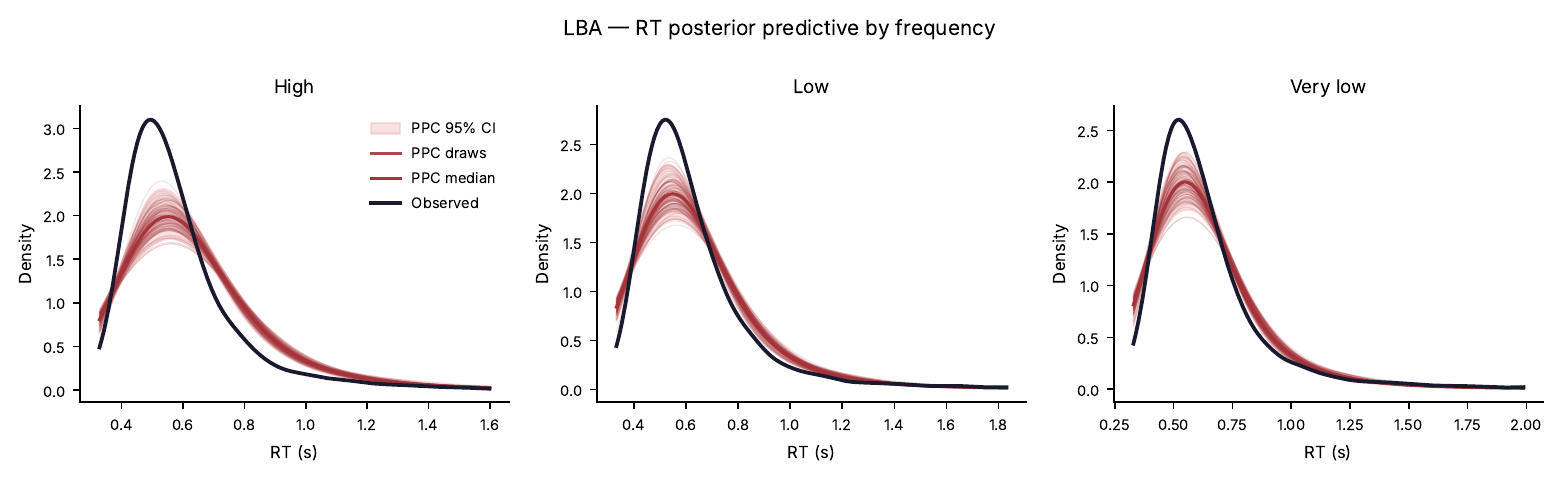}
    \caption{\textbf{LBA: RT posterior predictive by word frequency.} As \autoref{fig:exp3-ppc-rt-condition}, for high-, low-, and very-low-frequency words.}
    \label{fig:exp3-ppc-rt-frequency}
\end{figure}

\begin{figure}
    \centering
    \includegraphics[width=\linewidth]{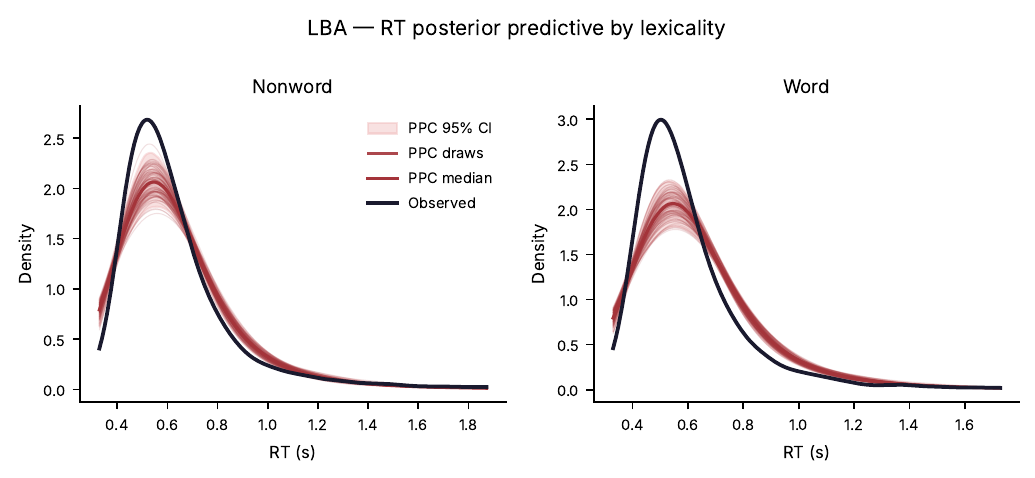}
    \caption{\textbf{LBA: RT posterior predictive by lexicality.} As \autoref{fig:exp3-ppc-rt-condition}, for nonwords and words.}
    \label{fig:exp3-ppc-rt-lexicality}
\end{figure}

\newpage

\subsection{Experiment 4: Transfer of encoder representation}
\label{app:experiment-4-figures}

\begin{figure}[!ht]
    \centering
    \includegraphics[width=\linewidth]{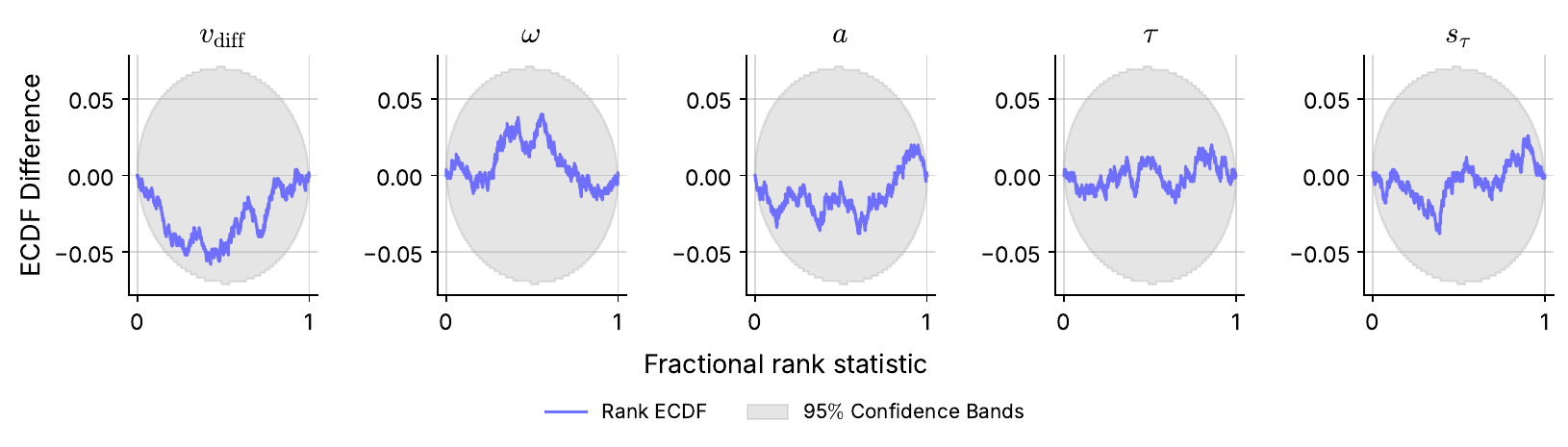}
    \includegraphics[width=\linewidth]{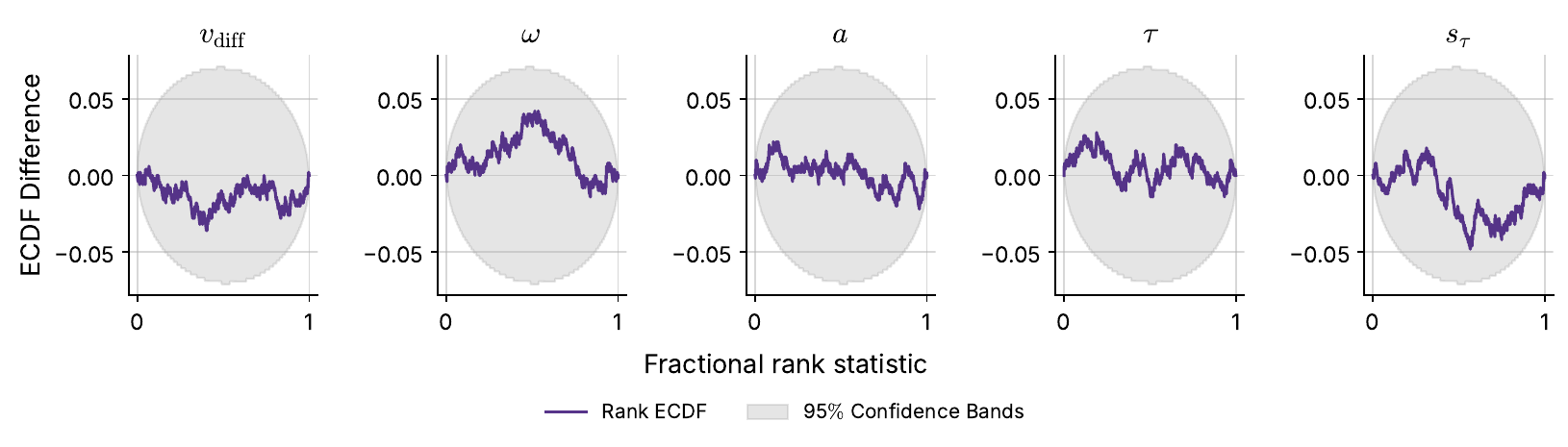}
    \caption{\textbf{aDDM: calibration of the two arms.} Rank ECDF minus the uniform CDF with 95\% bands for the task-specific Set Transformer (\emph{top}) and the frozen CogFormerV2 encoder (\emph{bottom}), each on its own 500 validation datasets.}
    \label{fig:experiment-4-calibration}
\end{figure}

\begin{figure}[!ht]
    \centering
    \includegraphics[width=\linewidth]{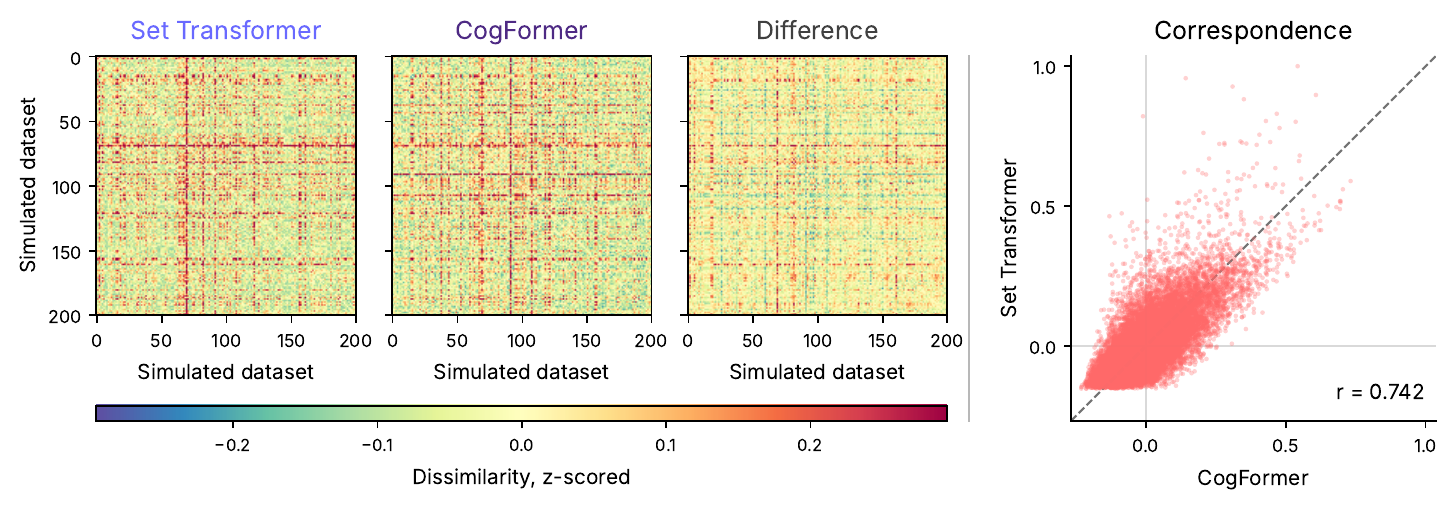}
    \caption{\textbf{aDDM: representational similarity of the two summaries.} Crossnobis dissimilarity matrices over 200 aDDM datasets for the Set Transformer summary, the frozen CogFormerV2 encoder, and their difference, each z-scored over off-diagonal entries; \textit{right}: off-diagonal entries against each other, with their Pearson correlation $r$.}
    \label{fig:experiment-4-rdm-rsa}
\end{figure}

This section reports additional diagnostics for the aDDM probe of Experiment~4 (\autoref{ssec:exp4-protocol}): calibration of both estimators (\autoref{fig:experiment-4-calibration}) and the representational similarity of their summaries (\autoref{fig:experiment-4-rdm-rsa}).

\end{appendices}

\pagebreak

\end{document}